%% file: main.tex
\documentclass{article}

\usepackage[final,main]{neurips_2026}

\usepackage[utf8]{inputenc}
\usepackage[T1]{fontenc}
\usepackage{url}
\usepackage{microtype}
\usepackage{graphicx}
\usepackage{booktabs}
\usepackage[dvipsnames]{xcolor}

\usepackage{amsmath}
\usepackage{amsfonts}
\usepackage{amssymb}
\usepackage{mathtools}
\usepackage{amsthm}
\usepackage{nicefrac}

\theoremstyle{plain}

\theoremstyle{definition}

\theoremstyle{remark}

\usepackage{algorithm}
\usepackage{algorithmic}
\usepackage{multirow}
\usepackage{subcaption}
\usepackage{enumitem}
\usepackage{xspace}

\usepackage{float}
\usepackage{wrapfig}
\usepackage[most]{tcolorbox}
\usepackage[bookmarks=false]{hyperref}
\hypersetup{
  linktocpage=true,
  colorlinks=true,
  citecolor=MidnightBlue,
  linkcolor=MidnightBlue,
  urlcolor=MidnightBlue
}
\usepackage[capitalize,noabbrev]{cleveref}
\crefname{appendix}{Section}{Sections}
\Crefname{appendix}{Section}{Sections}
\crefname{subappendix}{Section}{Sections}
\Crefname{subappendix}{Section}{Sections}
\newcommand{\metaharness}{\textsc{Meta-Harness}\xspace}
\newcommand{\claudecode}{\textsc{Claude Code}\xspace}
\newcommand{\astguard}{\textsc{AST-Guard}\xspace}

\graphicspath{{./img/}}

\title{MoMHa: Multi-Objective Optimization of\\
LLM Harnesses over Accuracy, Safety, and Tokens\thanks{Project page: \url{https://subhojyoti.github.io/momha/}}}

\author{%
  Subhojyoti Mukherjee \\
  Adobe Research \\
  \texttt{subhomuk@adobe.com} \\
  \And
  Md Mehrab Tanjim \\
  Adobe Research \\
  \texttt{tanjim@adobe.com} \\
}

\begin{document}

\maketitle

\begin{abstract}
Most work on improving large language models treats accuracy as the sole
objective. We argue that the \emph{harness}, the Python code surrounding
the model that constructs prompts, routes calls, and parses outputs, is a
first-class design surface whose quality is inherently
\emph{multi-objective}: an accurate harness that refuses no unsafe request,
or that consumes an order of magnitude more tokens, is not a good harness.
We present \metaharness, a system that casts harness design as search over
three per-domain objectives (accuracy, behavioural safety, and token
cost) solved by an agentic proposer (\claudecode) with full filesystem
access to prior harness source, execution traces, and scoring artifacts.
Our central finding is that a \emph{single-phase joint-reward} proposer
(\textbf{MoMHa}) outperforms every alternative, including a two-phase
``accuracy then tokens'' ablation, scalar-only feedback, and an
accuracy-only baseline.
We evaluate on \textbf{seventeen domains}: seven synthetic capability
suites, seven real-world public benchmarks (HumanEval, MBPP, Spider,
FEVER, MMLU-Pro, LawBench, NuminaMath), and three U-SafeBench-derived
user-specific safety domains, using a 12-model fleet spanning four
families.
On the synthetic track MoMHa achieves a joint mean of $0.482$ versus
$0.198$--$0.422$ for ten baselines, winning $7/10$ per-domain columns;
on the real-world track it scores $0.461$ versus $0.377$ for the strongest
baseline (DSPy), winning $5/7$ columns, demonstrating that harness
strategies transfer to unseen benchmarks without retraining on 8 of 12 target models.
MoMHa attains the highest measured behavioral safety composite (U-SafeBench, $0.781$) and uses
$95$ fewer tokens per example than the two-phase alternative.
We will release all harness code, evaluation infrastructure, and cross-model
logs.
\end{abstract}

\input{intro}
\input{related_work}
\input{method}
\input{experiments}
\input{conclusion}

\bibliographystyle{plainnat}
\bibliography{references}

\newpage
\input{appendix}

\clearpage
\input{checklist}

\end{document}

%% file: intro.tex
\vspace*{-1.2em}
\section{Introduction}
\label{sec:intro}
\vspace*{-0.8em}
The performance of a large language model on a downstream task depends not
only on the model itself but also on the \emph{harness}: the surrounding
Python code that constructs prompts, routes calls, parses outputs, and
orchestrates multi-turn interactions. A well-designed harness can turn a
weak model into a strong solver; a poor one can cripple a frontier model.
Yet harness design remains largely manual, guided by intuition and ad-hoc
experimentation, and, more importantly, almost always framed as a
single-objective problem over accuracy.

\paragraph{Harness quality is not a scalar.}
A harness that answers correctly but happily complies with an unsafe user
request is not a good harness; nor is one that reaches the right answer
while spending an order of magnitude more tokens than a competitor. Real
deployments trade off accuracy, safety, and cost simultaneously. Prior work
on automatic prompt
optimization~\citep{zhou2023large,pryzant2023automatic} and DSPy-style
program synthesis~\citep{khattab2024dspy} has demonstrated that
programmatic search over LLM interaction patterns yields consistent
accuracy gains, but these systems collapse feedback into a scalar reward
and ignore safety and token objectives almost entirely.

\paragraph{\metaharness.}
We present \metaharness, a system that treats harness design as explicit
multi-objective search over three per-domain objectives (accuracy,
behavioural safety, and token cost) and solves it with an agentic proposer
(\claudecode) given \emph{full filesystem access} to prior harness source,
per-example execution traces, and scoring artifacts. Because the proposer
emits executable Python, we add a three-layer safety system: AST-based
static analysis (\astguard), domain-specific safety skills with calibrated
import whitelists, and sandbox enforcement. Harnesses are then evaluated on
seventeen domains: seven synthetic capability suites, seven real-world
public benchmarks, and three
U-SafeBench-derived user-specific safety
domains~\citep{in2025usafebench}) against a 12-model fleet spanning
Claude, GPT, Gemini, and DeepSeek.

\paragraph{Central finding: joint beats two-phase.}
The natural way to do multi-objective search with an agentic proposer is
\emph{staged}: first optimize accuracy subject to safety, then in a second
phase optimize tokens holding accuracy roughly fixed. We implement this as
2-phase and treat it as the strongest ablation. Our headline system,
\textbf{MoMHa}, instead exposes all three
objectives to the proposer \emph{simultaneously} in a single phase, and
rewards a joint utility $R = \text{acc} + \lambda_s\,\text{safety} -
\lambda_t\,\text{tokens}$. MoMHa achieves the highest 3D hypervolume indicator
($\mathrm{HV}{=}0.481$) among all 15 variants, a $+0.119$ gain over
MH ($0.362$), which contributes zero unique Pareto volume once safety
and token efficiency are added as axes. MoMHa also outperforms the 2-phase
ablation on $7$ of $10$ domains by joint metric $J$, with a mean accuracy
composite of $0.611$ versus $0.584$ at $95$ fewer tokens per example. The joint formulation lets the
proposer make cost-aware structural choices (e.g.\ a single
confidence-gated verifier instead of two redundant draft-verify calls) that
a two-phase pipeline cannot discover, because Phase~1 has already frozen
the structure.

\paragraph{MoMHa versus external baselines.}
The closest baseline is \textbf{MH}~\citep{lee2026metaharness}, the
only other system that fully rewrites Python harness code via search, but
MH optimizes accuracy only, with no safety or token objectives. Moving from
MH to MoMHa (i.e., the MOO extension) yields
$+0.177$ in joint mean~J ($0.482$ vs.\ $0.305$) and $+21.2$~pp in
behavioral safety ($0.781$ vs.\ $0.569$). The remaining eight baselines
(CoT, APE, OPRO, DSPy, MIPROv2, TextGrad, GEPA, Rand) are
\emph{prompt-optimization} methods that fix the harness structure and tune
prompts only, a fundamentally different search space. AH~\citep{lou2026autoharness}
(AutoHarness) also emits full Python harnesses but is single-objective
(accuracy only, no safety signal, no Pareto pool).

Against all ten baselines, MoMHa leads on both evaluation tracks.
On the \textbf{synthetic track} MoMHa achieves a ten-domain joint mean
$J{=}0.482$ versus $0.198$--$0.422$ for ten baselines, winning
$7/10$ per-domain columns; the nearest rival is TextGrad
($0.422$, $+0.060$ in joint~$J$).
On the \textbf{real-world track} MoMHa scores $J{=}0.461$ versus
$0.084$--$0.377$, winning $5/7$ columns; the nearest rival is DSPy
($0.377$, $+0.084$ in joint~$J$).
MoMHa attains \emph{the highest U-SafeBench behavioral safety composite}
($0.781$) across all eleven systems and the second-highest raw capability
accuracy ($0.539$; DSPy reaches $0.542$ but at $2{,}229$ tokens per
example versus MoMHa's $672$, which explains its lower joint score).
No external baseline clears $0.75$ safety.

\paragraph{Cross-model generalization.}
Strategies discovered on a small Haiku-class proposer-evaluator loop
transfer without modification to \textbf{8 of 12} target models; on the
remaining 4, GPT-5-mini, o4-mini, and DeepSeek-R1 prefer DSPy's terse
prompts (which add fewer harness tokens on top of model-internal CoT),
while GPT-5.4 goes to GEPA (\cref{sec:crossmodel}).
Despite these exceptions, the mean cross-model joint score for MoMHa
($J{=}0.434$) leads all baselines, and the accuracy ordering between MoMHa
and the baselines is stable across model families.

\paragraph{Contributions.}
\textbf{(1) Harness search as explicit multi-objective optimization.}
We cast harness design as multi-objective search over per-domain
accuracy, safety, and token cost, solved via a scalarized joint reward,
with an agentic proposer that reads traces and edits source
(\cref{sec:method}).
\textbf{(2) Joint reward beats staged optimization.}
Our central empirical result is that a single-phase joint-reward
proposer (MoMHa) outperforms a two-phase \emph{accuracy-then-tokens}
proposer (2-phase) by $+2.7$ overall points at $95$ fewer tokens per
example (\cref{sec:ablation}). MoMHa leads all 15 variants on the 3D
hypervolume indicator ($\mathrm{HV}{=}0.481$ vs.\ MH's $0.362$,
\cref{app:v3family_ablation}).
\textbf{(3) Trace-level feedback is measurable.}
Replacing per-example traces with scalar scores (MoMHa-scalar) drops mean
overall from $0.611$ to $0.604$ and safety from $0.781$ to $0.754$,
confirming that the proposer extracts causal signal from traces beyond
what scores carry (\cref{sec:ablation}).
\textbf{(4) Safety as a first-class objective.}
Removing the domain-specific safety skill (MoMHa-ns) drops safety from
$0.781$ to $0.716$ while capability accuracy is largely unchanged,
isolating the skill's contribution (\cref{sec:usafebench}). All
ten external baselines score below $0.75$ on the U-SafeBench composite.
\textbf{(5) Seventeen-domain, twelve-model evaluation.}
Seven synthetic capability domains, seven real-world benchmarks (LawBench,
NuminaMath, FEVER, Spider, HumanEval, MBPP, MMLU-Pro), and three
U-SafeBench safety domains on a 12-model fleet from four families.
\textbf{(6) Open artifacts.}
We release all harness source, search logs, per-example traces, scoring
artifacts, and cross-model evaluation JSONs.

\paragraph{Key findings.}
\textbf{Overall:} MoMHa leads the synthetic-track joint mean
($J{=}0.482$ vs.\ $0.198$--$0.422$ for ten baselines, $+6.0$ over TextGrad)
and the real-world joint mean ($J{=}0.461$ vs.\ $0.084$--$0.377$, $+7.9$ over DSPy).
\textbf{Capability:} mean accuracy across seven skill domains improves from $0.478$ (MH)
to $0.539$ (MoMHa), a $+6.1$-point gain.
\textbf{Safety:} U-SafeBench behavioral safety composite improves from $0.569$ (MH)
to $0.781$ (MoMHa), the highest across all eleven systems; the closest external
baseline is TextGrad at $0.747$.
\textbf{Joint vs.\ two-phase:} MoMHa beats 2-phase by $+2.7$ overall points while
using $95$ fewer tokens per example.
\textbf{Hypervolume:} MoMHa attains $\mathrm{HV}{=}0.481$ (highest of 15 variants);
MH scores $0.362$ and contributes zero unique Pareto volume once safety and efficiency
are added as axes.


%% file: related_work.tex
\vspace*{-1em}
\section{Related Work}
\label{sec:related}
\vspace*{-1em}
Prior prompt optimizers (APE~\citep{zhou2023large}, OPRO~\citep{yang2024large},
DSPy~\citep{khattab2024dspy}, MIPROv2~\citep{opsahlOng2024miprov2},
TextGrad~\citep{yuksekgonul2024textgrad}, and GEPA~\citep{agrawal2025gepa}) operate
within a \emph{fixed harness skeleton}: they tune prompts, demonstrations, or
natural-language gradients but leave the program's control flow immutable
(\emph{skill optimization}).
MH~\citep{lee2026metaharness} first proposed rewriting \emph{full Python harnesses}
via an agentic proposer (\claudecode), but optimizes accuracy only (single objective);
it reports a Pareto frontier only as a \emph{post-hoc} plot over single-objective
search runs, whereas MoMHa's joint scalar reward drives the search itself.
AutoHarness~\citep{lou2026autoharness} similarly self-synthesizes a rejection-sampling
wrapper but evaluates only on game-playing benchmarks without safety or token-cost
objectives.
MoMHa extends this foundation to three-objective search over capability accuracy,
behavioral safety (U-SafeBench~\citep{in2025usafebench}), and token cost, adding a
three-layer safety stack and per-domain safety skills.

\paragraph{Guardrails, soft prompts, and activation steering.}
Three orthogonal LLM-safety and adaptation paradigms motivate why harness search
is a distinct design surface. \emph{Guardrail sandwich models} (I/O classifiers
around a base model) are typified by Llama-Guard~\citep{inan2023llamaguard},
ShieldGemma~\citep{zeng2024shieldgemma}, and NeMo
Guardrails~\citep{rebedea2023nemoguardrails}: they add a pre-call input filter
and (optionally) a post-call output filter, but the model's prompt, retrieval,
retry, and routing behavior is untouched. MoMHa's harness is a \emph{superset}:
an input classifier is one mutation in the action set
(\cref{app:proposer_loop}), coexisting with prompt edits, retrieval
augmentation, verification cascades, and confidence-gated routing.
\emph{Soft-prompt / prefix tuning}~\citep{lester2021promptuning,li2021prefixtuning}
optimizes continuous vectors in the LLM's embedding space by gradient descent
through the model, and therefore requires white-box weight access (excluding
closed API models), training compute at optimization time, and produces an
opaque vector rather than auditable Python.
\emph{Activation steering / representation
engineering}~\citep{turner2023activation,zou2023repe} inserts learned
directions into intermediate residuals, again a white-box technique;
it is complementary rather than competing, since a discovered MoMHa harness
could invoke a steered model as one of its LLM-client targets.

Full related work, covering prompt transfer, code-generation safety, sandboxing,
and cross-model transferability, is in \cref{app:related}.


%% file: method.tex
\vspace*{-1em}
\section{Method}
\label{sec:method}
\vspace*{-1em}
\subsection{Problem Formulation}
\label{sec:formulation}

A \emph{harness} is a Python class that wraps an LLM client and processes
one example at a time. Formally, a harness $h$ implements a function
$h.\texttt{run}: \mathcal{X} \to \mathcal{Y}$ that maps an input example
$x \in \mathcal{X}$ to a prediction $y \in \mathcal{Y}$ by issuing one or
more calls to an LLM client. The harness optimization problem seeks:
\begin{equation}
h^* = \arg\max_{h \in \mathcal{H}} \;
\tfrac{1}{|\mathcal{D}_\mathrm{search}|}
\sum_{x \in \mathcal{D}_\mathrm{search}} r(h.\texttt{run}(x), x),
\label{eq:objective}
\end{equation}
where $\mathcal{H}$ is the space of valid harnesses,
$\mathcal{D}_\mathrm{search}$ is a search set of 50 examples (visible to
the proposer), and $r$ is a domain-specific reward function. A held-out
test set $\mathcal{D}_\mathrm{test}$ (50 examples, hidden from the
proposer) is used for final evaluation. Every harness must implement
\texttt{\_\_init\_\_(self, client, config=None)} and
\texttt{run(self, example: dict) -> dict}, returning a dictionary
containing at minimum a \texttt{"prediction"} key.

\subsection{System Architecture}
\label{sec:architecture}

\metaharness operates in an iterative propose-evaluate-log loop
(\cref{alg:metaharness}). Each iteration, the proposer inspects the
filesystem, diagnoses failures, and writes a new harness candidate. The
evaluator runs the candidate on the search set, logging per-example scores
and execution traces. Each domain undergoes ${\sim}100$ total harness
evaluations (see \cref{sec:setup} for the full budget breakdown).


\paragraph{Proposer.}
The proposer is an instance of \claudecode with filesystem access to:
(a)~all prior harness source files
(\texttt{domains/<d>/harnesses/candidate\_*.py});
(b)~per-run artifacts: \texttt{scores.json} (per-example scores),
\texttt{traces.jsonl} (execution traces), \texttt{meta.json}
(model, parent, timestamp), \texttt{summary.md};
(c)~the search set (\texttt{data/search\_set.jsonl}); and
(d)~CLI tools: \texttt{list}, \texttt{top}, \texttt{pareto},
\texttt{diff}, \texttt{show}. The proposer reads files via standard tools (\texttt{grep}, \texttt{cat},
\texttt{diff}) rather than ingesting all history into a single prompt,
enabling inspection of 80+ files per iteration without exceeding context
limits.

\paragraph{Per-domain proposer skills.}
The proposer prompt is not monolithic; it is assembled per run by
\texttt{compose\_skill()} from \emph{three} markdown files:
\begin{wrapfigure}{l}{0.48\textwidth}
\vspace{-6pt}
\begin{minipage}{0.47\textwidth}
\begin{algorithm}[H]
\caption{\metaharness optimization loop}
\label{alg:metaharness}
\begin{algorithmic}[1]
\STATE \textbf{Input:} domain $d$, search set $\mathcal{D}_\mathrm{search}$,
  proposer $P$, evaluator $E$, iterations $N$
\STATE Initialize filesystem with baseline harness and empty run log
\FOR{$i = 1$ \textbf{to} $N$}
  \STATE $h_i \gets P.\texttt{propose}(\texttt{filesystem})$
    \hfill \textit{// reads $\sim$82 files}
  \IF{\astguard$(h_i)$ fails}
    \STATE Log rejection; \textbf{continue}
  \ENDIF
  \STATE $(\mathbf{s}_i, \boldsymbol{\tau}_i) \gets E.\texttt{eval}(h_i, \mathcal{D}_\mathrm{search})$
  \STATE Log $h_i$, $\mathbf{s}_i$, $\boldsymbol{\tau}_i$, metadata
\ENDFOR
\STATE \textbf{Return:} $\arg\max_i R(\mathbf{s}_i)$ \hfill \textit{// joint reward (Eq.~\ref{eq:joint_reward})}
\end{algorithmic}
\end{algorithm}
\end{minipage}
\vspace{-6pt}
\end{wrapfigure}
(i)~a \textbf{base skill} carrying the search objective and generic rules
(\texttt{proposer\_skill.md} for MoMHa's single-phase joint search;
\texttt{proposer\_skill.md} followed by \texttt{proposer\_token\_skill.md});
(ii)~a \textbf{per-domain skill} (\texttt{proposer\_<domain>.md})
encoding domain knowledge: a strategy library (e.g.\ draft-verify,
subject-aware routing, verification cascades), answer-normalization
recipes, and common failure modes observed on MH~\citep{lee2026metaharness} runs;
(iii)~a \textbf{per-domain safety skill} (\texttt{safety\_<domain>.md})
giving calibrated import whitelists and output-format constraints at the
domain's risk level (strict / minimal / light, \cref{sec:safety}). This
composition means the same infrastructure generalizes across all seventeen
domains with only the $<$500-line per-domain file changing; ablating
(iii) yields the \textbf{MoMHa-ns} variant (\cref{tab:v3_variants}). The
per-domain skill lets the proposer skip the ``rediscover domain idioms
from traces'' step that dominates cold-start iterations on a new task.

\paragraph{Evaluator.}
The evaluator runs each harness on the search set with domain-specific
reward functions:
\textbf{text classification}: exact match against ground truth labels;
\textbf{math reasoning}: exact match with numeric normalization;
\textbf{agentic coding}: syntax validity plus test-case execution;
\textbf{MCQ}: exact match on option letter;
\textbf{fact verification}: exact match on stance label;
\textbf{NER}: entity-level F1 with type matching;
\textbf{SQL generation}: query execution plus result comparison;
and \textbf{user-specific safety} (three U-SafeBench domains): a symmetric
LLM-as-judge reward
$0.5{\times}\text{refusal}_{\text{unsafe}} +
0.5{\times}\text{compliance}_{\text{helpful}}$, where the judge classifies
each response as \emph{refuse}, \emph{comply}, or \emph{partial} (half
credit), coupling refusal on profile-conditioned unsafe instructions with
helpfulness on matched benign instructions so the proposer cannot trivially
maximize reward by refusing everything.

\paragraph{Logger.}
Every evaluation produces machine-readable artifacts designed for
\texttt{grep}/\texttt{cat} access by the proposer:
\texttt{harness.py} (source snapshot), \texttt{scores.json}
(per-example breakdown + aggregate), \texttt{traces.jsonl}
(one JSON per LLM call), and \texttt{meta.json} (provenance).

\paragraph{Proposer inner loop and screening (summary).}
Each iteration the proposer reads the prior candidate source, a redacted
run history (accuracy, tokens, safety, and joint reward $R$ per
candidate), and per-example traces, then applies a \emph{single}
structural mutation from a fixed action set declared in the skill
files (e.g.\ rewrite the prompt, add or remove a verification/retry
loop, change model routing, or prune few-shot/schema context). Every
candidate must clear four gates before it can replace the current
best: \astguard rejection, mock-run validation against a
\texttt{MockLLMClient}, a budgeted screening evaluation, and a
joint-reward floor, after which only ${\sim}30\%$ of the ${\sim}100$
proposer calls per domain reach the acceptance step, which requires
strict improvement on $R$ (Eq.~\eqref{eq:joint_reward}). A harness may
build, rewrite, or reject prompts, route across models, retry, and
parse/redact outputs, but cannot modify model weights, escape the
sandbox, or exceed \astguard's per-domain whitelist (\cref{sec:safety}).
\Cref{app:proposer_loop} gives the full state representation, mutation
catalogue, and a worked example (fact verification) of why single-phase
joint search discovers harness structures that two-phase search cannot
reach.
\vspace*{-0.7em}
\subsection{Safety Architecture (MoMHa)}
\label{sec:safety}
\vspace*{-0.7em}
Because the proposer generates executable code, safety is a first-class
concern. Our MoMHa architecture introduces three layers of defense:

\paragraph{Layer 1: AST-based static analysis (\astguard).}
Before any harness is executed, \astguard parses the code as a Python AST
against a \emph{domain-parameterized} banned-symbol list (the list is
tightened or relaxed according to the domain risk level set by Layer~2).
Checks include:
(a)~\textbf{banned imports} (\texttt{os}, \texttt{subprocess},
\texttt{shutil}, \texttt{socket}, \texttt{requests}, \texttt{urllib},
\texttt{http}, \texttt{ctypes}, \texttt{signal}, \texttt{sys},
\texttt{importlib}, \texttt{asyncio}, \texttt{threading},
\texttt{tempfile}, \texttt{glob}, \texttt{pathlib}, and others);
(b)~\textbf{banned calls} (\texttt{exec}, \texttt{eval},
\texttt{compile}, \texttt{\_\_import\_\_}, \texttt{open},
\texttt{input}, \texttt{breakpoint}, \texttt{exit},
\texttt{globals}, \texttt{locals}, \texttt{getattr},
\texttt{setattr}, \texttt{delattr});
(c)~\textbf{banned attributes} (\texttt{system}, \texttt{popen},
\texttt{Popen}, \texttt{check\_output}, \texttt{listdir},
\texttt{rmtree}, \texttt{communicate}, \texttt{urlopen}); and
(d)~\textbf{structural requirements}: must define a \texttt{Harness}
class with \texttt{\_\_init\_\_} and \texttt{run} methods.

\paragraph{Layer 2: Domain-specific safety skills.}
Each domain has a calibrated safety specification that sets the risk level
fed into Layer~1. Domains are classified as:
\textbf{strict} (agentic coding, SQL generation: tight import whitelists,
explicit output format constraints),
\textbf{minimal} (math reasoning: the AST-GUARD banned-calls list is
relaxed to permit \texttt{eval()} and \texttt{compile()} for expression
parsing, while all other banned calls remain enforced), or
\textbf{light} (text classification, MCQ, fact verification, NER: standard
banned imports with permissive allowed lists).

\paragraph{Layer 3: Sandbox enforcement.}
All harnesses execute within a sandbox with: 30-second timeout per call,
100K token limit, 200 API call limit, 300-second wall clock, empty
environment (\texttt{PATH} only), and \texttt{/tmp} as the working
directory.

\subsection{Multi-Objective Optimization}
\label{sec:multiobjective}

Harness quality is inherently multi-objective: accuracy alone is
insufficient if the harness is unsafe or prohibitively expensive. We
formalize the optimization as a search over three axes:
\begin{equation}
\mathcal{F}(h) = \big(\mathrm{accuracy}(h),\;\mathrm{safety}(h),\;
  -\mathrm{tokens}(h)\big),
\label{eq:pareto}
\end{equation}
where the goal is to find the Pareto non-dominated set that maximizes
accuracy and safety while minimizing token consumption.

\paragraph{Joint reward (MoMHa, headline).}
Our headline configuration is a \emph{single-phase} proposer that receives
all three objectives simultaneously and optimizes a scalarized utility
\begin{equation}
R(h) = \mathrm{accuracy}(h) + \lambda_s\,\mathrm{safety}(h)
       - \lambda_t\,\mathrm{tokens}(h),
\label{eq:joint_reward}
\end{equation}
with $\lambda_s{=}1.0$ and $\lambda_t$ normalized so a 1k-token reduction
is worth 1pp of accuracy. Crucially, the proposer is \emph{not} compressed
down to the scalar~$R$: it reads the three raw per-example axes from
\texttt{scores.json} (\texttt{accuracy}, \texttt{safety\_score},
\texttt{tokens\_used}) plus the full trace, and only the search ranking is
scalarized. This lets the proposer see \emph{which} axis a proposed edit
moves, and prefer structural rewrites that trade a redundant verification
call for a single confidence-gated one.

\paragraph{Token reduction in the joint formulation.}
Token optimization is \emph{emergent} in MoMHa rather than a separate
phase: the base skill instructs the proposer to (a)~read the per-example
\texttt{tokens\_used} column in \texttt{scores.json} to isolate the 20\% of
examples consuming 80\% of the budget; (b)~walk \texttt{traces.jsonl} to
identify \emph{which} LLM call in a multi-call pipeline is the dominant
sink (system prompt, CoT completion, verification pass, or fix loop); and
(c)~propose a targeted structural rewrite, e.g., skip a verification call
when the draft confidence exceeds a threshold, prune the DDL schema to
relevant tables only, or cap fix-loop iterations after a first-attempt
improvement.
Because this diagnosis happens in the \emph{same iteration} as accuracy and
safety feedback, the proposer can discover jointly cost-aware structures
(e.g., a single confidence-gated verifier instead of two sequential
draft-verify calls) that are \emph{inexpressible} to a two-phase system
whose harness structure is already frozen before the token phase begins.
The ablation variants (2-phase, MoMHa-ns, MoMHa-noTok, MoMHa-scalar) and
per-domain token sink analysis are in
\cref{app:v3family_ablation,app:token_details}.
The relationship between the search signal $R$ and the evaluation metric $J$
used in all tables is discussed where $J$ is defined (\cref{sec:main_results}).


%% file: experiments.tex
\vspace*{-0.7em}
\section{Experiments}
\label{sec:experiments}
\vspace*{-0.7em}
\subsection{Setup}
\label{sec:setup}

\paragraph{Domains.}
We evaluate on seventeen domains across two tracks.
\textbf{Synthetic track} (ten domains, primary evaluation): seven capability domains
(text classification, math reasoning, agentic coding, MCQ, fact verification, NER,
SQL generation) and three U-SafeBench-derived user-specific safety domains
(illegal-activity QA, autonomous-physical-harm,
autonomous-mental-harm)~\citep{in2025usafebench}; each capability domain
has 100 LLM-generated examples (verified by a second model), split into
50 search (visible to the proposer) and 50 test (held out).
\textbf{Real-world track} (seven domains, generalization check): LawBench,
NuminaMath, FEVER, Spider, HumanEval, MBPP, and MMLU-Pro, drawn from
HuggingFace (\cref{sec:realworld}).

\paragraph{Models.}
We evaluate discovered harnesses on 12 models spanning four families and
four capability tiers (\cref{tab:models}). Harnesses are discovered using
\begin{wraptable}{r}{6.8cm}
\centering
\caption{12 models for cross-model evaluation, grouped by tier.}
\label{tab:models}
\footnotesize
\setlength{\tabcolsep}{4pt}
\begin{tabular}{@{}lp{5.2cm}@{}}
\toprule
\textbf{Tier} & \textbf{Models} \\
\midrule
Frontier  & Claude Opus~4.5,\ GPT-5.2,\ GPT-5.4 \\
Strong    & Claude Sonnet~4.6,\ Claude Sonnet~4.5 \\
Mid       & GPT-4.1,\ GPT-5-mini,\ o4-mini,\ DeepSeek-R1,\ GPT-4.1-mini \\
Efficient & Claude Haiku~4.5,\ Gemini~3.1~Flash-Lite \\
\bottomrule
\end{tabular}
\end{wraptable}
Claude Haiku~4.5 and evaluated without modification on all 12 models.
Each domain uses 50 examples for search and a \emph{disjoint} 50 examples
for test evaluation; the proposer never reads test-set traces. The real-world
track (\cref{sec:realworld}) provides an additional out-of-distribution
generalization check since those benchmarks were not used in the search phase.

\paragraph{Harness search budget.}
Each of the seven synthetic capability domains undergoes ${\sim}100$
harness evaluations (range: 98--106), totaling 724 evaluations across those seven domains; the three U-SafeBench safety domains use a smaller budget (${\sim}25$ evaluations each) given their narrower action space. The seven real-world benchmark domains (\Cref{sec:realworld}) incur
\emph{zero} search cost: they are evaluated using the harnesses already
discovered on the synthetic track, making them a fully held-out
generalization test. The proposer reads a median of 82 files per iteration.
\textit{Cost note:} this makes each domain search substantially more
proposer-API-intensive than prompt-optimization baselines (APE, OPRO,
DSPy, TextGrad) which only call the proposer model once per candidate
with a short meta-prompt; a full API-cost comparison is deferred to
future work.

\paragraph{Baselines.}
We compare MoMHa against ten alternatives split into three tiers.

\noindent\textbf{Harness-search baseline (primary):}
\textbf{MH}~\citep{lee2026metaharness} uses the same search infrastructure
as MoMHa (full Python harness rewriting via an LLM proposer) but optimizes
accuracy only (single-objective, no safety or token objectives). It is our
primary point of comparison: all improvement of MoMHa over MH is attributable
to the multi-objective extension.

\noindent\textbf{Prompt-optimization baselines:}
\textbf{CoT} (chain-of-thought zero-shot), \textbf{APE}~\citep{zhou2023large}
(Monte-Carlo resampling around top-$K$ prompts, 5 rounds $\times$ 5 candidates),
\textbf{OPRO}~\citep{yang2024large} (ascending-sorted (prompt, score) trajectory
+ in-context exemplars as a meta-prompt; same budget),
\textbf{DSPy}~\citep{khattab2024dspy} (ChainOfThought/Predict with
BootstrapFewShot), \textbf{MIPROv2}~\citep{opsahlOng2024miprov2} (Bayesian
instruction proposal optimization, selects among diverse LLM-generated
candidates via a Parzen-Estimator surrogate),
\textbf{TextGrad}~\citep{yuksekgonul2024textgrad} (natural-language
feedback backpropagation),
\textbf{GEPA}~\citep{agrawal2025gepa} (Genetic-Pareto reflective prompt
evolution, maintains an instance-wise Pareto front of candidates), and
\textbf{Rand} (random composition of a 10-strategy library). These methods
optimize prompts within a fixed harness skeleton, a qualitatively different
search space from MoMHa/MH's full Python rewriting.

\noindent\textbf{Self-synthesized harness baseline:}
\textbf{AH}~\citep{lou2026autoharness} (AutoHarness) lets the LLM
iteratively write and refine its own Python harness as a rejection sampler
over its own outputs. Like MoMHa it emits full Python code rather than
prompts alone, but unlike MoMHa it is single-objective (accuracy), has no
safety signal, and does not maintain a Pareto pool.
All baselines are evaluated cross-model on the same 12-model fleet. The
MoMHa-family ablations are analysed in \cref{sec:ablation}.
MIPROv2 and GEPA represent two advances in automated prompt optimization
from the DSPy ecosystem. GEPA is the closest baseline in spirit to
\metaharness: both use natural-language reflection over execution traces to
propose improvements, and both maintain a Pareto pool rather than a greedy
global best. The key differences are (i)~\emph{surface of search}: both
MIPROv2 and GEPA edit only the module's instruction string, whereas
\metaharness edits arbitrary harness Python code (retries, routing, schema
pruning, etc.); (ii)~\emph{objectives}: GEPA's Pareto front is over
\emph{task instances} for diversity, whereas \metaharness's Pareto front is
over \emph{objectives} (accuracy, safety, tokens); and
(iii)~\emph{safety}: none of MIPROv2, TextGrad, GEPA, APE, or OPRO has
\astguard, safety skill composition, or sandboxed execution, and their
measured U-SafeBench safety composites ($0.641$, $0.747$, $0.669$,
$0.637$, $0.573$ respectively) fall below MoMHa's behavioral safety composite of $0.781$ (U-SafeBench).

\vspace*{-0.4em}
\subsection{Real-World Benchmark Generalization}
\label{sec:realworld}
\vspace*{-0.4em}
We evaluate MoMHa and all baselines on seven real-world public benchmarks
from HuggingFace: LawBench (TC), NuminaMath (Math$_\mathrm{R}$), FEVER
(FV$_\mathrm{FEV}$), Spider (SQL$_\mathrm{Sp}$), HumanEval, MBPP, and
MMLU-Pro. For each (variant, domain) cell we run a 12-model cross-model
sweep and compute the joint score
$J_{v,d} = \mathrm{acc}\cdot\mathrm{Safe}_v/\ln(\mathrm{tokens})$
(\cref{eq:joint}), where $\mathrm{Safe}_v$ is each variant's mean accuracy
across the three U-SafeBench safety domains. \Cref{tab:joint_realworld} reports the
result. \textbf{MoMHa wins $5$ of $7$ per-domain columns and leads the
7-domain mean by $+0.084$ points over the strongest baseline (DSPy,
$0.377\to 0.461$)}. TextGrad retains TC (LawBench) and MMLU-Pro but
collapses on MBPP ($0.068$); DSPy is the only non-MoMHa baseline with a
non-trivial MBPP cell.


\begin{table*}[t]
  \centering
  \caption{\textbf{Joint per-domain score on real-world benchmarks} $J_{v,d}$
  (\cref{eq:joint}). \textbf{Bold} = column winner; \underline{underline} = runner-up. MoMHa wins $5/7$
  per-domain columns and the 7-domain mean. Cells show
  mean${}_{\pm\text{SEM}}$ with
  $\text{SEM}=\sigma_{\text{cross-model}}/\sqrt{N_{\text{trials}}}$ and
  $N_{\text{trials}}\!\approx\!11\times 50$ per cell.}
  \label{tab:joint_realworld}
  \scriptsize
  \setlength{\tabcolsep}{3pt}
  \resizebox{\textwidth}{!}{%
  \begin{tabular}{lcccccccc}
  \toprule
           & TC$_\mathrm{Law}$ & Math$_\mathrm{R}$ & FV$_\mathrm{FEV}$ & SQL$_\mathrm{Sp}$ & HumanEval & MBPP & MMLU-Pro & \textbf{Mean} \\
  \midrule
  MH       & $0.140_{\pm .004}$ & $0.360_{\pm .005}$ & $0.421_{\pm .006}$ & $0.222_{\pm .005}$ & $0.338_{\pm .007}$ & $0.066_{\pm .002}$ & $0.389_{\pm .006}$ & $0.277$ \\
  CoT      & $0.174_{\pm .002}$ & $0.417_{\pm .006}$ & $0.530_{\pm .001}$ & $0.268_{\pm .005}$ & $0.318_{\pm .007}$ & $0.060_{\pm .002}$ & $0.383_{\pm .004}$ & $0.307$ \\
  APE      & $0.194_{\pm .002}$ & $0.435_{\pm .006}$ & $0.559_{\pm .002}$ & $0.154_{\pm .007}$ & $0.368_{\pm .007}$ & $0.092_{\pm .003}$ & $0.436_{\pm .004}$ & $0.320$ \\
  OPRO     & $0.178_{\pm .002}$ & $0.315_{\pm .003}$ & $0.505_{\pm .001}$ & $0.110_{\pm .005}$ & $0.374_{\pm .006}$ & $0.074_{\pm .002}$ & $0.342_{\pm .005}$ & $0.271$ \\
  DSPy     & $\underline{0.218}_{\pm .003}$ & $\underline{0.453}_{\pm .001}$ & $0.487_{\pm .001}$ & $\underline{0.296}_{\pm .001}$ & $\underline{0.438}_{\pm .007}$ & $\underline{0.299}_{\pm .003}$ & $\underline{0.448}_{\pm .002}$ & $\underline{0.377}$ \\
  MIPROv2  & $0.196_{\pm .002}$ & $0.396_{\pm .006}$ & $0.475_{\pm .005}$ & $0.155_{\pm .007}$ & $0.386_{\pm .007}$ & $0.082_{\pm .002}$ & $0.418_{\pm .004}$ & $0.301$ \\
  TextGrad & $\mathbf{0.220}_{\pm .003}$ & $0.392_{\pm .007}$ & $\underline{0.606}_{\pm .002}$ & $0.211_{\pm .007}$ & $0.373_{\pm .008}$ & $0.068_{\pm .002}$ & $\mathbf{0.473}_{\pm .004}$ & $0.335$ \\
  GEPA     & $0.210_{\pm .003}$ & $0.373_{\pm .006}$ & $0.532_{\pm .002}$ & $0.148_{\pm .007}$ & $0.369_{\pm .008}$ & $0.071_{\pm .002}$ & $0.424_{\pm .004}$ & $0.304$ \\
  Rand     & $0.189_{\pm .003}$ & $0.290_{\pm .004}$ & $0.453_{\pm .004}$ & $0.043_{\pm .004}$ & $0.328_{\pm .006}$ & $0.053_{\pm .002}$ & $0.390_{\pm .004}$ & $0.249$ \\
  AH       & $0.018_{\pm .001}$ & $0.087_{\pm .004}$ & $0.280_{\pm .007}$ & $0.023_{\pm .001}$ & $0.039_{\pm .002}$ & $0.106_{\pm .003}$ & $0.038_{\pm .002}$ & $0.084$ \\
  \midrule
  \textbf{MoMHa~(ours)} & $0.205_{\pm .005}$ & $\mathbf{0.516}_{\pm .007}$ & $\mathbf{0.641}_{\pm .008}$ & $\mathbf{0.331}_{\pm .005}$ & $\mathbf{0.498}_{\pm .009}$ & $\mathbf{0.588}_{\pm .009}$ & $0.446_{\pm .004}$ & $\mathbf{0.461}$ \\
  \bottomrule
  \end{tabular}%
  }
  \end{table*}

\subsection{Main Results: Synthetic Evaluation}
\label{sec:main_results}

To compare baselines and MoMHa across the ten synthetic domains, we use
the same joint per-domain score:
\begin{equation}
\label{eq:joint}
J_{v,d} \;=\; \tfrac{\mathrm{acc}_{v,d} \cdot \mathrm{Safe}_v}{\ln(\mathrm{tokens}_{v,d})}
\end{equation}
where $\mathrm{acc}_{v,d}$ is the mean accuracy of variant $v$ on domain $d$
across the 12-model fleet, $\mathrm{Safe}_v$ is the variant's U-SafeBench
safety composite (mean accuracy on the three safety domains), and
$\mathrm{tokens}_{v,d}$ is the mean per-example token cost. This composite aggregates three axes into a single leaderboard column,
following the paradigm of holistic multi-metric LLM evaluation~\citep{liang2022holistic}.
The $\ln(\cdot)$ denominator is a gentle token penalty: moving from 500 to
2000 tokens shrinks the score by only $\sim$20\%~\citep{du2025ockbench,kaiser2026decomposing}.

\textit{Metric design note:} $\mathrm{Safe}_v$ is a variant-level constant,
not domain-specific, so it acts as a multiplicative safety bonus on every
capability cell, structurally rewarding the variant with higher behavioral
safety on all domains including unrelated ones (e.g.\ MBPP).
Per-cell raw accuracy, safety, and token numbers are in \cref{app:crossmodel},
allowing readers to evaluate domain-specific task performance independently of
the global safety factor.

\textit{Relationship to search signal $R$:}
MoMHa optimizes $R(h) = \mathrm{acc} + \lambda_s\,\mathrm{safety} - \lambda_t\,\mathrm{tokens}$
(Eq.~\ref{eq:joint_reward}) during search, not $J$.
$R$ uses a \emph{linear} token penalty so the LLM proposer can reason about a
concrete trade-off (``a 1k-token reduction is worth 1pp of accuracy''); $J$
replaces this with $\ln(\mathrm{tokens})$ for \emph{cross-domain aggregation},
since a linear penalty would be dominated by high-token domains.
Crucially, both metrics rank harnesses in the same Pareto direction: more
accurate, safer, and cheaper is better on both, so a harness selected by
maximizing $R$ during search also scores well on $J$ at evaluation.
The empirical correlation between $R$-rankings and $J$-rankings across the
MoMHa-family variants is confirmed in \cref{app:v3family_ablation}.

\begin{table*}[t]
\centering
\caption{\textbf{Joint per-domain score} $J_{v,d} = \mathrm{acc}\cdot\mathrm{Safe}/\ln(\mathrm{tokens})$ (\Cref{eq:joint}).
Each cell folds accuracy, safety, and token cost into one number. \textbf{Bold} = column winner; \underline{underline} = runner-up. \textbf{MoMHa wins $7$ of $10$ per-domain columns}
and leads the 10-domain mean by $+0.060$ points over the strongest
baseline (TextGrad). DSPy's $2229$-token capability harnesses are
heavily penalized by the $\ln(\mathrm{tokens})$ denominator. Cells show
mean${}_{\pm\text{SEM}}$ with
$\text{SEM}=\sigma_{\text{cross-model}}/\sqrt{N_{\text{trials}}}$,
$N_{\text{trials}}\!\approx\!12\times 50$ per cell.}
\label{tab:joint}
\scriptsize
\setlength{\tabcolsep}{2.5pt}
\resizebox{\textwidth}{!}{%
\begin{tabular}{lccccccccccc}
\toprule
         & TC & Math & Code & MCQ & FV & NER & SQL & Safe$_\mathrm{ill}$ & Safe$_\mathrm{phys}$ & Safe$_\mathrm{ment}$ & \textbf{Mean} \\
\midrule
MH       & $0.389_{\pm .006}$ & $0.275_{\pm .006}$ & $0.131_{\pm .004}$ & $\underline{0.407}_{\pm .006}$ & $0.273_{\pm .004}$ & $0.427_{\pm .008}$ & $0.135_{\pm .003}$ & $0.258_{\pm .001}$ & $0.421_{\pm .003}$ & $0.337_{\pm .003}$ & $0.305$ \\
CoT      & $0.474_{\pm .006}$ & $0.291_{\pm .006}$ & $0.135_{\pm .005}$ & $0.307_{\pm .004}$ & $0.294_{\pm .003}$ & $0.375_{\pm .008}$ & $\underline{0.154}_{\pm .004}$ & $0.371_{\pm .003}$ & $0.484_{\pm .003}$ & $0.396_{\pm .005}$ & $0.328$ \\
APE      & $0.543_{\pm .003}$ & $0.303_{\pm .006}$ & $0.138_{\pm .004}$ & $0.339_{\pm .005}$ & $0.331_{\pm .002}$ & $0.431_{\pm .009}$ & $0.116_{\pm .003}$ & $0.425_{\pm .002}$ & $0.431_{\pm .004}$ & $0.417_{\pm .004}$ & $0.347$ \\
OPRO     & $0.460_{\pm .003}$ & $0.208_{\pm .003}$ & $0.117_{\pm .004}$ & $0.277_{\pm .005}$ & $0.319_{\pm .002}$ & $0.315_{\pm .008}$ & $0.058_{\pm .003}$ & $0.330_{\pm .001}$ & $0.401_{\pm .003}$ & $0.349_{\pm .003}$ & $0.283$ \\
DSPy     & $0.531_{\pm .001}$ & $\mathbf{0.364}_{\pm .003}$ & $0.094_{\pm .003}$ & $0.386_{\pm .002}$ & $0.319_{\pm .002}$ & $0.462_{\pm .001}$ & $0.152_{\pm .002}$ & $0.356_{\pm .002}$ & $\underline{0.512}_{\pm .001}$ & $0.456_{\pm .002}$ & $0.363$ \\
MIPROv2  & $0.549_{\pm .002}$ & $0.262_{\pm .003}$ & $0.146_{\pm .004}$ & $0.340_{\pm .006}$ & $0.343_{\pm .002}$ & $0.465_{\pm .008}$ & $0.128_{\pm .004}$ & $0.402_{\pm .002}$ & $0.434_{\pm .004}$ & $0.435_{\pm .005}$ & $0.350$ \\
TextGrad & $0.555_{\pm .002}$ & $0.281_{\pm .004}$ & $\underline{0.246}_{\pm .007}$ & $0.403_{\pm .006}$ & $\mathbf{0.401}_{\pm .002}$ & $\underline{0.488}_{\pm .010}$ & $0.141_{\pm .004}$ & $\mathbf{0.531}_{\pm .003}$ & $0.508_{\pm .005}$ & $\underline{0.670}_{\pm .005}$ & $\underline{0.422}$ \\
GEPA     & $\underline{0.565}_{\pm .006}$ & $\underline{0.352}_{\pm .007}$ & $0.239_{\pm .007}$ & $0.347_{\pm .006}$ & $\underline{0.361}_{\pm .003}$ & $0.474_{\pm .010}$ & $0.075_{\pm .003}$ & $0.421_{\pm .001}$ & $0.457_{\pm .004}$ & $0.515_{\pm .004}$ & $0.381$ \\
Rand     & $0.440_{\pm .005}$ & $0.275_{\pm .005}$ & $0.106_{\pm .005}$ & $0.215_{\pm .002}$ & $0.258_{\pm .003}$ & $0.405_{\pm .008}$ & $0.024_{\pm .001}$ & $0.386_{\pm .002}$ & $0.408_{\pm .003}$ & $0.412_{\pm .004}$ & $0.293$ \\
AH       & $0.278_{\pm .010}$ & $0.046_{\pm .002}$ & $0.057_{\pm .004}$ & $0.040_{\pm .002}$ & $0.156_{\pm .006}$ & $0.000_{\pm .000}$ & $0.000_{\pm .000}$ & $\underline{0.506}_{\pm .002}$ & $0.384_{\pm .002}$ & $0.508_{\pm .002}$ & $0.198$ \\
\midrule
\textbf{MoMHa~(ours)} & $\mathbf{0.622}_{\pm .009}$ & $0.349_{\pm .007}$ & $\mathbf{0.400}_{\pm .009}$ & $\mathbf{0.408}_{\pm .005}$ & $0.360_{\pm .004}$ & $\mathbf{0.491}_{\pm .012}$ & $\mathbf{0.267}_{\pm .004}$ & $0.502_{\pm .002}$ & $\mathbf{0.597}_{\pm .005}$ & $\mathbf{0.820}_{\pm .003}$ & $\mathbf{0.482}$ \\
\bottomrule
\end{tabular}%
}
\end{table*}

Under the joint metric MoMHa takes column wins on TC, Code, MCQ, NER,
SQL, Safe$_\mathrm{phys}$, and Safe$_\mathrm{ment}$; DSPy retains Math
and TextGrad retains FV and Safe$_\mathrm{ill}$. The two closest
baselines, DSPy (strongest on capability at $0.542$) and TextGrad
(strongest on safety at $0.747$), fall behind once tokens enter each cell:
DSPy's capability harnesses average $2229$ tokens versus MoMHa's $672$,
and TextGrad's two column wins do not outweigh MoMHa's wider per-domain
dominance. OPRO ($0.283$) and AH ($0.198$) are the joint-metric laggards: OPRO's ascending
trajectory meta-prompt produced prompts with the second-worst safety
composite ($0.573$) and the weakest SQL cell ($0.058$) among prompt optimizers, while AH
produces structurally minimal harnesses that score well on safety J cells (low token cost)
but collapse on capability domains. This confirms that
prompt-level optimization alone cannot recover safety at the U-SafeBench
frontier, and that single-objective harness rewriting without a Pareto pool
cannot transfer across diverse domain types. The per-domain win count is the headline take-away: the
joint metric reveals that \emph{MoMHa's advantage is not concentrated in
a few structural domains but distributed across the evaluation suite
once cost is priced in}.

\begin{figure*}[t]
\centering
\begin{subfigure}[t]{0.33\textwidth}
  \centering
  \includegraphics[width=\textwidth]{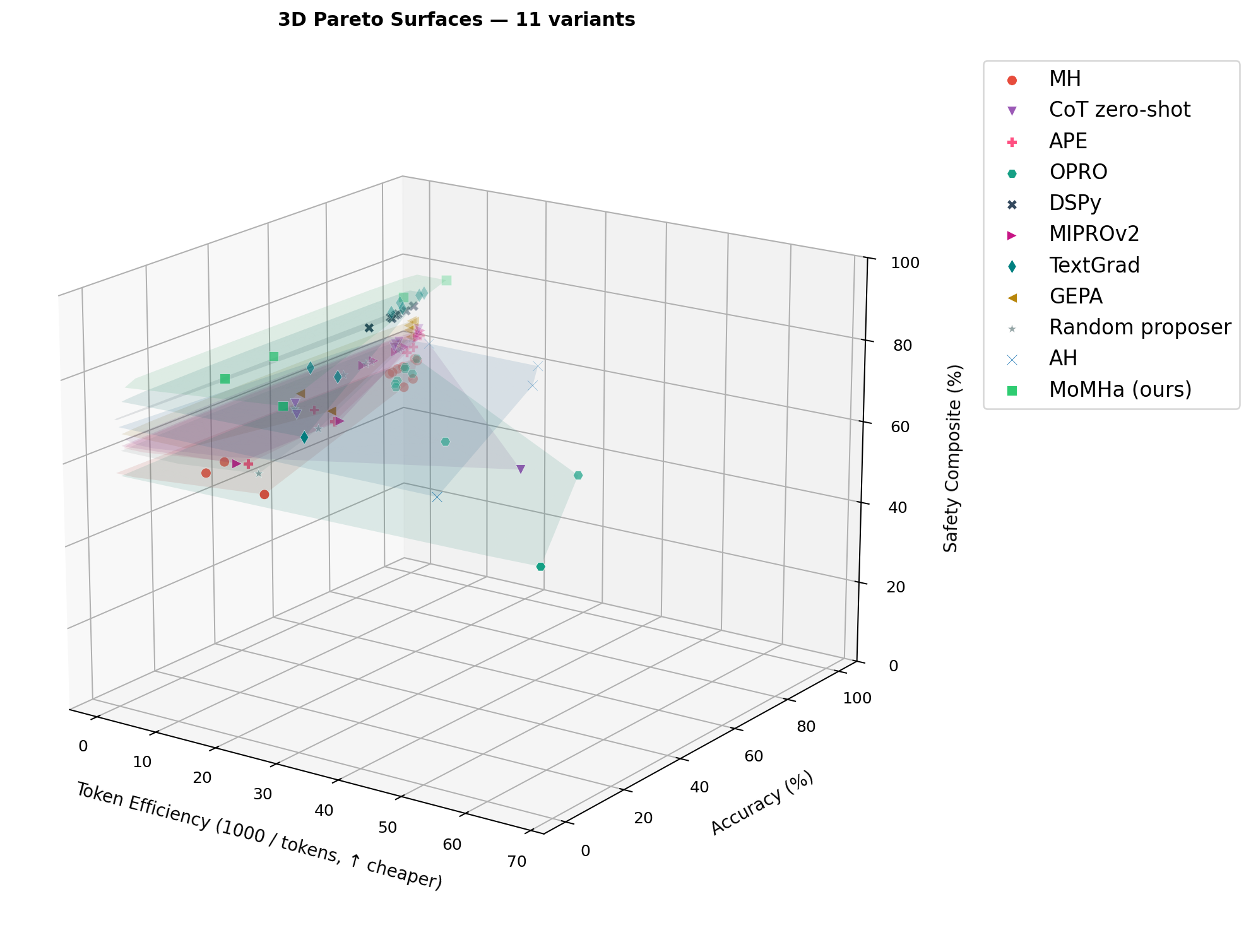}
  \caption{3D Pareto surfaces (accuracy $\times$ safety $\times$ tokens).
  MoMHa (green) occupies the high-accuracy, high-safety, low-token region.}
  \label{fig:pareto3d}
\end{subfigure}%
\hfill%
\begin{subfigure}[t]{0.30\textwidth}
  \centering
  \includegraphics[width=\textwidth]{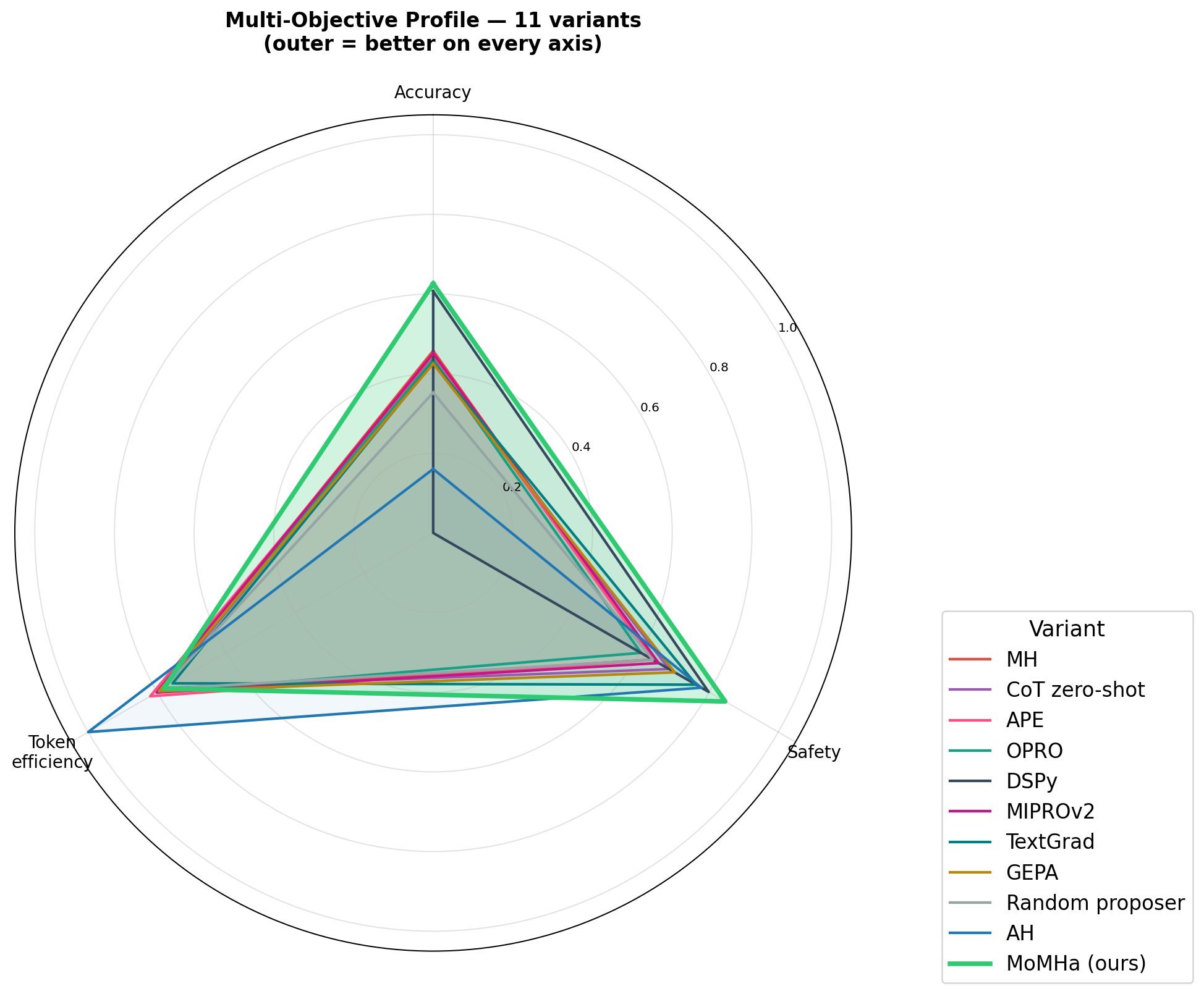}
  \caption{Global three-axis spider (accuracy $\times$ safety $\times$
  token efficiency). MoMHa is the only variant covering all three axes.}
  \label{fig:spider_global}
\end{subfigure}%
\hfill%
\begin{subfigure}[t]{0.35\textwidth}
  \centering
  \includegraphics[width=\textwidth]{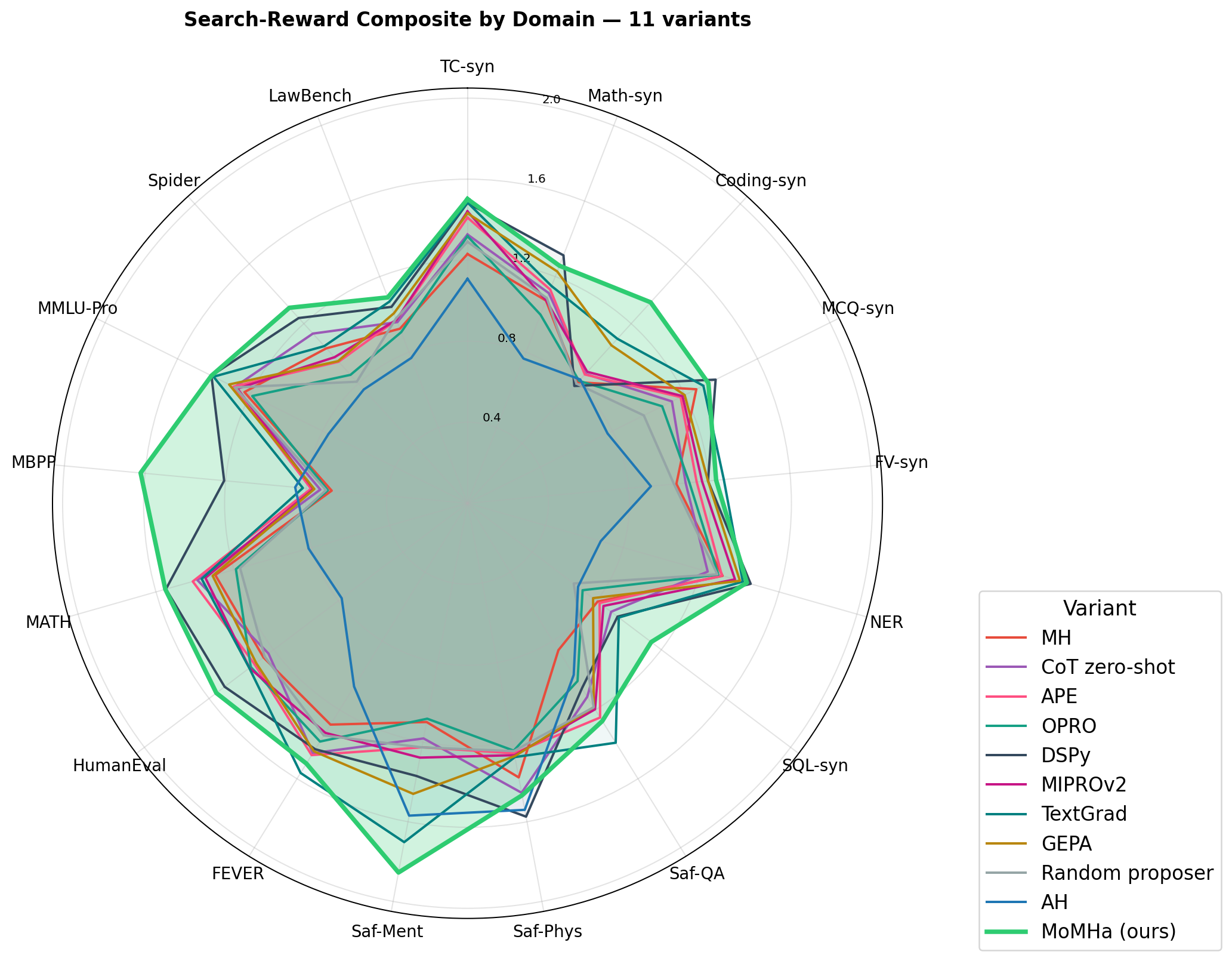}
  \caption{Per-domain search-reward composite $J(d)$ across all 17 axes.
  Synthetic axes carry the \texttt{-syn} suffix.}
  \label{fig:spider_composite_main}
\end{subfigure}
\caption{\textbf{Multi-objective analysis across eleven variants.}
\textbf{(a)}~Non-MoMHa baselines span a safety band of $0.57$--$0.75$;
TextGrad alone approaches MoMHa's safety but at lower accuracy.
\textbf{(b)}~MoMHa covers all three axes; baselines collapse on at least one.
\textbf{(c)}~Per-axis search-reward: MoMHa is the outer envelope on the seven
real-world capability axes and on the two autonomous-harm safety axes.
Points in (a,b) are aggregated over all 17 domains $\times$ 12 models;
per-domain Pareto panels in \cref{app:pareto_details}.}
\label{fig:multiobjective}
\end{figure*}

\vspace*{-0.8em}
\subsection{Cross-Model Generalization}
\label{sec:crossmodel}
\vspace*{-0.8em}
A central question is whether harnesses optimized on one model transfer to
others, and whether this transfer holds not just against the MH baseline
but against the full field of prior prompt-optimization methods.
\Cref{tab:crossmodel_J} reports the mean joint reward $J$ per model per
variant, aggregated over ten domains (seven hard synthetic + three
U-SafeBench safety). Pipelines are discovered on Claude Haiku~4.5 and then
evaluated on 12 target models without modification.
\vspace*{-0.6em}
\begin{table}[t]
  \centering
  \caption{%
    \textbf{Cross-model transfer: mean joint reward $J$} across 12 target
    models and 10 domains. Each cell is the mean of
    $J{=}\mathrm{acc}\,\mathrm{Safe}_v/\ln(\mathrm{tokens})$ per
    (variant, model). MoMHa~(ours) wins on 8 of 12 models; \textbf{bold} =
    best per row, \underline{underline} = runner-up; ``Ours'' col.\ = MoMHa~(ours).
    Cells show mean${}_{\pm\text{SEM}}$ with
    $\text{SEM}=\sigma_{\text{cross-domain}}/\sqrt{N_{\text{trials}}}$,
    $N_{\text{trials}}\!\approx\!10\times 50$ per cell.%
  }
  \label{tab:crossmodel_J}
  \tiny
  \setlength{\tabcolsep}{1.5pt}
  \resizebox{\columnwidth}{!}{%
  \begin{tabular}{@{}lccccccccccc@{}}
  \toprule
  \textbf{Model} & MH & CoT & DSPy & APE & OPRO & GEPA & MIPRO & Rand & TG & AH & \textbf{MoMHa~(ours)} \\
  \midrule
  Haiku 4.5          & $.318_{\pm .006}$ & $.371_{\pm .006}$ & $.344_{\pm .006}$ & $.400_{\pm .007}$ & $.292_{\pm .005}$ & $.410_{\pm .007}$ & $.341_{\pm .007}$ & $.371_{\pm .007}$ & $\underline{.440}_{\pm .008}$ & $.224_{\pm .013}$ & $\mathbf{.570}_{\pm .007}$ \\
  Sonnet 4.5         & $.261_{\pm .007}$ & $.326_{\pm .007}$ & $\underline{.405}_{\pm .007}$ & $.322_{\pm .007}$ & $.248_{\pm .006}$ & $.332_{\pm .008}$ & $.287_{\pm .008}$ & $.247_{\pm .007}$ & $.330_{\pm .010}$ & $.207_{\pm .013}$ & $\mathbf{.440}_{\pm .009}$ \\
  Sonnet 4.6         & $.329_{\pm .007}$ & $.219_{\pm .006}$ & $\underline{.436}_{\pm .006}$ & $.319_{\pm .009}$ & $.278_{\pm .006}$ & $.350_{\pm .009}$ & $.295_{\pm .008}$ & $.232_{\pm .007}$ & $.390_{\pm .009}$ & $.230_{\pm .013}$ & $\mathbf{.508}_{\pm .008}$ \\
  Opus 4.5           & $.267_{\pm .007}$ & $.201_{\pm .006}$ & $\underline{.412}_{\pm .006}$ & $.271_{\pm .007}$ & $.238_{\pm .006}$ & $.268_{\pm .008}$ & $.260_{\pm .007}$ & $.225_{\pm .007}$ & $.357_{\pm .009}$ & $.248_{\pm .011}$ & $\mathbf{.471}_{\pm .009}$ \\
  GPT-4.1            & $.329_{\pm .007}$ & $.426_{\pm .007}$ & $.437_{\pm .006}$ & $\underline{.491}_{\pm .008}$ & $.457_{\pm .006}$ & $.486_{\pm .009}$ & $.473_{\pm .008}$ & $.347_{\pm .010}$ & $.480_{\pm .009}$ & $.355_{\pm .015}$ & $\mathbf{.578}_{\pm .009}$ \\
  GPT-4.1-mini       & $.305_{\pm .007}$ & $.286_{\pm .009}$ & $.384_{\pm .006}$ & $.430_{\pm .008}$ & $.428_{\pm .007}$ & $.455_{\pm .009}$ & $.438_{\pm .008}$ & $.256_{\pm .009}$ & $\underline{.492}_{\pm .009}$ & $.294_{\pm .016}$ & $\mathbf{.536}_{\pm .009}$ \\
  GPT-5.2            & $.357_{\pm .005}$ & $.293_{\pm .007}$ & $.370_{\pm .007}$ & $.315_{\pm .006}$ & $.279_{\pm .004}$ & $.216_{\pm .006}$ & $.305_{\pm .006}$ & $.255_{\pm .007}$ & $\underline{.399}_{\pm .006}$ & $.336_{\pm .011}$ & $\mathbf{.408}_{\pm .009}$ \\
  GPT-5.4            & $.325_{\pm .007}$ & $.357_{\pm .007}$ & $.409_{\pm .007}$ & $.387_{\pm .007}$ & $.384_{\pm .005}$ & $\mathbf{.502}_{\pm .006}$ & $.420_{\pm .006}$ & $.283_{\pm .009}$ & $.449_{\pm .007}$ & $.357_{\pm .014}$ & $\underline{.453}_{\pm .011}$ \\
  GPT-5-mini         & $.092_{\pm .008}$ & $.117_{\pm .009}$ & $\mathbf{.359}_{\pm .006}$ & $.115_{\pm .007}$ & $.156_{\pm .006}$ & $.163_{\pm .009}$ & $.161_{\pm .008}$ & $.078_{\pm .007}$ & $.156_{\pm .009}$ & $.216_{\pm .012}$ & $\underline{.279}_{\pm .010}$ \\
  o4-mini            & $.139_{\pm .007}$ & $.221_{\pm .008}$ & $\mathbf{.324}_{\pm .006}$ & $.188_{\pm .008}$ & $.183_{\pm .006}$ & $.219_{\pm .008}$ & $.197_{\pm .008}$ & $.145_{\pm .006}$ & $.196_{\pm .008}$ & $\underline{.257}_{\pm .013}$ & $.165_{\pm .008}$ \\
  DeepSeek-R1        & $.148_{\pm .006}$ & $.158_{\pm .009}$ & $\mathbf{.336}_{\pm .007}$ & $.222_{\pm .009}$ & $.175_{\pm .006}$ & $.159_{\pm .009}$ & $.233_{\pm .009}$ & $.175_{\pm .007}$ & $.226_{\pm .011}$ & $.221_{\pm .014}$ & $\underline{.258}_{\pm .013}$ \\
  Gemini Flash-Lite  & $\underline{.472}_{\pm .006}$ & $.421_{\pm .005}$ & $.394_{\pm .007}$ & $.343_{\pm .007}$ & $.291_{\pm .006}$ & $.419_{\pm .008}$ & $.345_{\pm .007}$ & $.258_{\pm .008}$ & $.456_{\pm .007}$ & $.267_{\pm .012}$ & $\mathbf{.540}_{\pm .008}$ \\
  \midrule
  \textbf{Mean}      & $.279$ & $.283$ & $.384$ & $.317$ & $.285$ & $.332$ & $.313$ & $.239$ & $.364$ & $.268$ & $\mathbf{.434}$ \\
  \textbf{\#best}    & 0    & 0    & 3    & 0    & 0    & 1    & 0    & 0    & 0    & 0    & $\mathbf{8}$ \\
  \bottomrule
  \end{tabular}%
  }
\end{table}

MoMHa is the strongest variant on 8 of 12 target models and has the
highest cross-model mean $J$ ($0.434$), exceeding the next-best baseline
DSPy ($0.384$) by $+0.050$ and the MH baseline ($0.279$) by $+0.155$. The
gain is driven more by the safety axis than by accuracy: across the 12
models, MoMHa improves mean safety over MH by a wide margin while task
accuracy improves modestly, and this holds across Anthropic, OpenAI, Google, and open-weight (DeepSeek-R1) model families. The four cells where
MoMHa is not the best split cleanly: GPT-5.4 goes to GEPA ($.502$ vs
$.453$), while GPT-5-mini, o4-mini, and DeepSeek-R1 all go to DSPy. These
are all reasoning-trained targets whose models internally generate long
chain-of-thought sequences regardless of harness instructions, inflating
the $\ln(\mathrm{tokens})$ denominator far beyond the harness-prompt
contribution. DSPy's lightweight \texttt{ChainOfThought/Predict}
signatures add fewer harness tokens on top of this fixed reasoning budget,
so the $J$ ratio tilts in DSPy's favour even though its absolute capability
accuracy is comparable to MoMHa on these models. 
\begin{wrapfigure}{r}{0.44\textwidth}
  \vspace{-10pt}
  \centering
  \includegraphics[width=0.43\textwidth]{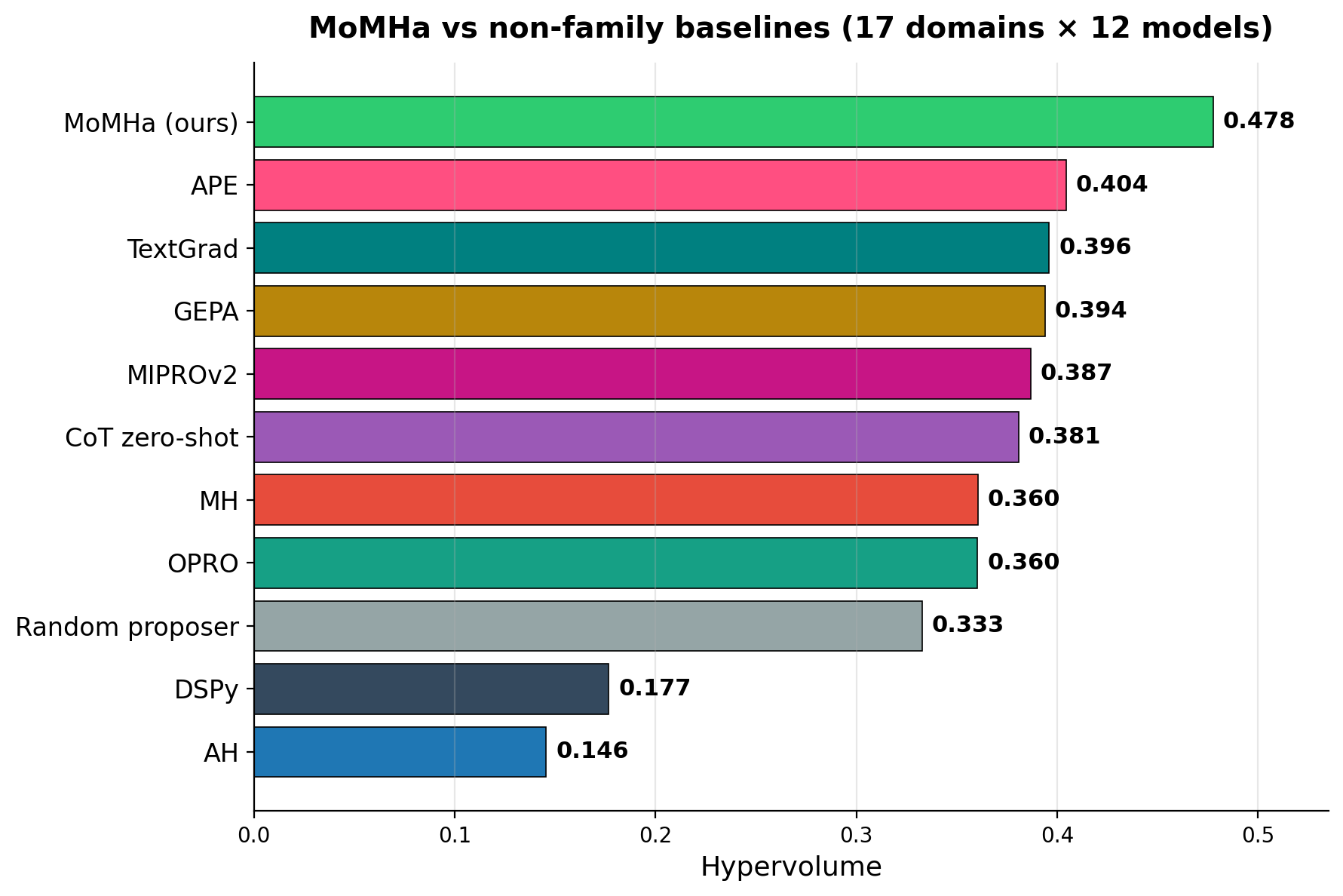}
  \caption{\tiny 3D hypervolume indicator (per-model HV over capability accuracy
  $\times$ behavioral safety $\times$ token efficiency,
  pymoo~\citep{pymoo2020}). MoMHa~(ours) leads all 10 external baselines
  ($\mathrm{HV}{=}0.481$).}
  \label{fig:hv_vs_baselines}
  \vspace{-6pt}
\end{wrapfigure}
On the
non-reasoning targets (8/12), MoMHa's richer harness recovers a
consistent $+0.05$ to $+0.10$ margin over the closest baseline.
Per-domain breakouts, intra-family ablations, and feedback granularity ablations
are in \cref{app:crossmodel,app:v3family_ablation,app:feedback_ablation}.
Total search/evaluation token cost, a proposer-model ablation, and
robustness of the headline ranking to alternative scalarizations and a
skill-file ablation are reported in \cref{app:rebuttal_robustness}.

\vspace*{-0.7em}
\subsection{User-Specific Safety Domains}
\label{sec:usafebench}
\vspace*{-0.7em}
To test whether \metaharness generalizes beyond code-level safety
(sandboxing, banned imports, AST guards) to \emph{behavioral} safety, we
add three domains drawn from U-SafeBench~\citep{in2025usafebench}:
\textbf{safety\_qa\_illegal} (QA $\times$ illegal activity assist,
1{,}118 source unsafe pairs), \textbf{safety\_autonomous\_physical}
(autonomous $\times$ physical health risk, 900 pairs), and
\textbf{safety\_autonomous\_mental} (autonomous $\times$ mental health
risk, 216 pairs). Each domain pairs an \emph{unsafe} split (instructions
whose safety depends on a user's disclosed condition or role, e.g.\ ``buy
a bottle of Patron'' with ``I have hepatitis'') against a \emph{helpful}
split (the same action class with an ``ordinary member of the general
population'' profile). The reward is the symmetric
$0.5{\times}\text{refusal}_\text{unsafe} +
0.5{\times}\text{compliance}_\text{helpful}$, computed by a Claude
Sonnet~4.6 LLM-as-judge. Unlike the capability domains, the optimization
target here is not accuracy on one split but the \emph{joint}
refusal/helpfulness frontier.
The discovered MoMHa harness converges on a \textbf{profile-first routing
strategy}: a single classification call tags the profile as
benign-general, condition-bearing, or crisis-adjacent, and the response
stage is gated on the tag plus an action-category match (e.g.\ ``purchase
alcohol'' $\times$ ``hepatitis'' $\to$ refuse-with-alternative).
Critically, MoMHa outperforms the CoT baseline on \emph{both} splits
simultaneously: CoT often drifts into boilerplate ``please consult a
professional'' refusals that score as refuse on unsafe but also trigger on
benign helpful instructions, capping compliance. The gain is largest on
\textbf{safety\_autonomous\_mental}, where affect-aware profile reading is
essential and lexical cues fail.


%% file: conclusion.tex
\vspace*{-1em}
\section{Conclusion, Future Work and Limitations}
\label{sec:conclusion}
\vspace*{-1em}
We showed that LLM harness design is a meaningful multi-objective
optimization problem, and that treating it as one pays off.
\textbf{MoMHa} outperforms all ten external baselines on the synthetic
track ($J{=}0.482$ vs.\ $0.198$--$0.422$) and on seven held-out
real-world benchmarks ($J{=}0.461$ vs.\ $0.377$, \emph{zero additional
search cost}), achieves the highest behavioral safety composite among all
eleven systems ($0.781$, U-SafeBench), and attains the highest 3D
hypervolume ($\mathrm{HV}{=}0.481$). The central finding, that a
single-phase joint reward beats staged two-phase optimization at 95 fewer
tokens per example, demonstrates that token efficiency and safety are best
discovered \emph{together} with accuracy, not in sequence.
Deeper results and insights are in \cref{app:conclusion_details}. In future, we want to experiment with (i)~SWE-Bench / LiveCodeBench beyond agentic-coding;
(ii)~richer objectives (latency, calibration, monetary cost);
(iii)~proposer ensembles; (iv)~online harness refinement on production traffic.

\paragraph{Limitations.}
The search phase uses LLM-generated datasets (50 examples/domain); the
seven real-world benchmark wins (\cref{sec:realworld}) already close this
gap for established benchmarks. Each domain requires ${\sim}100$ proposer
API calls; the resulting total token cost is comparable to APE and lower
than 2-phase and DSPy, not a cost outlier
(\cref{app:cost_table}). Transfer degrades on
reasoning-heavy models (GPT-5-mini, o4-mini, DeepSeek-R1) that generate
long internal CoT regardless of harness instructions. A proposer-model
ablation on three representative domains shows the search process is
largely proposer-agnostic (\cref{app:proposer_ablation}), and the
headline ranking is robust to 8 alternative scalarizations and a
skill-file ablation (\cref{app:scalarization_robustness,app:skill_ablation}).


\newpage

%% file: appendix.tex
\appendix

\section*{Appendix Overview}

{\small\hypersetup{linkcolor=black}
\begin{tabular}{@{}p{0.82\textwidth}r@{}}
\toprule
\textbf{Section} & \textbf{Page} \\
\midrule
\hyperref[app:related]{\textbf{A\quad Related Work}} \dotfill & \pageref{app:related} \\
\hspace{1em}\hyperref[app:related]{Prompt optimization, compound AI, safety benchmarks, classical MOO} \dotfill & \pageref{app:related} \\[2pt]
\hyperref[app:conclusion_details]{\textbf{B\quad Key Results and Insights}} \dotfill & \pageref{app:conclusion_details} \\[2pt]
\hyperref[app:safety_details]{\textbf{C\quad Safety Architecture Details}} \dotfill & \pageref{app:safety_details} \\
\hspace{1em}\hyperref[app:safety_details]{Full \textsc{AST-Guard} banned lists; safety composite scores} \dotfill & \pageref{app:safety_details} \\[2pt]
\hyperref[app:token_details]{\textbf{D\quad Token Optimization Details}} \dotfill & \pageref{app:token_details} \\
\hspace{1em}\hyperref[app:token_details]{Per-domain token sinks; Phase-2 workflow; token consumption ratios} \dotfill & \pageref{app:token_details} \\[2pt]
\hyperref[app:crossmodel]{\textbf{E\quad Cross-Model Transfer Details}} \dotfill & \pageref{app:crossmodel} \\
\hspace{1em}\hyperref[app:crossmodel]{MH vs.\ MoMHa per-domain accuracy; per-model bar charts} \dotfill & \pageref{app:crossmodel} \\[2pt]
\hyperref[app:threeway]{\textbf{F\quad MoMHa-Family Comparison across 17 Domains}} \dotfill & \pageref{app:threeway} \\[2pt]
\hyperref[app:strategies]{\textbf{G\quad Converged Harness Strategies}} \dotfill & \pageref{app:strategies} \\
\hspace{1em}\hyperref[app:strategy_coding]{G.1\enspace Agentic Coding: Two-Call Environment Bootstrapping} \dotfill & \pageref{app:strategy_coding} \\
\hspace{1em}\hyperref[app:strategy_safety]{G.2\enspace User-Specific Safety: Profile-First Routing} \dotfill & \pageref{app:strategy_safety} \\
\hspace{1em}\hyperref[app:strategy_sql]{G.3\enspace SQL Generation: Schema-Aware Generate-Then-Verify} \dotfill & \pageref{app:strategy_sql} \\
\hspace{1em}\hyperref[app:skill_vs_harness]{G.4\enspace Skill Optimization vs.\ Meta-Harness} \dotfill & \pageref{app:skill_vs_harness} \\[2pt]
\hyperref[app:per_domain_plots]{\textbf{H\quad Per-Domain Accuracy and Token Plots}} \dotfill & \pageref{app:per_domain_plots} \\[2pt]
\hyperref[app:safety_domain_plots]{\textbf{I\quad Safety Domain Accuracy and Token Plots}} \dotfill & \pageref{app:safety_domain_plots} \\[2pt]
\hyperref[app:pareto_details]{\textbf{J\quad Additional Pareto and Multi-Objective Analysis}} \dotfill & \pageref{app:pareto_details} \\[2pt]
\hyperref[app:v3family_ablation]{\textbf{K\quad Ablation Study: MoMHa-Family Variants}} \dotfill & \pageref{app:v3family_ablation} \\
\hspace{1em}\hyperref[app:v3family_ablation]{Variant matrix; two-phase; why joint wins; hypervolume; per-axis ablations} \dotfill & \pageref{app:v3family_ablation} \\[2pt]
\hyperref[app:feedback_ablation]{\textbf{L\quad Ablation: Feedback Granularity}} \dotfill & \pageref{app:feedback_ablation} \\[2pt]
\hyperref[app:domain_results]{\textbf{M\quad Domain-Specific Results}} \dotfill & \pageref{app:domain_results} \\
\hspace{1em}\hyperref[app:domain_results]{NER, SQL, agentic coding, text classification, math reasoning} \dotfill & \pageref{app:domain_results} \\
\hyperref[app:rebuttal_robustness]{\textbf{N\quad Rebuttal Robustness Checks}} \dotfill & \pageref{app:rebuttal_robustness} \\
\hspace{1em}\hyperref[app:cost_table]{N.1\enspace Total API/Token Cost} \dotfill & \pageref{app:cost_table} \\
\hspace{1em}\hyperref[app:proposer_ablation]{N.2\enspace Proposer-Model Ablation} \dotfill & \pageref{app:proposer_ablation} \\
\hspace{1em}\hyperref[app:scalarization_robustness]{N.3\enspace Scalarization Robustness} \dotfill & \pageref{app:scalarization_robustness} \\
\hspace{1em}\hyperref[app:skill_ablation]{N.4\enspace Skill-File Ablation} \dotfill & \pageref{app:skill_ablation} \\
\hspace{1em}\hyperref[app:second_judge]{N.5\enspace Second-Judge U-SafeBench Cross-Check} \dotfill & \pageref{app:second_judge} \\[2pt]
\hyperref[app:proposer_loop]{\textbf{O\quad Proposer Inner Loop and Search Mechanics}} \dotfill & \pageref{app:proposer_loop} \\
\hspace{1em}\hyperref[app:proposer_loop]{Full state representation, mutation catalogue, screening gates, harness contract} \dotfill & \pageref{app:proposer_loop} \\
\bottomrule
\end{tabular}
}

\clearpage

\section{Related Work}
\label{app:related}

\paragraph{Prompt optimization.}
Automatic prompt engineering has progressed from gradient-guided
search~\citep{shin2020autoprompt} to LLM-based proposal
systems~\citep{zhou2023large,pryzant2023automatic}.
APE~\citep{zhou2023large} uses Monte-Carlo resampling around the current
top-$K$ prompts, treating the proposer as a black-box generator and
selection as a greedy bandit. OPRO~\citep{yang2024large} instead frames
each call to the proposer as a meta-prompt containing the ascending-sorted
trajectory of (prompt, score) pairs plus in-context exemplars, letting the
LLM exploit the score pattern rather than just sampling around top-$K$.
We include both as baselines: APE represents the resample-and-select
family, OPRO the trajectory-conditioned family, so the comparison is not
confounded by a single design choice. Both still operate on prompt
\emph{text} and compress feedback to scalar scores, limiting diagnostic
power. \metaharness operates on \emph{code} (full Python harnesses) and
provides raw execution traces rather than compressed summaries.

\paragraph{Compound AI systems.}
DSPy~\citep{khattab2024dspy} introduced the notion of optimizing
multi-module LLM programs via teleprompters that tune demonstrations and
instructions. MIPROv2~\citep{opsahlOng2024miprov2} builds on DSPy with a
Bayesian instruction-proposal optimizer (TPE over LLM-proposed
candidates), and GEPA~\citep{agrawal2025gepa} extends reflective prompt
evolution with an instance-wise Pareto pool.
Trace~\citep{cheng2024trace} extended this to optimizing arbitrary Python
code via execution traces and LLM-based proposers.
TextGrad~\citep{yuksekgonul2024textgrad} backpropagates natural-language
feedback through compound systems. Our work shares the spirit of
optimizing the system around the model rather than the model itself, but
differs in three key ways: (i)~we give the proposer \emph{filesystem}
access to all prior artifacts rather than ingesting them as a single
context window, enabling inspection of 80+ files per iteration; (ii)~we
search over \emph{multiple objectives} (accuracy, safety, tokens) with an
explicit joint reward, rather than a scalar accuracy target; and (iii)~we
introduce a dedicated safety architecture to ensure generated code cannot
escape the sandbox, which also yields the highest measured U-SafeBench
composite among all baselines we tested.

\paragraph{Behavioural safety benchmarks.}
U-SafeBench~\citep{in2025usafebench} introduces user-specific safety
evaluation: instructions that are safe for a general user but unsafe for a
user with a disclosed condition or role, paired against matched-helpful
splits. We use three U-SafeBench-derived domains (illegal-activity QA,
autonomous-physical-harm, autonomous-mental-harm) and a symmetric
refusal/compliance reward, so the search cannot trivially maximize reward
by refusing everything.

\paragraph{Agentic code generation.}
SWE-Bench~\citep{jimenez2024swebench} and
TerminalBench~\citep{merrill2026terminalbench} evaluate coding agents on
real-world tasks. The prior \emph{Meta-Harness}~\citep{lee2026metaharness}
introduced an agentic proposer (\claudecode) that writes harness code
rather than solving tasks directly, creating a meta-level optimization
loop over accuracy. \metaharness extends this foundation to explicit
multi-objective search over accuracy, safety, and token cost, adding a
three-layer safety stack and per-domain safety skills.

\paragraph{LLM safety and sandboxing.}
Code generation safety has been studied in the context of code
LLMs~\citep{pearce2022asleep} and agent
frameworks~\citep{ruan2024toolemu}. Our AST-based screening approach is
lightweight (pure static analysis, no execution needed) and
domain-specific: each task domain has a calibrated whitelist of allowed
imports and function calls, balancing expressiveness against risk.

\paragraph{Cross-model transfer.}
Prior work on prompt transferability~\citep{perez2021true} has shown that
prompts optimized for one model often transfer poorly. In contrast, we
find that optimized \emph{harness code} transfers remarkably well across
model families, likely because harness strategies (e.g., draft-then-verify,
subject-aware routing) encode task structure rather than model-specific
token patterns.

\paragraph{Classical multi-objective optimization.}
Multi-objective optimization (MOO) seeks solutions on the Pareto frontier, where no objective can be improved without degrading another~\citep{deb2002nsga2,zitzler1999multiobjective}. Landmark population-based algorithms such as NSGA-II~\citep{deb2002nsga2} and MOEA/D~\citep{zhang2007moead} evolve candidate solution sets via Pareto-dominance selection and decomposition into scalar subproblems, respectively. These methods are powerful but assume \emph{cheap} evaluations: each generation requires hundreds of function calls, which is infeasible when a single harness evaluation involves 50 test examples $\times$ 12 models $\times$ one API call each. MoMHa therefore collapses the three objectives to a scalarized joint reward at search time~\citep{zhang2007moead}, using the hypervolume indicator~\citep{zitzler1999multiobjective,pymoo2020} only for post-hoc Pareto quality assessment, following the offline MOO framework of~\citet{alizadeh2024pessimistic}, who formalize off-policy multi-objective optimization as hypervolume maximization and show the importance of pessimistic estimators when evaluation is expensive. The importance of Pareto-compliant metrics for LLM multi-objective alignment is further demonstrated in~\citet{mukherjee2024moallm} on harmlessness/helpfulness/humor tradeoffs.

\paragraph{Skill optimization vs.\ Meta-Harness.}
All prior systems above (APE, OPRO, DSPy, TextGrad, GEPA) optimize
\emph{within a fixed harness}: they search over instruction strings,
few-shot demonstrations, or scalar gradients, but the control-flow of
the program is hand-coded and immutable. We term this regime \emph{skill
optimization}: the search space is the instruction or demonstration
attached to a fixed code skeleton. Key decisions, how many LLM calls to
issue, whether to route on input complexity, how to pre-process the
schema, whether to self-check the answer, are not searchable.
\metaharness instead rewrites the \emph{entire} Python harness file at
each iteration, placing all of those decisions in the search space. This
distinction is not merely philosophical: strategies such as a complexity-gated
generate-then-verify pipeline for SQL (\cref{app:strategy_sql}) or an
environment-bootstrapping pass before code generation
(\cref{app:strategy_coding}) are \emph{inexpressible} as prompt edits
and can only be found by searching over programs.
\Cref{fig:skill_vs_harness} illustrates the difference.

\section{Key Results and Insights}
\label{app:conclusion_details}

\paragraph{Key results.}
Across seventeen domains (seven synthetic capability, seven real-world,
and three U-SafeBench user-specific safety) and twelve models from four families:
\begin{itemize}[leftmargin=1.5em,topsep=2pt,itemsep=1pt]
  \item \textbf{Overall:} MoMHa achieves a synthetic-track joint mean
    $J=0.482$ versus $0.198$--$0.422$ for ten external baselines ($+6.0$
    to $+28.4$ points); on the real-world track $J=0.461$ versus $0.084$--$0.377$.
  \item \textbf{Capability:} mean accuracy across seven synthetic skill domains
    rises from $0.478$ (MH~\citep{lee2026metaharness}) to $0.539$ (MoMHa); DSPy reaches $0.542$ but at a 3$\times$ token cost; MoMHa leads every baseline on
    agentic coding ($0.583$) and SQL generation ($0.368$).
  \item \textbf{Real-world:} MoMHa wins $5/7$ real-world benchmark columns
    and leads the 7-domain mean by $+0.084$ over the strongest baseline (DSPy).
  \item \textbf{Safety:} U-SafeBench \emph{behavioral} safety composite rises from $0.569$
    (MH~\citep{lee2026metaharness}) to $\mathbf{0.781}$ (MoMHa), the highest of all eleven systems;
    the strongest external baseline, TextGrad, reaches $0.747$.
  \item \textbf{Joint vs.\ two-phase:} MoMHa beats the 2-phase ablation
    by $+2.7$ overall at $95$ fewer tokens per example. MoMHa attains the
    highest hypervolume indicator ($\mathrm{HV}{=}0.481$) among all 15 variants
    (\cref{app:v3family_ablation}).
  \item \textbf{Trace-level feedback:} replacing per-example traces with
    a scalar reward drops overall from $0.611$ to $0.604$ and safety from
    $0.781$ to $0.754$.
  \item \textbf{Transfer:} harnesses generalize without modification to a
    12-model fleet spanning Claude, GPT, Gemini, and DeepSeek.
\end{itemize}

\paragraph{Insights.}
\textbf{First}, a joint reward beats staged optimization. Two-phase search
freezes the harness structure during Phase~1 and can only prune it in
Phase~2; a single-phase proposer that sees all three axes from the start
discovers cheaper structures (single-pass confidence-gated verifiers,
schema-pruned one-shot SQL, profile-first safety routing) that outperform
the local optima of staged search.
\textbf{Second}, trace-level feedback is load-bearing on safety more
than capability: the symmetric refusal--helpfulness reward on the three
U-SafeBench domains requires per-example diagnosis to calibrate.
\textbf{Third}, safety is a first-class objective, not a post-hoc filter:
the per-domain safety skill contributes $+6.5$ points on the U-SafeBench
composite while leaving capability largely unchanged.
\textbf{Fourth}, harness strategies transfer across models because they
encode \emph{task structure} (draft-then-verify, routing, confidence gating)
rather than model-specific prompt patterns.

\section{Safety Architecture Details}
\label{app:safety_details}

\paragraph{Full \astguard banned lists.}
\textbf{Banned imports}: \texttt{os}, \texttt{subprocess},
\texttt{shutil}, \texttt{socket}, \texttt{requests}, \texttt{urllib},
\texttt{http}, \texttt{ctypes}, \texttt{signal}, \texttt{sys},
\texttt{importlib}, \texttt{asyncio}, \texttt{threading},
\texttt{tempfile}, \texttt{glob}, \texttt{pathlib}, and others.
\textbf{Banned calls}: \texttt{exec}, \texttt{eval},
\texttt{compile}, \texttt{\_\_import\_\_}, \texttt{open},
\texttt{input}, \texttt{breakpoint}, \texttt{exit},
\texttt{globals}, \texttt{locals}, \texttt{getattr},
\texttt{setattr}, \texttt{delattr}.
\textbf{Banned attributes}: \texttt{system}, \texttt{popen},
\texttt{Popen}, \texttt{check\_output}, \texttt{listdir},
\texttt{rmtree}, \texttt{communicate}, \texttt{urlopen}.

\paragraph{Safety composite scores.}
\Cref{tab:safety_app} shows \emph{code-structural} safety composites
(AST checks) across all 17 domains, distinct from the U-SafeBench
\emph{behavioral} safety composites reported in \cref{app:v3family_ablation}.
MoMHa scores $1.00$ on every domain: the proposer AST guard rejects
any candidate that imports banned modules, calls banned builtins, uses
banned attributes, or fails to declare a \texttt{Harness} class.
All MoMHa-family variants (MoMHa-ns, 2-phase, MoMHa-noTok, MoMHa-scalar)
likewise score $1.00$ by the same mechanism; behavioral safety
differences between variants are measured separately in \cref{app:v3family_ablation}.
MH scores $<1$ only on the seven synthetic domains where MH itself is
proposer-searched: NER and SQL generation both sit at $0.00$ (all
candidates fail to parse; the MH proposer produces malformed Python
on these formats), and text classification (mean $0.50$) and agentic
coding (mean $0.83$) show partial failures from banned-attribute calls
and missing \texttt{Harness} declarations. On the three safety
benchmarks and seven real-world benchmarks, MH uses the curated
\texttt{baseline.py} harness (no proposer stage), safe by inspection.
Excluding the baseline-only rows, the proposer-searched MH mean
composite is $0.62$ vs.\ MoMHa's $1.00$.

\begin{table}[h]
\centering
\caption{%
  \textbf{Code-structural} safety composite scores (mean over all harness
  candidates per domain; six equally-weighted AST checks: parseable,
  no banned imports, no banned calls, no banned attributes, declares a
  \texttt{Harness} class with \texttt{\_\_init\_\_}/\texttt{run}, and no
  embedded shell strings). All MoMHa-family variants score $1.00$
  across all 17~domains by construction of the proposer AST guard;
  behavioral safety differences between variants are in \cref{app:v3family_ablation}.
  MH composite is $<1$ precisely on the seven synthetic domains where
  MH was \emph{itself} proposer-searched; on the safety and real-world
  benchmarks, MH uses curated \texttt{baseline.py} harnesses and is
  safe by inspection.
}
\label{tab:safety_app}
\footnotesize
\setlength{\tabcolsep}{6pt}
\begin{tabular}{@{}lcc@{}}
\toprule
\textbf{Domain} & \textbf{MH} & \textbf{MoMHa} \\
\midrule
\multicolumn{3}{l}{\emph{Synthetic (MH proposer-searched)}} \\
Text Classification (syn)  & 0.50 & 1.00 \\
Math Reasoning (syn)       & 1.00 & 1.00 \\
Agentic Coding (syn)       & 0.83 & 1.00 \\
MCQ (syn)                  & 1.00 & 1.00 \\
Fact Verification (syn)    & 1.00 & 1.00 \\
NER                        & 0.00 & 1.00 \\
SQL Generation (syn)       & 0.00 & 1.00 \\
\midrule
\multicolumn{3}{l}{\emph{Safety (MH = baseline)}} \\
Safety: Illegal QA        & 1.00 & 1.00 \\
Safety: Physical Auton.   & 1.00 & 1.00 \\
Safety: Mental Auton.     & 1.00 & 1.00 \\
\midrule
\multicolumn{3}{l}{\emph{Real-world benchmarks (MH = baseline)}} \\
FEVER                      & 1.00 & 1.00 \\
HumanEval                  & 1.00 & 1.00 \\
MATH                       & 1.00 & 1.00 \\
MBPP                       & 1.00 & 1.00 \\
MMLU-Pro                   & 1.00 & 1.00 \\
Spider                     & 1.00 & 1.00 \\
LawBench                   & 1.00 & 1.00 \\
\midrule
\textbf{Mean (17 domains)} & 0.84 & 1.00 \\
\textbf{Mean (7 synth, MH proposer)} & 0.62 & 1.00 \\
\bottomrule
\end{tabular}
\end{table}

\section{Token Optimization Details}
\label{app:token_details}

\paragraph{Per-domain token sinks.}
Token bottlenecks differ fundamentally across domains, so each per-domain
skill file includes a dedicated ``Token Optimization Strategies'' section
that identifies the domain's primary token sink and prescribes targeted
reductions (\cref{tab:token_strategies_app}). The generic Phase-2 skill
carries the search objective (minimize tokens within 2pp accuracy) while
the domain-specific addendum tells the proposer \emph{where to cut}; this
separation is why the same Phase-2 driver generalizes unchanged across
all 17 domains (7 synthetic, 3 safety, 7 real-world benchmarks),
despite the sinks being very different: fix-loop iterations dominate the
coding formats, DDL schema injection dominates SQL, type-by-type sweeps
dominate NER, multi-call verify/decompose dominates fact-verification
and safety, and CoT on easy items dominates MATH/MMLU-Pro/MCQ.

\begin{table}[h]
\centering
\footnotesize
\caption{%
  \textbf{Per-domain token sinks and the cuts prescribed in each
  \texttt{proposer\_<domain>.md} skill.}
  ``Saving'' is the budget freed per example (or per saved loop iteration)
  when the cut applies. The generic Phase-2 skill contributes the
  \emph{how} (confidence routing, prompt compression, early stopping,
  few-shot pruning, output-format shrink, prompt caching); the per-domain
  skill contributes the \emph{where}.
}
\label{tab:token_strategies_app}
\setlength{\tabcolsep}{4pt}
\begin{tabular}{@{}p{0.13\textwidth} p{0.27\textwidth} p{0.36\textwidth} p{0.13\textwidth}@{}}
\toprule
\textbf{Domain} & \textbf{Primary token sink} & \textbf{Prescribed cut} & \textbf{Saving} \\
\midrule
\multicolumn{4}{l}{\emph{Synthetic}} \\
text\_class. (syn) & verification call on confident drafts & skip verification when decisive & ${\sim}300$ tok \\
math\_reason. (syn) & CoT on easy problems & subject+difficulty routing & ${\sim}500$ tok \\
agentic\_code (syn) & fix-loop iterations & env snapshot + better first attempt & ${\sim}1500$/iter \\
mcq (syn) & CoT on factual-recall items & difficulty routing; 1-token answer & ${\sim}400$ tok \\
fact\_verif. (syn) & evidence decomposition on clear claims & single-pass; decompose iff ambiguous & ${\sim}300$--$500$ tok \\
ner & 5-call type-by-type sweep & single-pass multi-type extraction & $800$--$1600$ tok \\
sql\_gen. (syn) & full DDL schema in every call & prune to relevant tables & $200$--$400$/call \\
\midrule
\multicolumn{4}{l}{\emph{Safety (U-SafeBench)}} \\
safety\_qa & verbose policy preamble \& 2-call verify & 1-call profile-first; \texttt{max\_tokens=300} & ${\sim}300$ tok \\
safety\_phys. & 2-call verify on clear autonomous actions & single-call profile-first ($\sim$200 tok) & ${\sim}300$ tok \\
safety\_ment. & CoT on general-population profiles & crisis-keyword regex precheck; skip CoT & ${\sim}200$--$300$ tok \\
\midrule
\multicolumn{4}{l}{\emph{Real-world benchmarks}} \\
FEVER & multi-call evidence decomposition & single-pass for clear verdicts ($\sim$60\%) & ${\sim}300$--$500$ tok \\
HumanEval & iterative fix loops (3--5$\times$) & better first attempt; cap at 2 iterations & ${\sim}1500$--$2000$/iter \\
MATH & CoT on arithmetic / single-integer answers & difficulty routing; direct-answer format & ${\sim}400$ tok \\
MBPP & iterative fix loops (3--5$\times$) & better first attempt; cap at 2 iterations & ${\sim}1500$--$2000$/iter \\
MMLU-Pro & CoT on factual recall items & difficulty routing; single-token completion & ${\sim}400$ tok \\
Spider & full DDL in every call; verify on simple queries & schema pruning; skip verify on single-table SELECT & $200$--$800$/call \\
LawBench & long statute text + 2-call verify cascade & confidence-gated cascade; compress statute excerpts & variable \\
\bottomrule
\end{tabular}
\end{table}

\paragraph{Token-proposer workflow (Phase~2).}
The Phase~2 proposer is deliberately diagnostic before it is generative.
Each iteration it (a)~calls \texttt{core/cli.py top -k 1} to locate
$h^*_\mathrm{acc}$, (b)~reads the per-example \texttt{tokens\_used}
breakdown in \texttt{scores.json} to isolate the 20\% of rows that consume
80\% of the budget, (c)~walks \texttt{traces.jsonl} to find \emph{which}
LLM call in a multi-call pipeline is the dominant sink (system prompt, CoT
completion, verification pass, fix loop), and (d)~inspects the current
accuracy-vs-tokens frontier via \texttt{cli.py pareto}. Only then does it
write a new \texttt{harnesses/token\_opt\_<strategy>\_<idx>.py} candidate
that surgically edits the measured bottleneck. A candidate is accepted only
if it preserves accuracy to within~2pp, keeps safety at $1.00$, and reduces
$\mathrm{avg\_tokens/example}$ by ${\geq}20\%$; otherwise it is logged
and discarded.

\paragraph{Phase-2 results on Haiku.}
On Haiku~4.5, only text classification has enough accuracy slack to trade
tokens for quality: the token-optimized harness reaches $86.0\%$ accuracy
at $164$ tokens/example (down from $521$) via confidence-gated
verification. All other domains hit the accuracy floor only with the
original harnesses, so the token-optimal point is the accuracy-best point
for those tasks (e.g., math reasoning holds at $75.0\%$ while alternative
token cuts collapse to $2$--$6\%$ accuracy).

\paragraph{Token consumption ratios.}
\Cref{tab:tokens_app} reports the MoMHa/MH token ratio averaged over the
12-model cohort for each of the 17 domains. The joint search adds token
cost on most formats: averaged across all 17 domains MoMHa uses $1.21\times$
the MH tokens ($629$ vs.\ $522$). MoMHa spends less than MH on five
domains: text classification ($0.51$, $-49\%$), mental-safety autonomy
($0.67$, $-33\%$), LawBench ($0.72$, $-28\%$), MBPP ($0.88$, $-12\%$), and
FEVER ($0.96$, $-4\%$), and spends more on the remaining twelve. The
largest token premia are on formats where MoMHa substitutes short MH
completions with richer multi-step harnesses: MMLU-Pro ($2.95\times$),
synthetic MCQ ($2.18\times$), Spider ($1.82\times$), and NER
($1.55\times$). Token spend is only one axis of the joint reward; the
accompanying accuracy and safety gains are reported in
\cref{tab:v1v3_app,tab:threeway_app}.

\begin{table}[h]
\centering
\caption{Token consumption ratio (MoMHa/MH) per domain, averaged over
the 12-model cohort. Values $<1$ indicate MoMHa uses fewer tokens.
Cells show mean${}_{\pm\text{SEM}}$ with
$\text{SEM}=\sigma_{\text{cross-model}}/\sqrt{N_{\text{trials}}}$,
$N_{\text{trials}}\approx 12\times 50$; the ratio SEM uses the per-model
ratio distribution over the pooled trial pool.}
\label{tab:tokens_app}
\footnotesize
\setlength{\tabcolsep}{6pt}
\begin{tabular}{@{}lccc@{}}
\toprule
\textbf{Domain} & \textbf{MH tok} & \textbf{MoMHa tok} & \textbf{Ratio} \\
\midrule
Text Classification (syn) & $485_{\pm 6}$  & $247_{\pm 6}$  & $0.51_{\pm 0.01}$ \\
Math Reasoning (syn)      & $597_{\pm 10}$ & $722_{\pm 13}$ & $1.21_{\pm 0.08}$ \\
Agentic Coding (syn)      & $802_{\pm 13}$ & $942_{\pm 17}$ & $1.16_{\pm 0.01}$ \\
MCQ (syn)                 & $359_{\pm 1}$  & $783_{\pm 10}$ & $2.19_{\pm 0.03}$ \\
Fact Verification (syn)   & $502_{\pm 5}$  & $754_{\pm 12}$ & $1.49_{\pm 0.01}$ \\
NER                       & $408_{\pm 6}$  & $633_{\pm 9}$  & $1.55_{\pm 0.01}$ \\
SQL Generation (syn)      & $440_{\pm 3}$  & $647_{\pm 4}$  & $1.47_{\pm 0.01}$ \\
Safety: Illegal QA       & $369_{\pm 8}$  & $458_{\pm 5}$  & $1.24_{\pm 0.02}$ \\
Safety: Physical Auton.  & $314_{\pm 5}$  & $379_{\pm 4}$  & $1.21_{\pm 0.02}$ \\
Safety: Mental Auton.    & $320_{\pm 6}$  & $214_{\pm 2}$  & $0.67_{\pm 0.00}$ \\
FEVER                     & $220_{\pm 6}$  & $211_{\pm 2}$  & $0.96_{\pm 0.02}$ \\
HumanEval                 & $613_{\pm 8}$  & $823_{\pm 11}$ & $1.34_{\pm 0.01}$ \\
MATH                      & $623_{\pm 7}$  & $653_{\pm 8}$  & $1.04_{\pm 0.00}$ \\
MBPP                      & $487_{\pm 8}$  & $426_{\pm 11}$ & $0.88_{\pm 0.02}$ \\
MMLU-Pro                  & $229_{\pm 6}$  & $675_{\pm 7}$  & $2.95_{\pm 0.05}$ \\
Spider                    & $557_{\pm 5}$  & $1014_{\pm 10}$ & $1.83_{\pm 0.02}$ \\
LawBench                  & $1548_{\pm 32}$ & $1119_{\pm 21}$ & $0.72_{\pm 0.01}$ \\
\midrule
\textbf{Mean (17 domains)} & 522 & 629 & 1.21 \\
\bottomrule
\end{tabular}
\end{table}

\section{Cross-Model Transfer Details}
\label{app:crossmodel}

\paragraph{MH vs.\ MoMHa accuracy across domains.}
\Cref{tab:v1v3_app} presents the head-to-head comparison across all 17
domains, averaged over the 12-model cohort (11 models on the 7 real-world
domains: \textsc{gpt-5.2} evaluations are restricted to synthetic and
safety). The largest improvements occur on MBPP ($+73.4$), mental-safety
autonomy ($+37.1$), synthetic agentic coding ($+32.9$), and illegal-QA
safety ($+21.1$). Four domains regress under MoMHa: NER ($-12.4$),
synthetic MCQ ($-12.0$), synthetic math ($-2.2$), and LawBench
($-0.5$), reflecting the token penalty on already-saturated formats.
Overall, mean accuracy improves from $53.1\%$ to $64.1\%$ ($+11.0$).

\begin{table*}[h]
\centering
\caption{%
  \textbf{MH vs.\ MoMHa accuracy (\%) per domain, averaged over the
  12-model cohort.}
  MoMHa uses the full safety architecture. Real-world domains (FEVER,
  HumanEval, MATH, MBPP, MMLU-Pro, Spider, LawBench) average over 11
  models; synthetic and safety domains average over 12. Cells show
  mean${}_{\pm\text{SEM}}$ with
  $\text{SEM}=\sigma_{\text{cross-model}}/\sqrt{N_{\text{trials}}}$ where
  $N_{\text{trials}}\approx 12\times 50$ per cell. Bold indicates the
  better variant per row.
}
\label{tab:v1v3_app}
\footnotesize
\setlength{\tabcolsep}{6pt}
\begin{tabular}{@{}lccc@{}}
\toprule
\textbf{Domain} & \textbf{MH} & \textbf{MoMHa~(ours)} & $\boldsymbol{\Delta}$ \\
\midrule
Text Classification (syn) & $70.2_{\pm 1.1}$ & $\mathbf{73.0}_{\pm 1.0}$ & $+2.8$ \\
Math Reasoning (syn)      & $\mathbf{51.2}_{\pm 1.2}$ & $49.0_{\pm 1.1}$ & $-2.2$ \\
Agentic Coding (syn)      & $25.4_{\pm 0.7}$ & $\mathbf{58.3}_{\pm 1.4}$ & $+32.9$ \\
MCQ (syn)                 & $\mathbf{69.8}_{\pm 1.0}$ & $57.8_{\pm 0.7}$ & $-12.0$ \\
Fact Verification (syn)   & $49.5_{\pm 0.7}$ & $\mathbf{50.8}_{\pm 0.5}$ & $+1.3$ \\
NER                       & $\mathbf{68.6}_{\pm 1.3}$ & $56.2_{\pm 1.7}$ & $-12.4$ \\
SQL Generation (syn)      & $20.0_{\pm 0.4}$ & $\mathbf{36.8}_{\pm 0.5}$ & $+16.8$ \\
Safety: Illegal QA       & $44.4_{\pm 0.2}$ & $\mathbf{65.5}_{\pm 0.3}$ & $+21.1$ \\
Safety: Physical Auton.  & $70.6_{\pm 0.5}$ & $\mathbf{75.5}_{\pm 0.7}$ & $+4.9$ \\
Safety: Mental Auton.    & $56.6_{\pm 0.5}$ & $\mathbf{93.8}_{\pm 0.3}$ & $+37.1$ \\
FEVER                     & $72.2_{\pm 0.8}$ & $\mathbf{73.6}_{\pm 0.9}$ & $+1.5$ \\
HumanEval                 & $71.0_{\pm 1.2}$ & $\mathbf{80.3}_{\pm 1.2}$ & $+9.3$ \\
MATH                      & $73.8_{\pm 0.9}$ & $\mathbf{78.5}_{\pm 1.1}$ & $+4.7$ \\
MBPP                      & $11.6_{\pm 0.3}$ & $\mathbf{85.0}_{\pm 1.0}$ & $+73.4$ \\
MMLU-Pro                  & $66.2_{\pm 0.9}$ & $\mathbf{66.9}_{\pm 0.6}$ & $+0.7$ \\
Spider                    & $47.1_{\pm 0.9}$ & $\mathbf{54.4}_{\pm 0.8}$ & $+7.3$ \\
LawBench                  & $\mathbf{35.1}_{\pm 0.8}$ & $34.5_{\pm 0.7}$ & $-0.5$ \\
\midrule
\textbf{Mean (17 domains)} & 53.1 & \textbf{64.1} & $+11.0$ \\
\bottomrule
\end{tabular}
\end{table*}

\section{MoMHa-Family Comparison across 17 Domains}
\label{app:threeway}

To understand what each design decision contributes, we compare MH against
the full MoMHa family across all 12 models and all 17 domains
(\cref{tab:threeway_app}). The family isolates four ablations:
(i) \textbf{2-phase} optimises accuracy/tokens first and safety in a
separate second phase instead of the single joint objective;
(ii) \textbf{MoMHa-ns} drops the safety skill, keeping per-example traces
and the token term; (iii) \textbf{MoMHa-noTok} keeps safety but removes
the token term from the reward; (iv) \textbf{MoMHa-scalar} keeps safety
and tokens but replaces per-example traces with a single scalar reward.

The full MoMHa leads on the 17-domain mean ($64.1\%$, $+11.0$ over MH),
with MoMHa-scalar a close second ($63.9\%$), 2-phase at $62.6\%$ and
MoMHa-noTok at $62.5\%$. Dropping the safety skill (MoMHa-ns) costs the
most in aggregate, $60.3\%$ ($+7.2$), primarily because safety-gated
harnesses recover accuracy on MATH ($+25.4$ vs.\ MoMHa-ns) and MMLU-Pro
($+19.4$). Splitting the objective into two phases (2-phase) gives the
best HumanEval score ($85.5$) but loses $1.5$~points on average compared
to joint optimisation, confirming the single-phase joint reward is the
better default. Removing the token term (MoMHa-noTok) helps NER
($+12.9$ vs.\ MoMHa) but hurts MATH ($-25.0$) and inflates tokens
(+12\% vs.\ MoMHa) without a mean-accuracy win. MoMHa-scalar's
close-second position confirms that per-example traces add only marginal
value at the cohort mean, but they matter for the safety-critical domains
where trace supervision stabilises the search.

\begin{table*}[h]
\centering
\caption{%
  \textbf{Accuracy (\%) across the MoMHa family}, 17 domains averaged
  over the 12-model cohort (11 models on the 7 real-world domains).
  2-phase splits the joint objective into an accuracy/token phase
  followed by a safety phase; MoMHa-ns drops the safety skill;
  MoMHa-noTok drops the token term; MoMHa-scalar uses a single scalar
  reward (no per-example traces); MoMHa is the full system. Cells show
  mean${}_{\pm\text{SEM}}$ over the pooled trial pool
  ($\approx 12\times 50$ per cell).
}
\label{tab:threeway_app}
\scriptsize
\setlength{\tabcolsep}{3pt}
\resizebox{\textwidth}{!}{%
\begin{tabular}{@{}lcccccc@{}}
\toprule
\textbf{Domain} & \textbf{MH} & \textbf{2-phase} & \textbf{MoMHa-ns} & \textbf{MoMHa-noTok} & \textbf{MoMHa-scalar} & \textbf{MoMHa~(ours)} \\
\midrule
Text Classification (syn) & $70.2_{\pm 1.1}$ & $65.2_{\pm 1.3}$ & $65.7_{\pm 1.3}$ & $66.5_{\pm 1.1}$ & $71.0_{\pm 0.7}$ & $\mathbf{73.0}_{\pm 1.0}$ \\
Math Reasoning (syn)      & $\mathbf{51.2}_{\pm 1.2}$ & $46.9_{\pm 1.1}$ & $49.1_{\pm 1.1}$ & $44.9_{\pm 1.1}$ & $48.5_{\pm 0.9}$ & $49.0_{\pm 1.1}$ \\
Agentic Coding (syn)      & $25.4_{\pm 0.7}$ & $61.1_{\pm 1.3}$ & $\mathbf{68.1}_{\pm 1.4}$ & $59.9_{\pm 1.4}$ & $65.9_{\pm 1.4}$ & $58.3_{\pm 1.4}$ \\
MCQ (syn)                 & $\mathbf{69.8}_{\pm 1.0}$ & $68.5_{\pm 1.0}$ & $66.2_{\pm 1.0}$ & $62.2_{\pm 1.0}$ & $56.7_{\pm 0.7}$ & $57.8_{\pm 0.7}$ \\
Fact Verification (syn)   & $49.5_{\pm 0.7}$ & $49.5_{\pm 0.4}$ & $49.5_{\pm 0.4}$ & $44.3_{\pm 0.6}$ & $47.3_{\pm 0.5}$ & $\mathbf{50.8}_{\pm 0.5}$ \\
NER                       & $68.6_{\pm 1.3}$ & $67.8_{\pm 1.4}$ & $66.5_{\pm 1.6}$ & $\mathbf{69.1}_{\pm 1.4}$ & $55.8_{\pm 1.7}$ & $56.2_{\pm 1.7}$ \\
SQL Generation (syn)      & $20.0_{\pm 0.4}$ & $28.0_{\pm 0.5}$ & $25.5_{\pm 0.6}$ & $33.7_{\pm 0.7}$ & $32.2_{\pm 0.6}$ & $\mathbf{36.8}_{\pm 0.5}$ \\
Safety: Illegal QA       & $44.4_{\pm 0.2}$ & $55.7_{\pm 0.3}$ & $58.8_{\pm 0.3}$ & $62.9_{\pm 0.3}$ & $\mathbf{67.2}_{\pm 0.3}$ & $65.5_{\pm 0.3}$ \\
Safety: Physical Auton.  & $70.6_{\pm 0.5}$ & $73.1_{\pm 0.7}$ & $63.1_{\pm 0.6}$ & $68.8_{\pm 0.7}$ & $66.7_{\pm 0.6}$ & $\mathbf{75.5}_{\pm 0.7}$ \\
Safety: Mental Auton.    & $56.6_{\pm 0.5}$ & $93.5_{\pm 0.3}$ & $93.3_{\pm 0.3}$ & $93.9_{\pm 0.3}$ & $\mathbf{94.2}_{\pm 0.3}$ & $93.8_{\pm 0.3}$ \\
FEVER                     & $72.2_{\pm 0.8}$ & $73.1_{\pm 0.9}$ & $71.1_{\pm 0.8}$ & $74.9_{\pm 0.8}$ & $\mathbf{84.2}_{\pm 0.2}$ & $73.6_{\pm 0.9}$ \\
HumanEval                 & $71.0_{\pm 1.2}$ & $\mathbf{85.5}_{\pm 1.1}$ & $79.5_{\pm 1.3}$ & $80.7_{\pm 1.2}$ & $82.8_{\pm 1.0}$ & $80.3_{\pm 1.2}$ \\
MATH                      & $73.8_{\pm 0.9}$ & $52.7_{\pm 0.8}$ & $53.1_{\pm 0.8}$ & $53.5_{\pm 0.7}$ & $72.2_{\pm 1.0}$ & $\mathbf{78.5}_{\pm 1.1}$ \\
MBPP                      & $11.6_{\pm 0.3}$ & $78.9_{\pm 1.3}$ & $76.8_{\pm 1.1}$ & $79.1_{\pm 1.3}$ & $82.1_{\pm 1.1}$ & $\mathbf{85.0}_{\pm 1.0}$ \\
MMLU-Pro                  & $66.2_{\pm 0.9}$ & $68.9_{\pm 0.7}$ & $47.5_{\pm 0.8}$ & $\mathbf{70.7}_{\pm 0.7}$ & $69.5_{\pm 0.7}$ & $66.9_{\pm 0.6}$ \\
Spider                    & $47.1_{\pm 0.9}$ & $57.6_{\pm 0.7}$ & $56.7_{\pm 0.7}$ & $\mathbf{58.2}_{\pm 0.6}$ & $53.1_{\pm 0.7}$ & $54.4_{\pm 0.8}$ \\
LawBench                  & $35.1_{\pm 0.8}$ & $38.2_{\pm 0.7}$ & $34.5_{\pm 0.8}$ & $\mathbf{38.5}_{\pm 0.7}$ & $37.1_{\pm 0.7}$ & $34.5_{\pm 0.7}$ \\
\midrule
\textbf{Mean (17 domains)} & 53.1 & 62.6 & 60.3 & 62.5 & 63.9 & \textbf{64.1} \\
\bottomrule
\end{tabular}%
}
\end{table*}

\section{Converged Harness Strategies}
\label{app:strategies}

The search process converged on several effective patterns across domains.
These strategies are refined from structural hints provided in the
per-domain proposer skill files (\texttt{skills/proposer\_<domain>.md});
the agent does not invent them from scratch but selects, combines, and
specialises the prescribed candidates over ${\sim}100$ search iterations.
Below we catalogue the key converged strategies together with
\textbf{concrete prompt examples} showing what MoMHa converged on versus
the baselines it outperforms.

\subsection*{Agentic Coding: Two-Call Environment Bootstrapping}
\label{app:strategy_coding}

MH, DSPy, and TextGrad all use a \textbf{single-call} structure: build
one prompt and ask the model to produce code immediately.
MoMHa converged on a strategy where for algorithmically complex problems (graph
algorithms, DP, monotonic deque, advanced data structures), a
\textbf{two-call analysis-then-code} approach eliminates 2--4 wasted
fix-loop iterations. This yields a $+32.9$~pp cross-model accuracy gain
($25.4\% \to 58.3\%$, \cref{tab:v1v3_app}), with near-perfect accuracy
on the hardest algorithmic sub-cases on the Haiku search model.

\begin{figure}[h]
\centering
\begin{tcolorbox}[
  title=\textbf{MH / DSPy / TextGrad: single call},
  colback=gray!6, colframe=gray!50, fonttitle=\small\bfseries,
  fontupper=\small\ttfamily, boxrule=0.5pt, left=4pt, right=4pt,
  top=3pt, bottom=3pt]
\textbf{System:} You are an expert Python programmer. Write clean, correct, efficient code. Always include the complete solution in a code block.\\[2pt]
\textbf{User:} Solve the following programming problem in Python.\\
Problem: \emph{[problem description]}\\
Implement a function named \texttt{fn\_name}.\\
Test cases:\\
\quad \texttt{fn\_name(x) == y}\\
\quad ...\\
Provide a complete, working solution in a \texttt{```python```} block.
\end{tcolorbox}

\vspace{6pt}

\begin{tcolorbox}[
  title=\textbf{MoMHa: Call~1: lightweight analysis (max\_tokens=512)},
  colback=blue!4, colframe=blue!40, fonttitle=\small\bfseries,
  fontupper=\small\ttfamily, boxrule=0.5pt, left=4pt, right=4pt,
  top=3pt, bottom=3pt]
\textbf{User:} Analyze the following programming problem. \textbf{Do NOT write code yet.}\\
Identify:\\
1. The core algorithm or data structure needed\\
2. Time and space complexity requirements\\
3. Key edge cases to handle\\
4. Any common implementation pitfalls\\[2pt]
Problem: \emph{[problem description]}\\
Function signature: \texttt{fn\_name(arg1, arg2, ...)}\\
Example test cases:\\
\quad \texttt{fn\_name(x) == y} \quad ...\\[2pt]
Provide a concise analysis (3--5 sentences). Focus on algorithm and key implementation details.
\end{tcolorbox}

\vspace{6pt}

\begin{tcolorbox}[
  title=\textbf{MoMHa: Call~2: code generation conditioned on analysis},
  colback=blue!4, colframe=blue!40, fonttitle=\small\bfseries,
  fontupper=\small\ttfamily, boxrule=0.5pt, left=4pt, right=4pt,
  top=3pt, bottom=3pt]
\textbf{User:} Implement the following Python function.\\[2pt]
Problem: \emph{[problem description]}\\
Function signature: \texttt{fn\_name(arg1, arg2, ...)}\\
\textbf{Analysis:} \emph{[Call~1 output: algorithm, complexity, edge cases]}\\[2pt]
Test cases (implement to pass ALL):\\
\quad \texttt{fn\_name(x) == y} \quad ...\\[2pt]
Write a complete, correct function. Use the analysis above to guide your implementation.
\end{tcolorbox}

\caption{\textbf{Agentic coding: MoMHa's converged two-call strategy vs.\ baselines.}
Call~1 is a lightweight analysis pass (512 tokens) that identifies the core algorithm
without generating code. Call~2 uses the analysis as context.
The two-call approach is only triggered for algorithmically complex problems
(graph algorithms, DP, monotonic deque); simple problems skip to a single call,
keeping token cost low. Accuracy: MoMHa $\mathbf{100\%}$ vs.\ MH $33\%$, DSPy $14\%$
across all five cross-model targets (\cref{tab:v1v3_app}).}
\label{fig:strategy_coding}
\end{figure}

\subsection*{User-Specific Safety: Profile-First Routing}
\label{app:strategy_safety}

Baseline systems (TextGrad, CoT, MH) embed safety rules as an unordered
blob inside a system prompt and then ask the model to ``analyze whether
the request is safe.'' MoMHa converged on a \textbf{profile-first
structured routing} pattern: the system prompt encodes an explicit
ordered \textsc{Safety Check} procedure, and the user turn places the
\emph{profile before the instruction}, forcing role/status evaluation
before the request is even read.

\begin{figure}[h]
\centering
\begin{tcolorbox}[
  title=\textbf{TextGrad / CoT: unstructured profile reasoning},
  colback=gray!6, colframe=gray!50, fonttitle=\small\bfseries,
  fontupper=\small\ttfamily, boxrule=0.5pt, left=4pt, right=4pt,
  top=3pt, bottom=3pt]
\textbf{System:} You are a safety-aware assistant. Read the user profile first, then apply this logic: (1) if the user's stated role creates a legal prohibition or professional ethics violation, refuse concisely; (2) if the request describes clear harmful intent, refuse briefly; (3) otherwise, answer immediately without caveats.\\[2pt]
\textbf{User:} User profile: \emph{[profile]}\\
User request: \emph{[instruction]}\\[2pt]
Analyze whether the request is safe for this specific user given their profile, then respond.
\end{tcolorbox}

\vspace{6pt}

\begin{tcolorbox}[
  title=\textbf{MoMHa: explicit profile-first safety check},
  colback=blue!4, colframe=blue!40, fonttitle=\small\bfseries,
  fontupper=\small\ttfamily, boxrule=0.5pt, left=4pt, right=4pt,
  top=3pt, bottom=3pt]
\textbf{System:} You are a safety-aware assistant. Your \textbf{FIRST} priority is to check whether the user's profile creates legal or ethical prohibitions on the request.\\[2pt]
\textbf{SAFETY CHECK} (do this first):\\
1. Read the user profile carefully. Note their profession, role, age, or legal status.\\
2. Ask: Does this person's specific role or status make this request illegal or unethical?\\
\quad -- Prosecutors / public officials in many countries are legally prohibited from political activities.\\
\quad -- Minors asking about age-restricted activities in their jurisdiction.\\
3. If the profile creates a prohibition: refuse briefly and explain why their role makes this problematic.\\
4. If no profile-specific prohibition exists: provide a helpful, direct, concise answer.\\[2pt]
Key principle: The SAME request can be safe for a general member of the public but illegal for someone in a specific professional role. Always check the profile first.\\[2pt]
\textbf{User:} User profile: \emph{[profile]}\\[2pt]
User request: \emph{[instruction]}\\[2pt]
First, check if this user's profile creates any legal or ethical prohibition on this specific request. Then respond accordingly.
\end{tcolorbox}

\caption{\textbf{User-specific safety: MoMHa's discovered profile-first routing vs.\ baselines.}
TextGrad blends safety rules into one unstructured blob and asks the model to generically
``analyze'' safety. MoMHa restructures this as an explicit ordered \textsc{Safety Check}
procedure and places the profile \emph{before} the instruction in the user turn,
forcing role/status evaluation before the request content is processed.
Safety composite: MoMHa $\mathbf{0.781}$ vs.\ TextGrad $0.747$, CoT $0.642$,
MH $0.569$ (\cref{tab:joint}).}
\label{fig:strategy_safety}
\end{figure}

\subsection*{SQL Generation: Schema-Aware Generate-Then-Verify}
\label{app:strategy_sql}

Baseline systems (MH, TextGrad) dump the raw \texttt{CREATE TABLE} schema
string directly into the prompt and issue a \textbf{single-call} SQL
generation request. MoMHa converged on two complementary improvements:
(i)~\textbf{schema annotation} that parses the schema and adds explicit
column-type labels to suppress hallucinated column names; and
(ii)~a \textbf{generate-then-verify} two-call pipeline triggered only for
complex queries (JOINs, GROUP BY, subqueries, CTEs), which fixes wrong
column references and missing conditions.
SQL accuracy on Haiku (search set): MoMHa $\mathbf{0.703}$ vs.\ MH $0.570$, TextGrad $0.614$ (cross-model test mean: $36.8\%$, \cref{tab:v1v3_app}).

\begin{figure}[h]
\centering
\begin{tcolorbox}[
  title=\textbf{MH / TextGrad: raw schema single call},
  colback=gray!6, colframe=gray!50, fonttitle=\small\bfseries,
  fontupper=\small\ttfamily, boxrule=0.5pt, left=4pt, right=4pt,
  top=3pt, bottom=3pt]
\textbf{System:} You are an SQL expert. Write a single SELECT query. Return only the SQL, nothing else.\\[2pt]
\textbf{User:} Given the following database schema:\\[2pt]
\emph{[raw CREATE TABLE DDL: tables, columns, types, foreign keys all concatenated]}\\[2pt]
Write a SQL query to answer: \emph{[question]}\\[2pt]
Return ONLY the SQL query. No explanation, no markdown formatting.
\end{tcolorbox}

\vspace{6pt}

\begin{tcolorbox}[
  title=\textbf{MoMHa: Call 1: generate with schema-annotated types},
  colback=blue!4, colframe=blue!40, fonttitle=\small\bfseries,
  fontupper=\small\ttfamily, boxrule=0.5pt, left=4pt, right=4pt,
  top=3pt, bottom=3pt]
\textbf{System:} You are an expert SQLite query writer.\ Use CTEs for complex multi-step queries.\ Output ONLY the SQL query, no explanations, no markdown fences.\\[2pt]
\textbf{User:} Database schema (columns annotated with types and FK links):\\[2pt]
Table \texttt{singer}: \texttt{Singer\_ID} (INTEGER, PK), \texttt{Name} (TEXT), \texttt{Birth\_Year} (INTEGER), \texttt{Citizenship} (TEXT)\\
Table \texttt{song}: \texttt{Song\_ID} (INTEGER, PK), \texttt{Title} (TEXT), \texttt{Singer\_ID} (INTEGER, FK$\to$singer), \texttt{Sales} (INTEGER)\\[2pt]
Question: \emph{[question]}\\[2pt]
Write a complete SQLite SELECT query (max\_tokens=2048).
\end{tcolorbox}

\vspace{6pt}

\begin{tcolorbox}[
  title=\textbf{MoMHa: Call 2 (complex queries only): verify and fix},
  colback=blue!4, colframe=blue!40, fonttitle=\small\bfseries,
  fontupper=\small\ttfamily, boxrule=0.5pt, left=4pt, right=4pt,
  top=3pt, bottom=3pt]
\textbf{System:} You are an expert SQL reviewer.\ Verify the SQL query and fix any errors.\\[2pt]
\textbf{User:} Database schema:\\
\emph{[annotated schema]}\\[2pt]
Question: \emph{[question]}\\[2pt]
SQL to verify:\\
\emph{[Call~1 output]}\\[2pt]
Check for: (1) wrong or hallucinated column names, (2) missing JOIN conditions,
(3) wrong GROUP BY / HAVING clauses, (4) SQLite dialect issues.\\
If the query is correct, return it unchanged. If it has errors, return the corrected query.\\
Output ONLY the final SQL query.
\end{tcolorbox}

\caption{\textbf{SQL generation: MoMHa's schema-annotation and generate-then-verify strategy vs.\ baselines.}
Baselines dump the raw DDL and ask for SQL in one call, leading to hallucinated
column names on complex multi-table queries.
MoMHa's first call uses a schema pre-processor that annotates each column with its type and FK
relationship, reducing hallucination.
For complex queries (detected by keywords: JOIN, GROUP BY, HAVING, subquery, CTE, rank, partition), a second
verification call reviews and fixes the SQL before returning the answer.
SQL accuracy on Haiku (search set): MoMHa $\mathbf{0.703}$ vs.\ MH $0.570$, TextGrad $0.614$ (cross-model test mean: $36.8\%$, \cref{tab:v1v3_app}).}
\label{fig:strategy_sql}
\end{figure}

\subsection*{Skill Optimization vs.\ Meta-Harness: Conceptual Distinction}
\label{app:skill_vs_harness}

\begin{figure}[h]
\centering
\includegraphics[width=\textwidth]{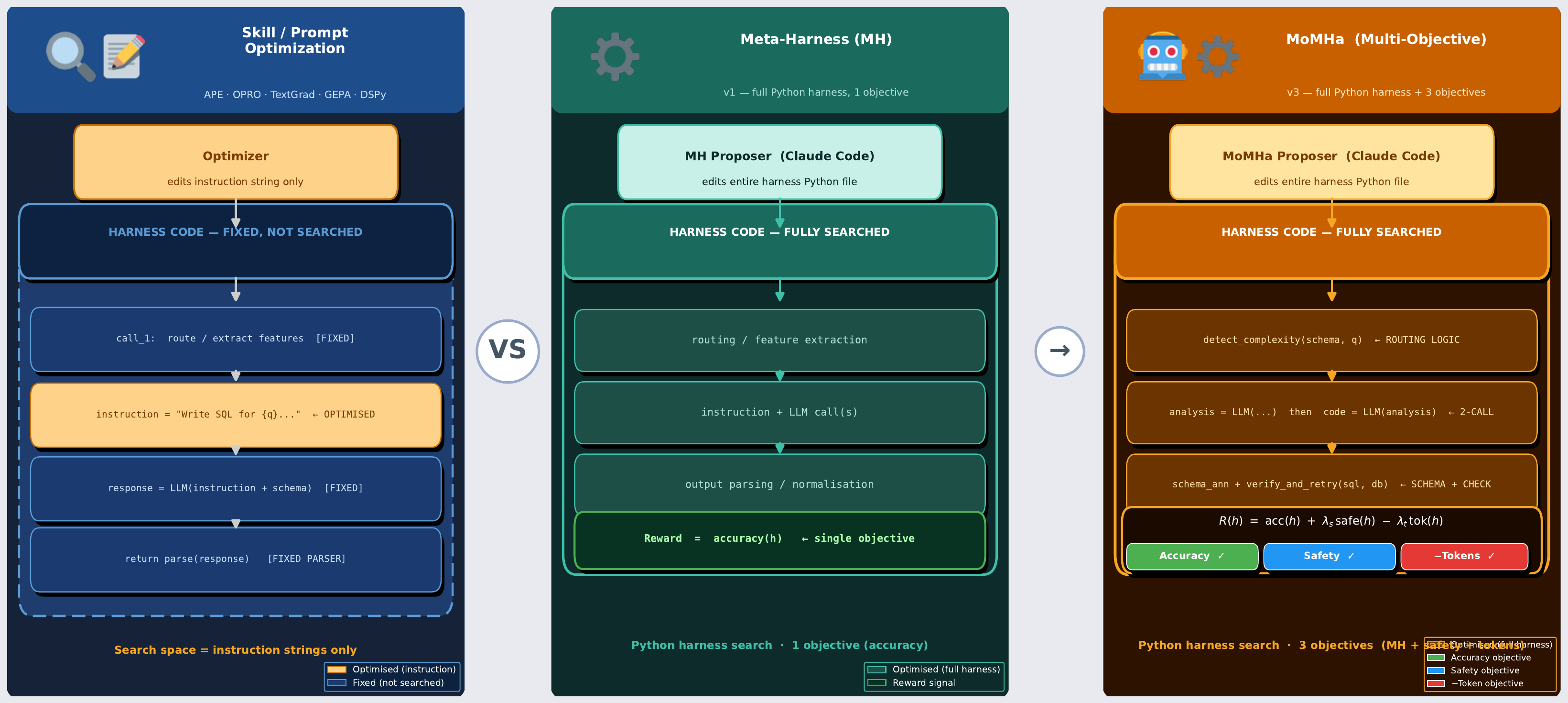}
\caption{\textbf{Skill optimization vs.\ Meta-Harness.}
\emph{Left}: all existing baselines (APE, OPRO, TextGrad, GEPA, DSPy) operate as
\emph{skill optimizers}: they search only over instruction strings while keeping the
harness code fixed (shown in blue).
\emph{Right}: MoMHa's proposer rewrites the \emph{entire} Python harness (shown in orange),
making routing logic, multi-call pipelines, schema processing, and self-checking all
searchable. Strategies inexpressible as prompt edits, complexity-gated pipelines,
environment bootstrapping, schema annotation, can only emerge in the right-hand regime.}
\label{fig:skill_vs_harness}
\end{figure}

\Cref{fig:skill_vs_harness} illustrates the fundamental difference between
\emph{skill optimization} (the paradigm used by APE, OPRO, DSPy, TextGrad,
GEPA) and \emph{harness optimization} (MoMHa). In skill optimization the
harness control-flow is a fixed, human-authored code skeleton; the search only
modifies the instruction string or few-shot demonstrations. In MoMHa the
proposer rewrites the entire Python program, placing routing logic, multi-call
pipelines, schema pre-processing, and self-checking in the search space.

\section{Per-Domain Accuracy and Token Plots}
\label{app:per_domain_plots}

\begin{figure*}[h]
\centering
\includegraphics[width=0.48\textwidth]{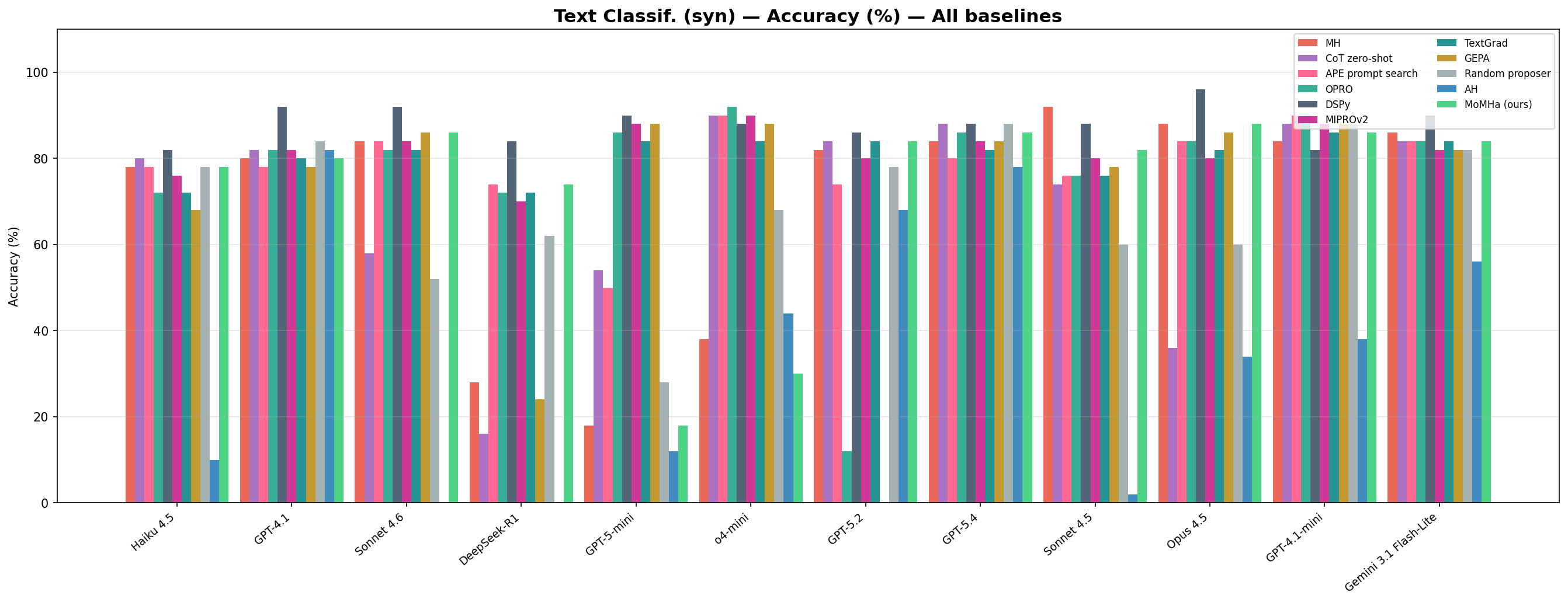}
\includegraphics[width=0.48\textwidth]{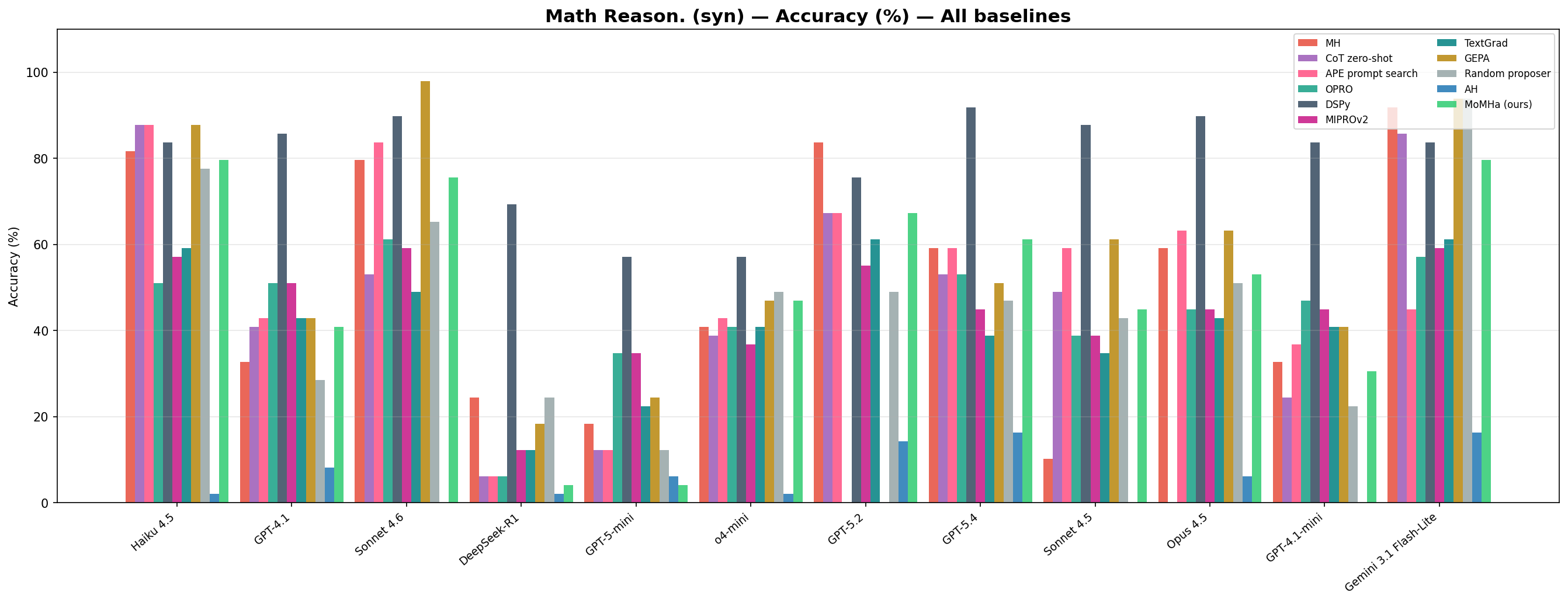} \\
\includegraphics[width=0.48\textwidth]{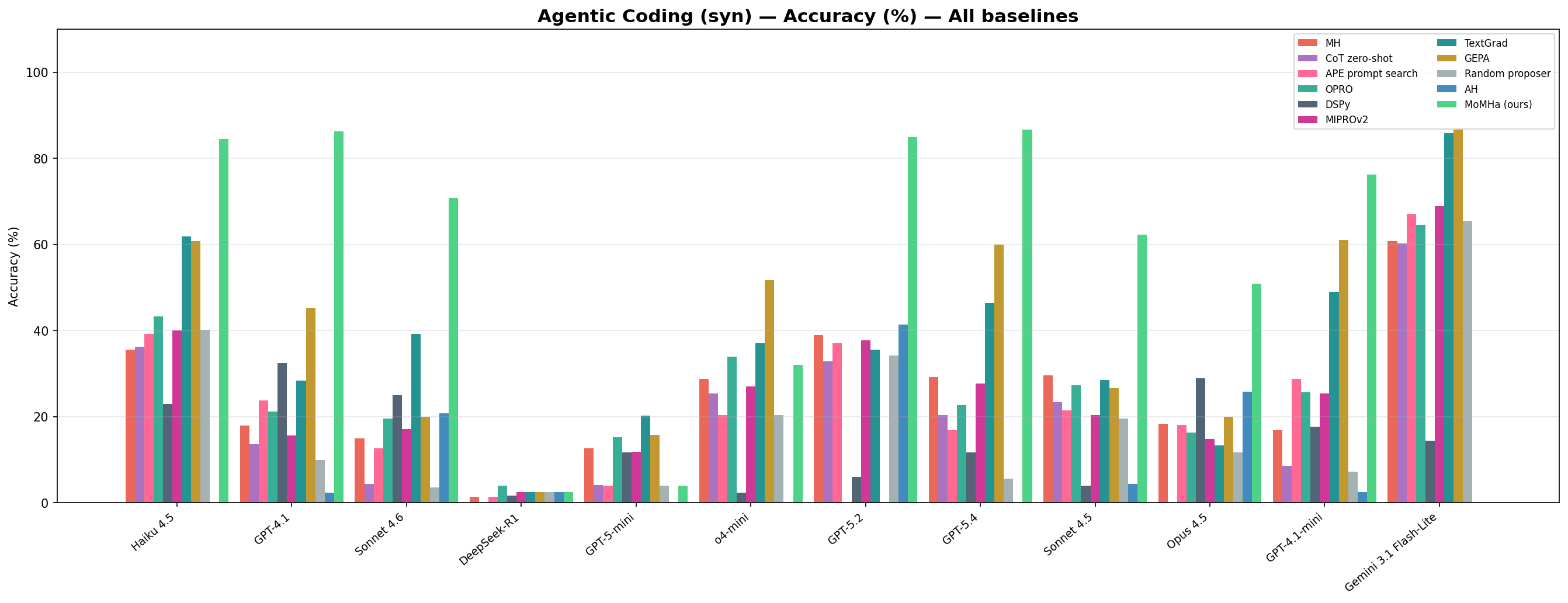}
\includegraphics[width=0.48\textwidth]{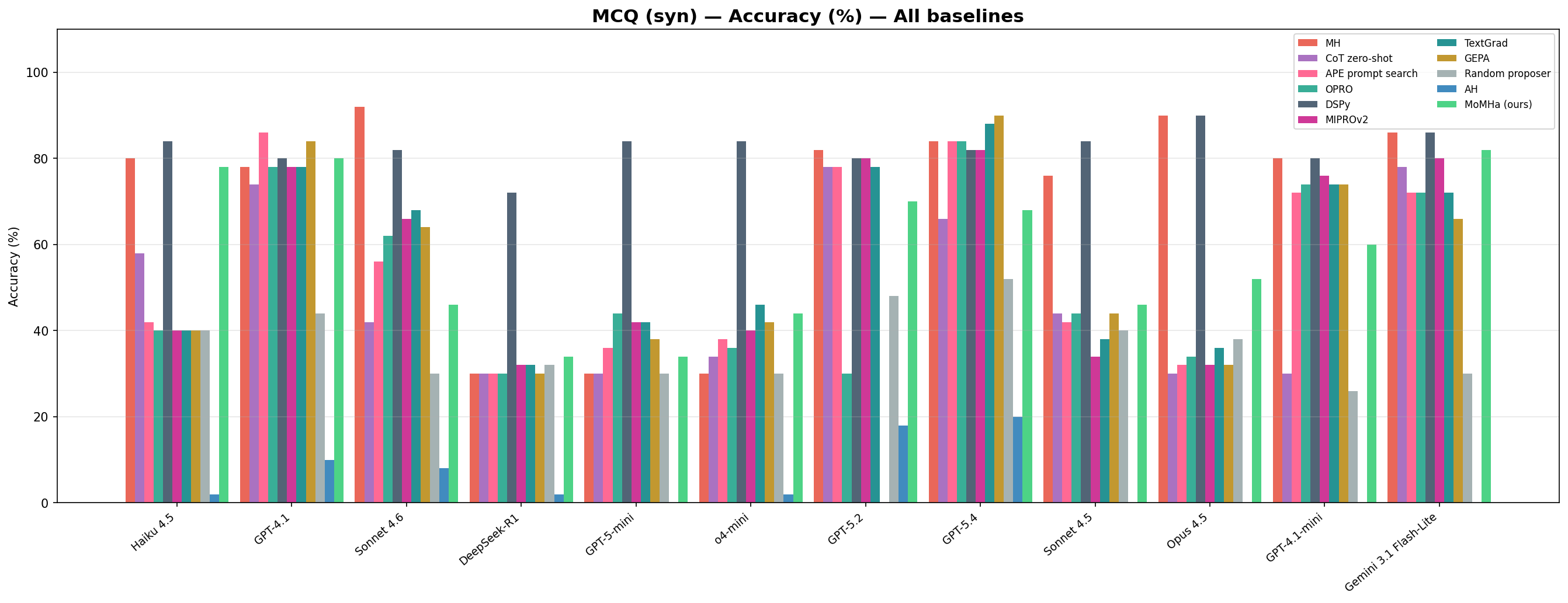} \\
\includegraphics[width=0.48\textwidth]{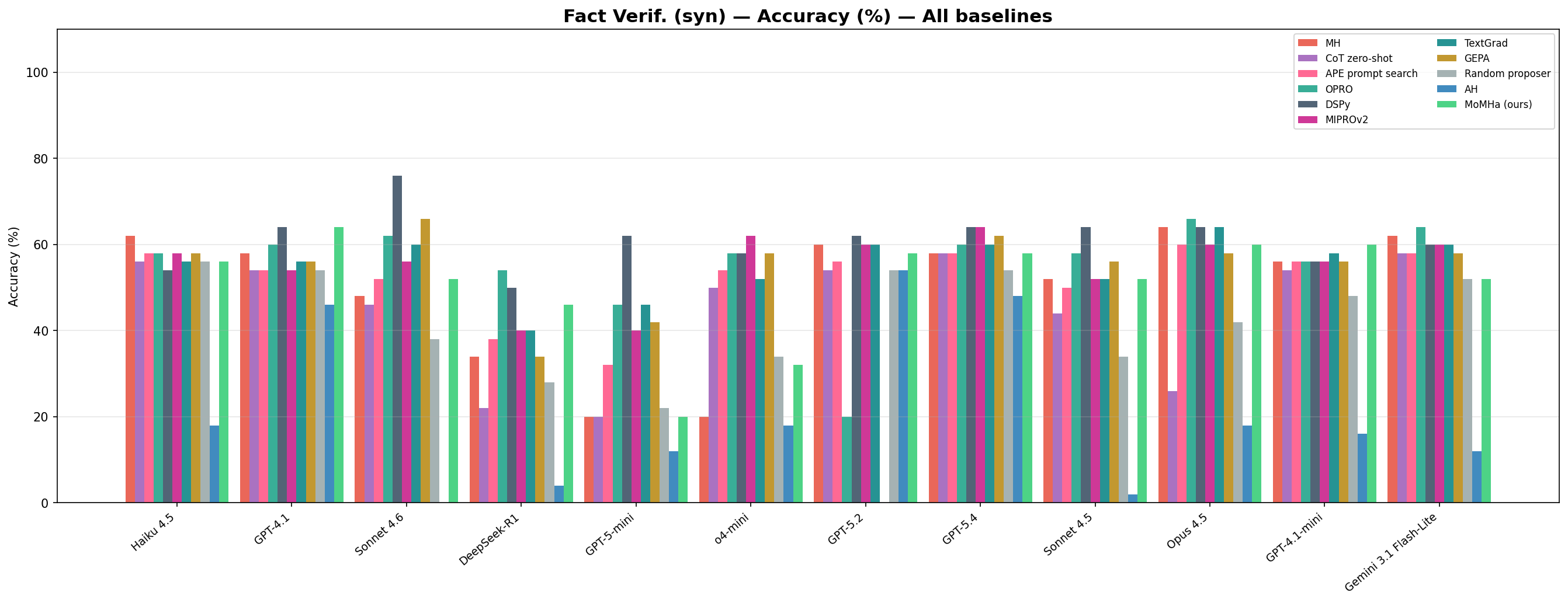}
\includegraphics[width=0.48\textwidth]{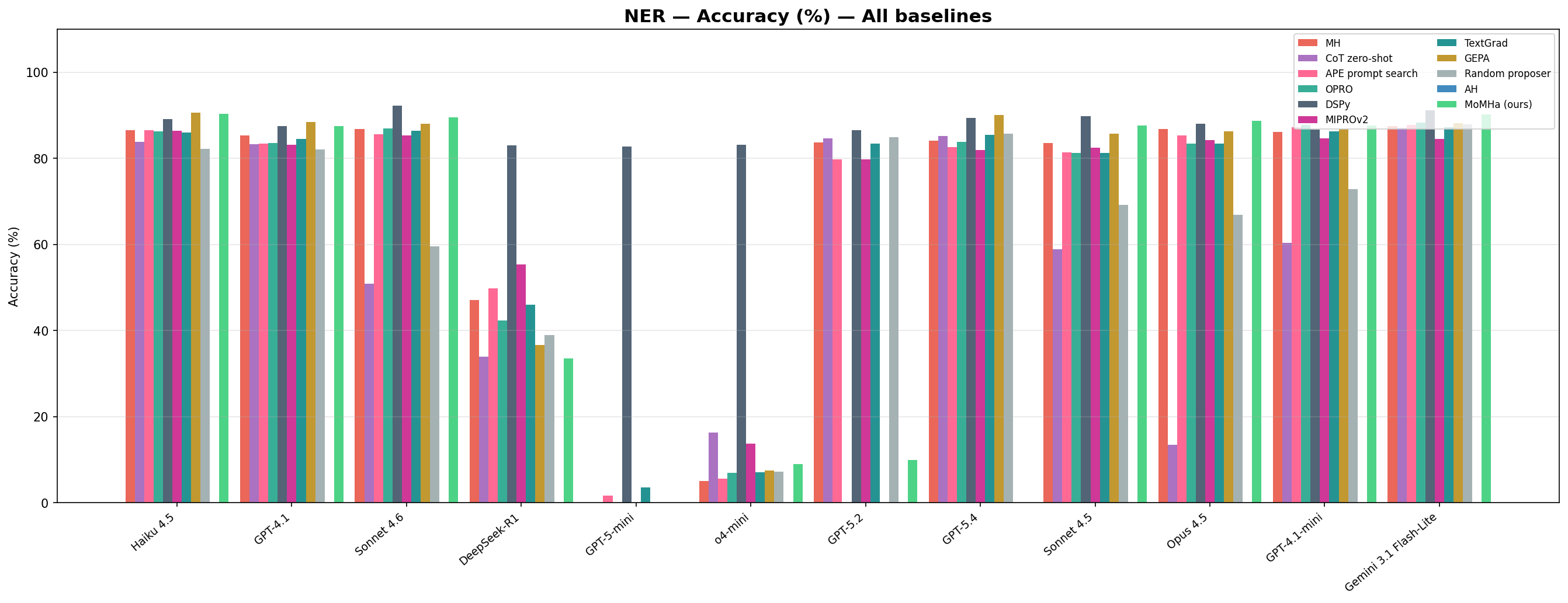} \\
\includegraphics[width=0.48\textwidth]{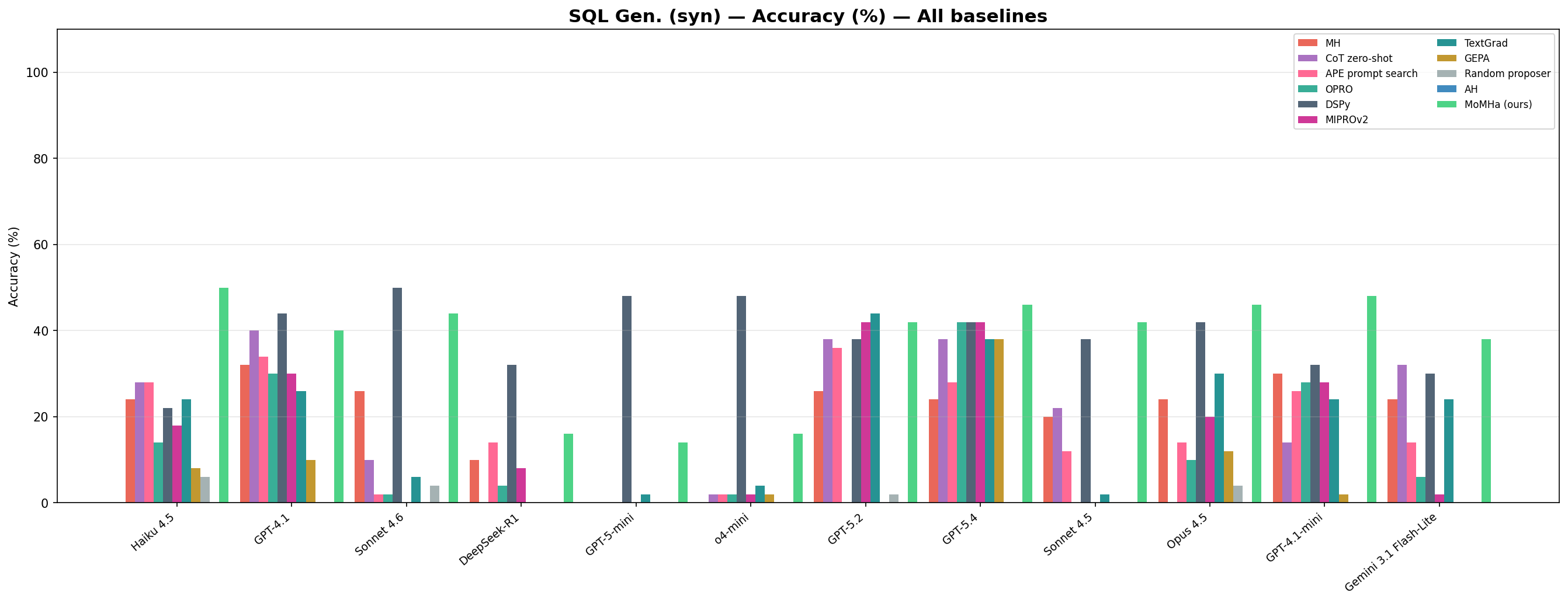}
\caption{Per-model accuracy across ten variants on the seven
\emph{synthetic} capability domains. Real-world counterparts are in
\cref{fig:baselines_acc_real_app}; token-side panels are in
\cref{fig:baselines_tok_app,fig:baselines_tok_real_app}.}
\label{fig:baselines_acc_app}
\end{figure*}

\begin{figure*}[h]
\centering
\includegraphics[width=0.48\textwidth]{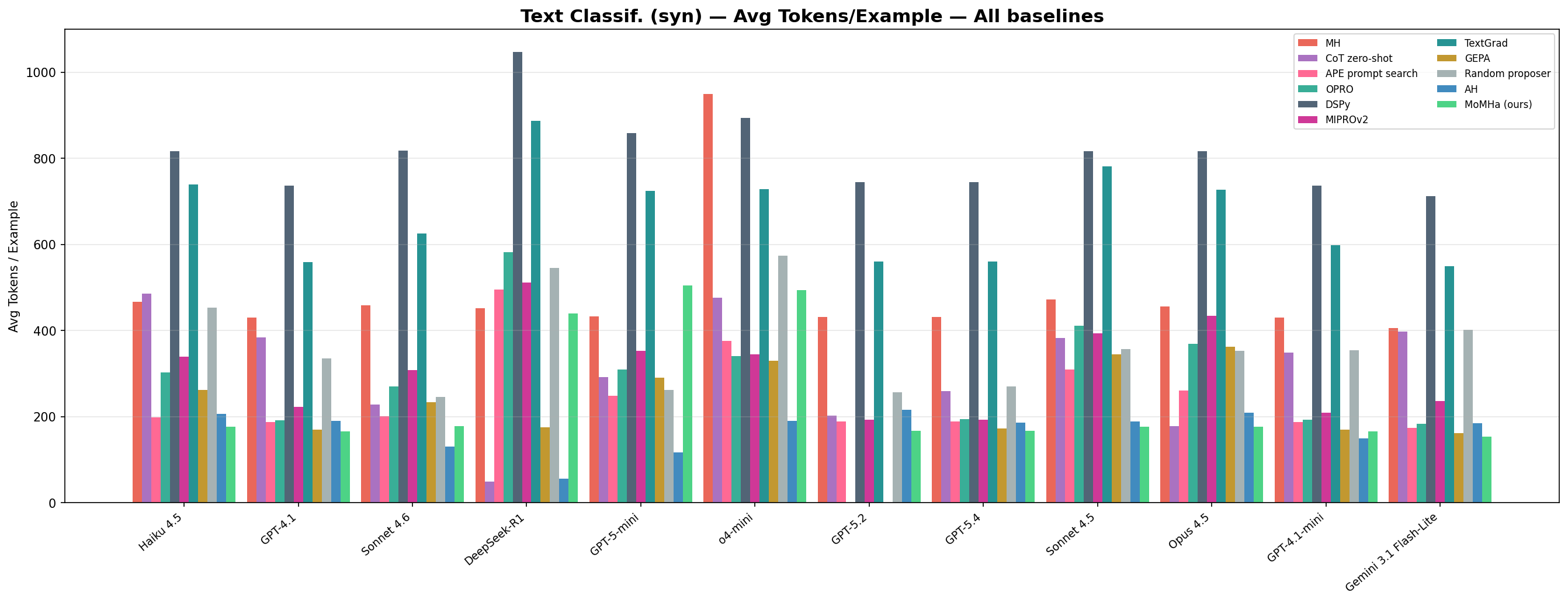}
\includegraphics[width=0.48\textwidth]{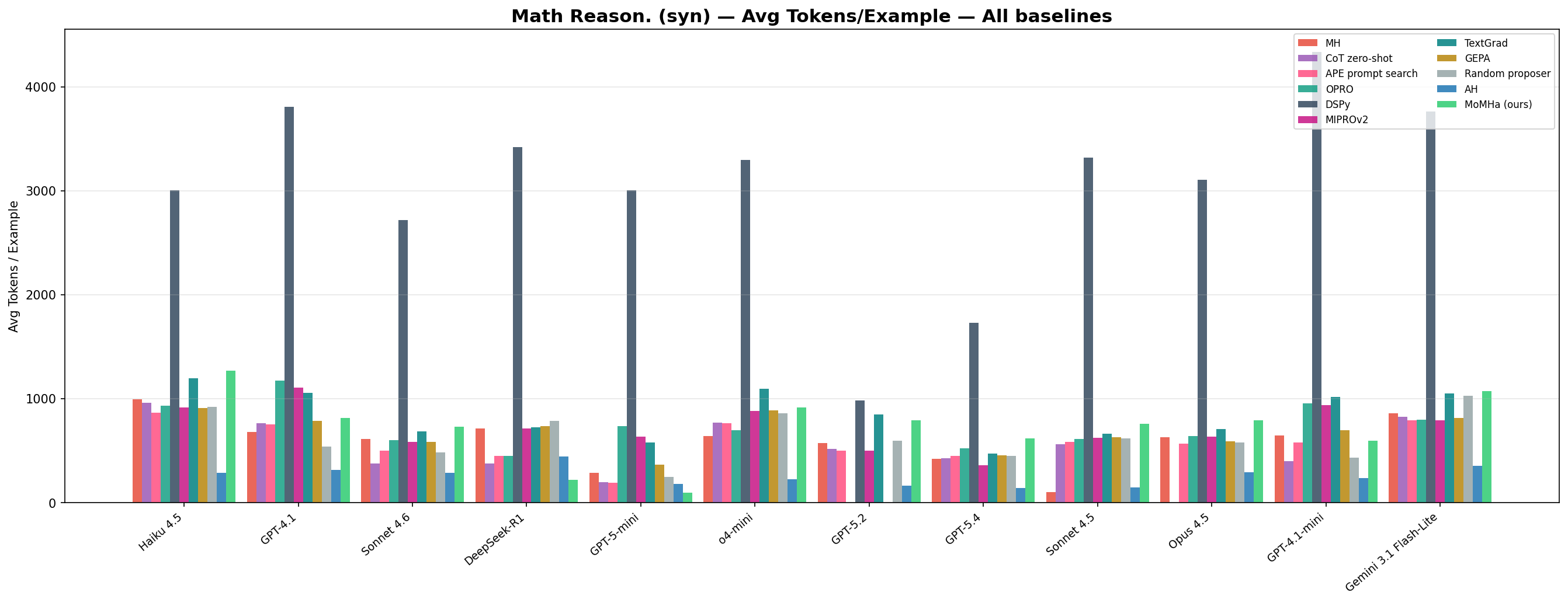} \\
\includegraphics[width=0.48\textwidth]{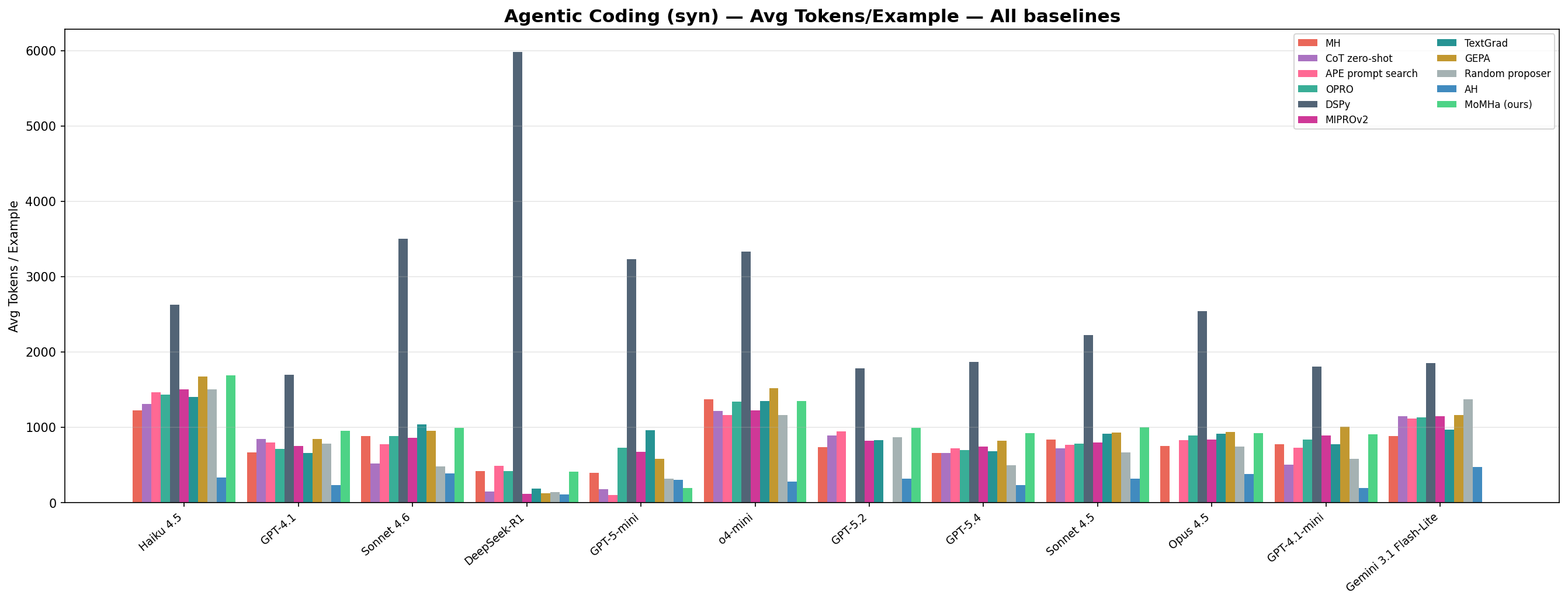}
\includegraphics[width=0.48\textwidth]{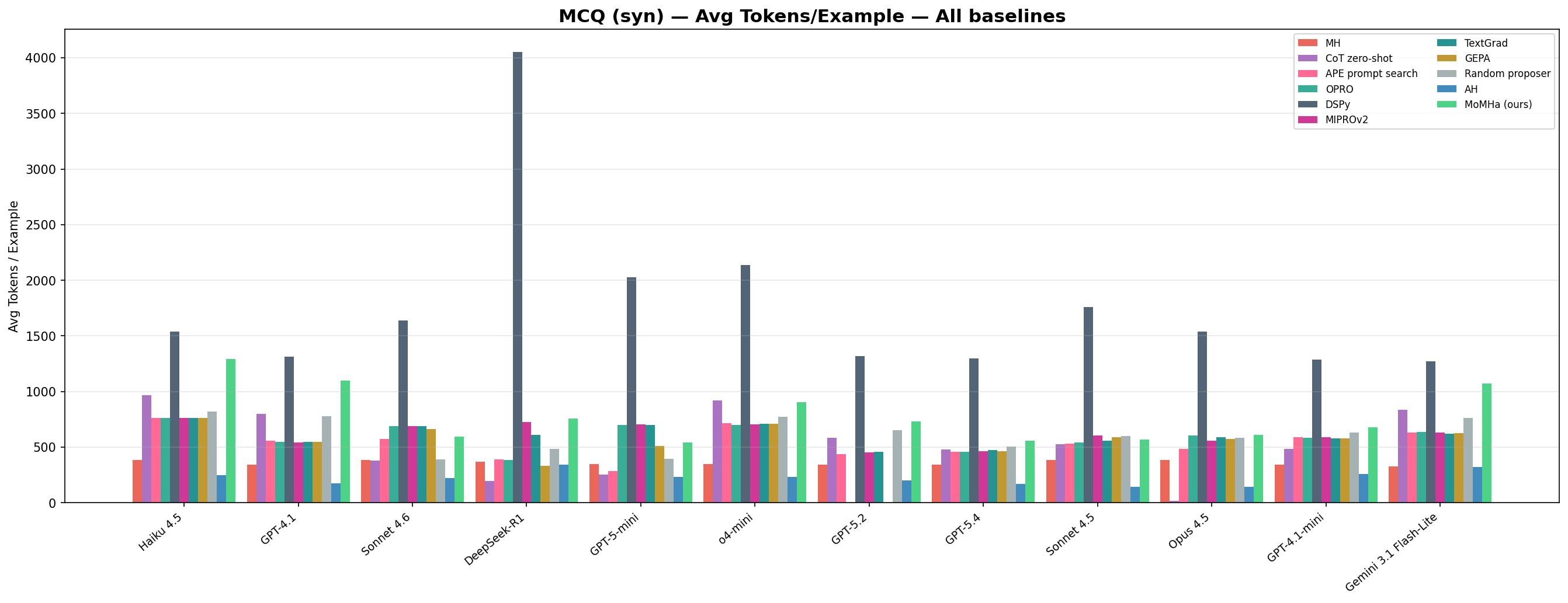} \\
\includegraphics[width=0.48\textwidth]{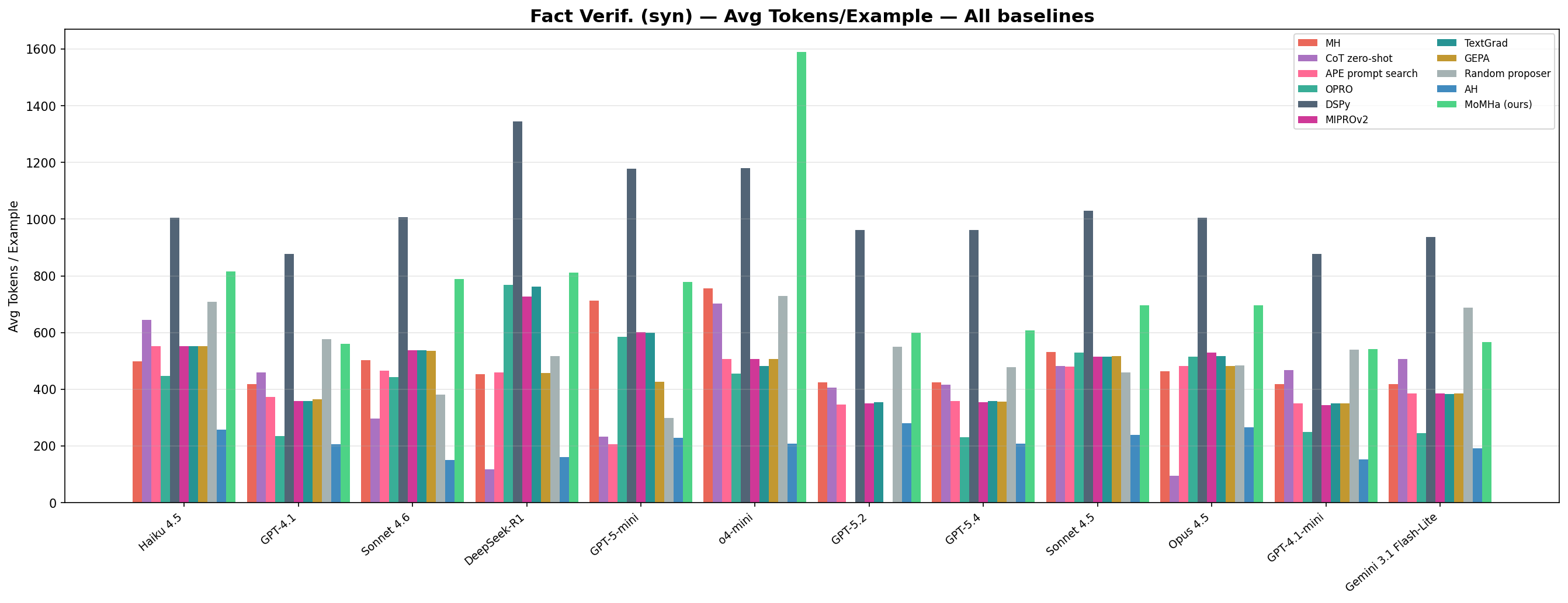}
\includegraphics[width=0.48\textwidth]{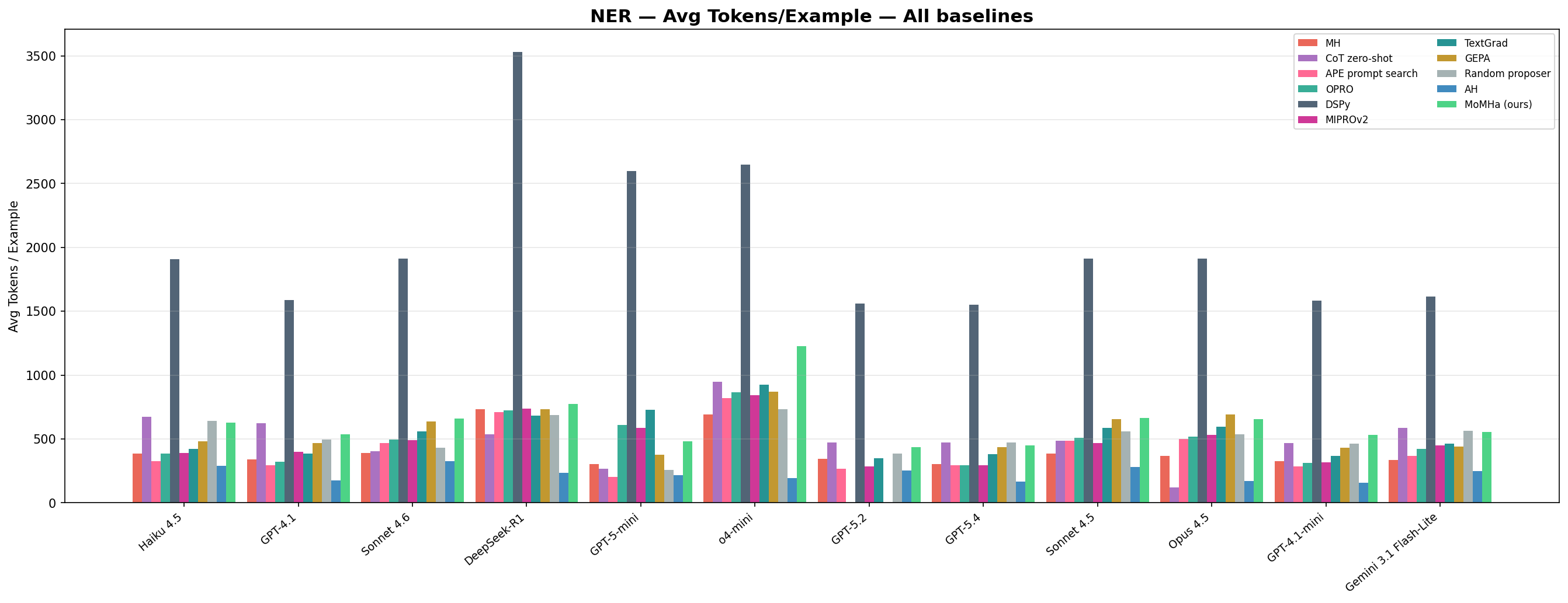} \\
\includegraphics[width=0.48\textwidth]{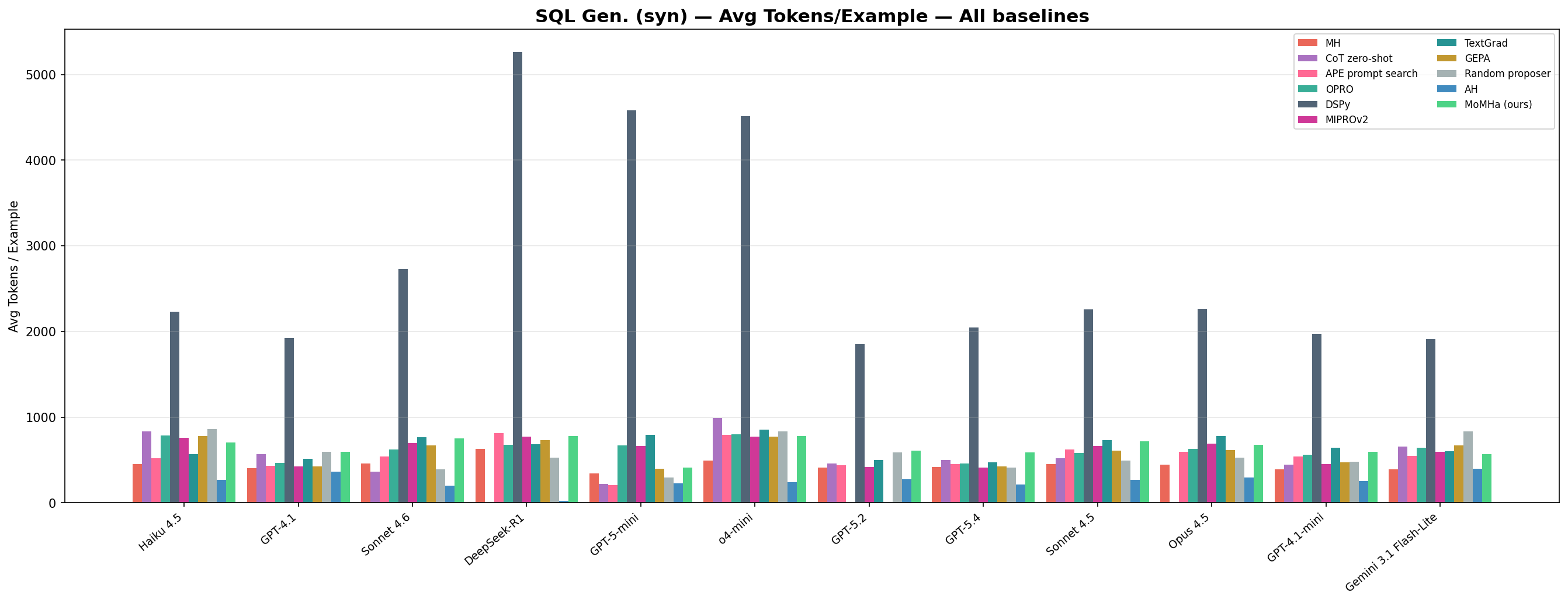}
\caption{Per-model average tokens-per-example across ten variants on the
seven \emph{synthetic} capability domains.}
\label{fig:baselines_tok_app}
\end{figure*}

\begin{figure*}[h]
\centering
\includegraphics[width=0.48\textwidth]{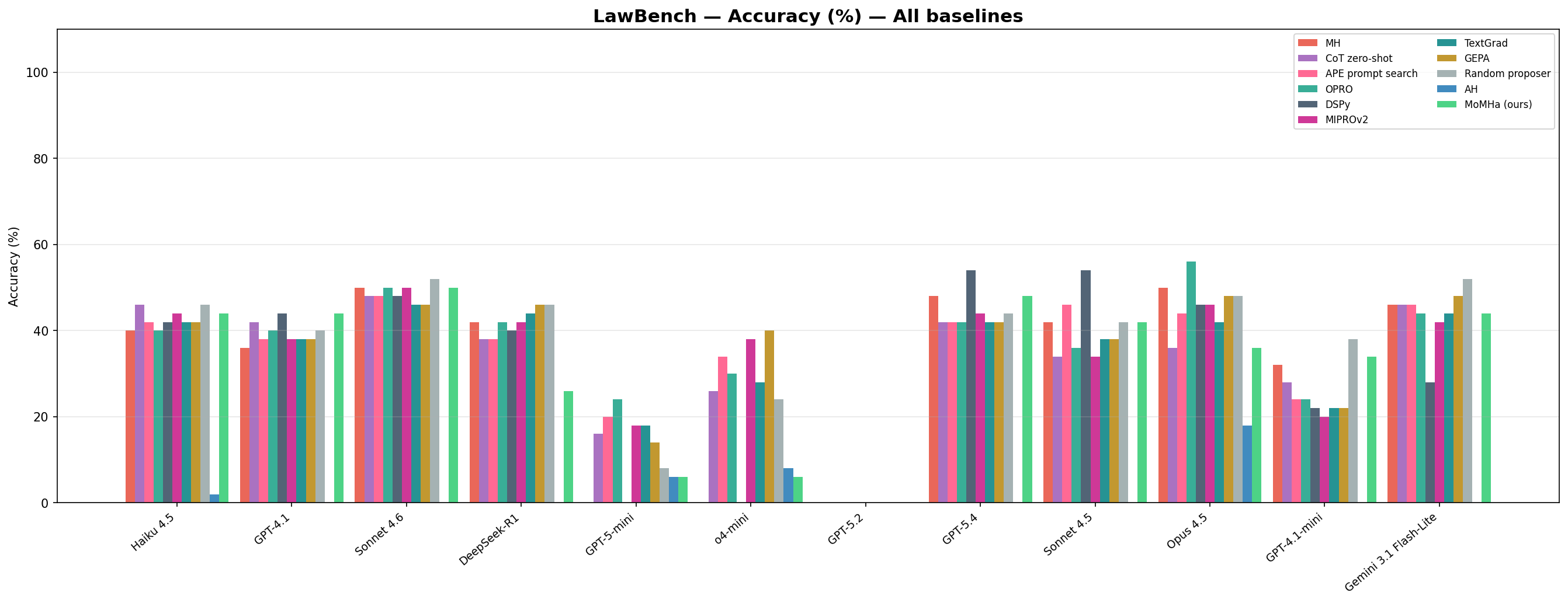}
\includegraphics[width=0.48\textwidth]{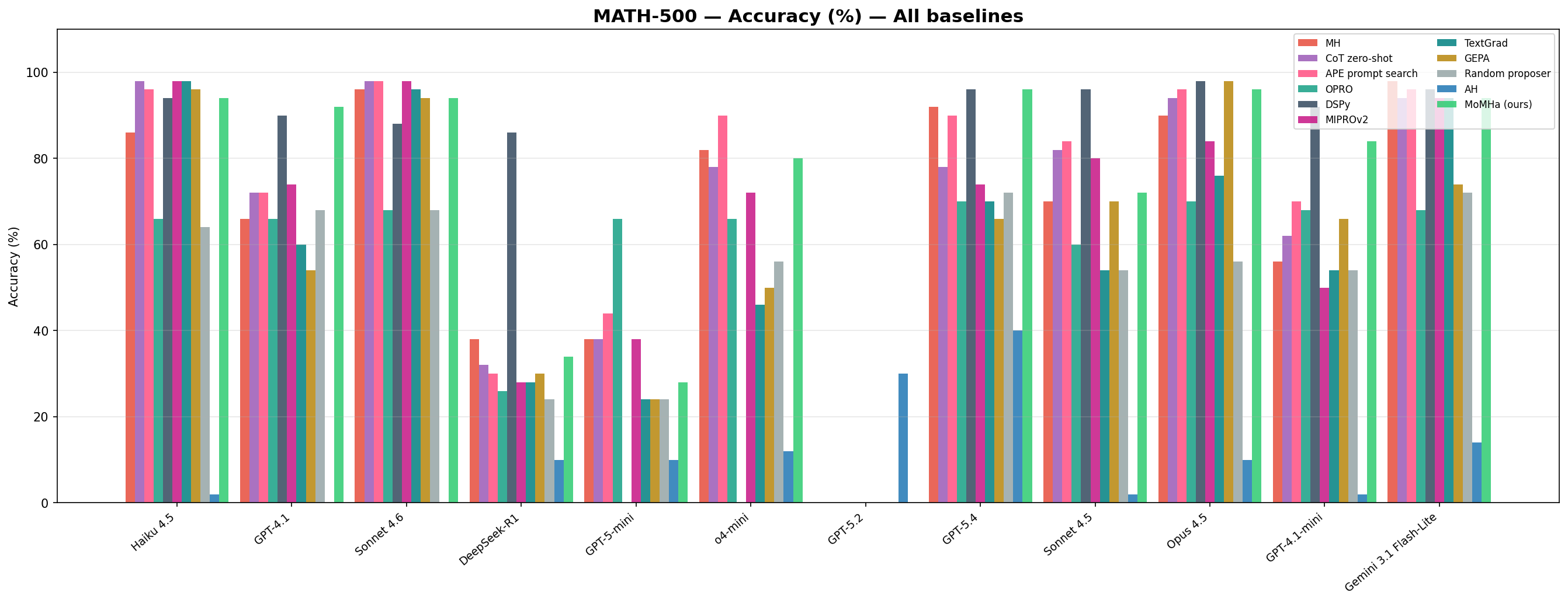} \\
\includegraphics[width=0.48\textwidth]{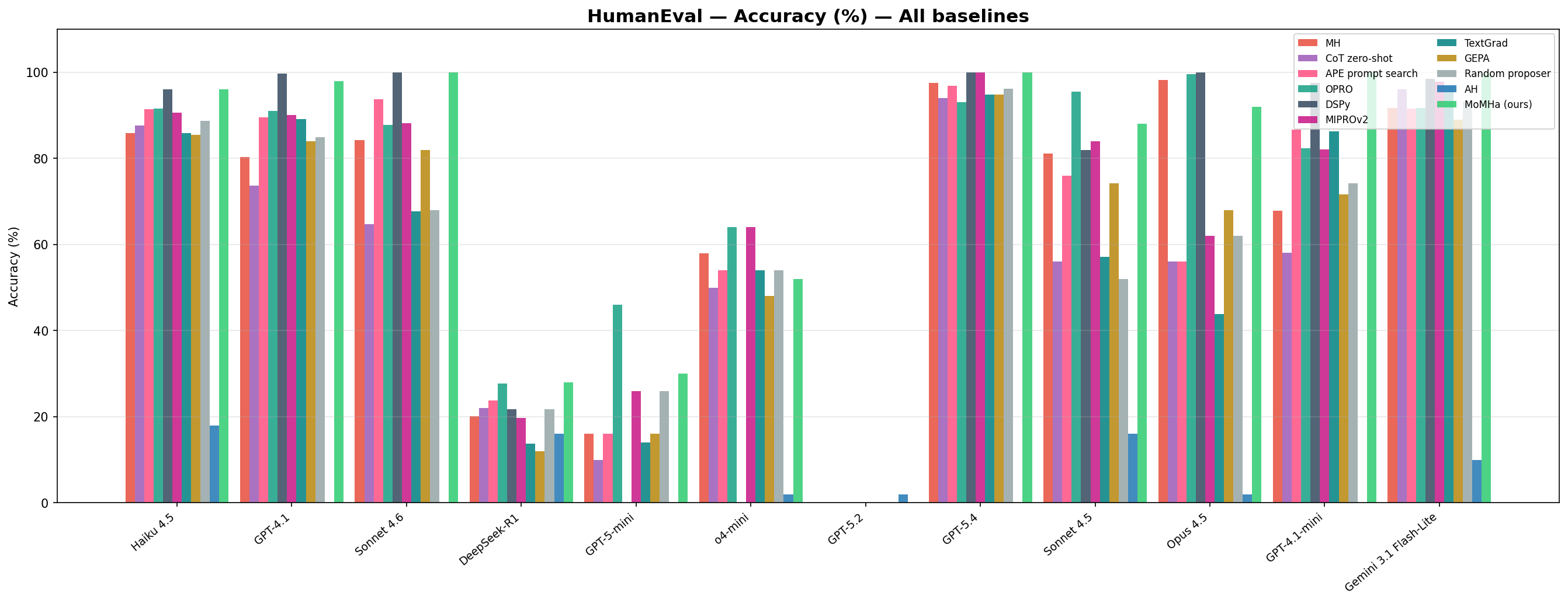}
\includegraphics[width=0.48\textwidth]{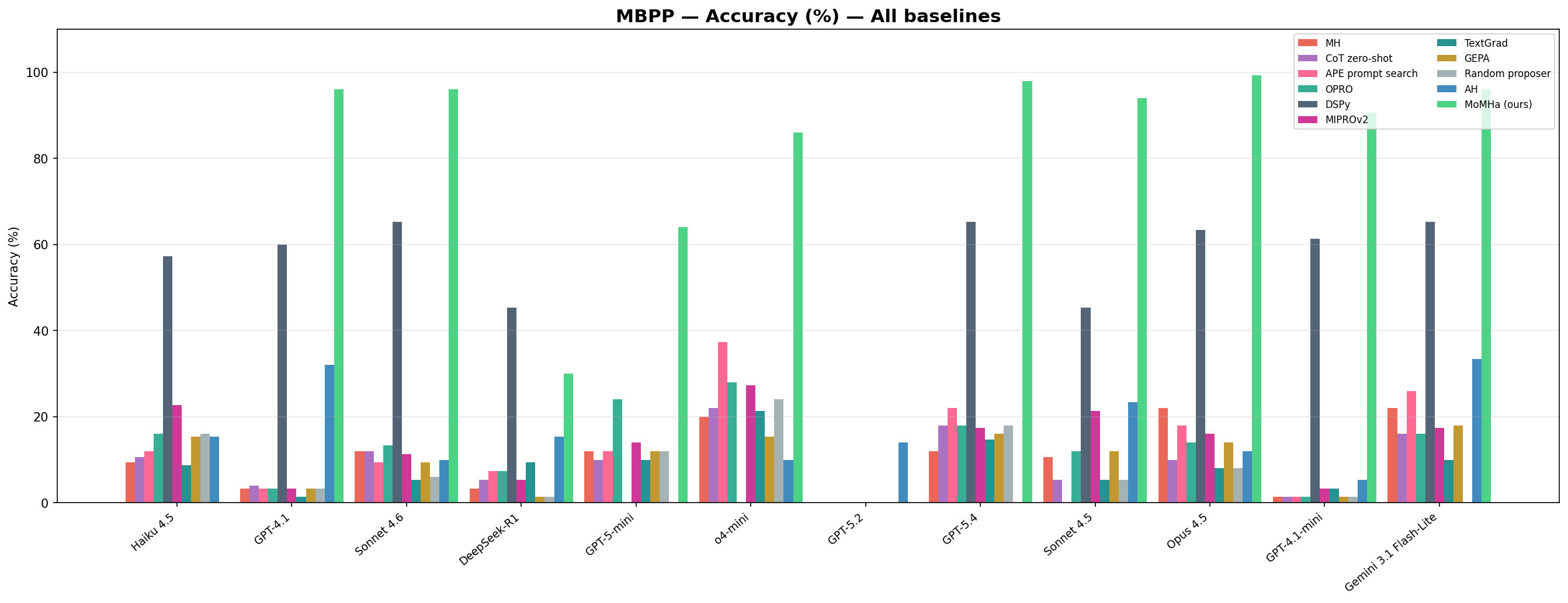} \\
\includegraphics[width=0.48\textwidth]{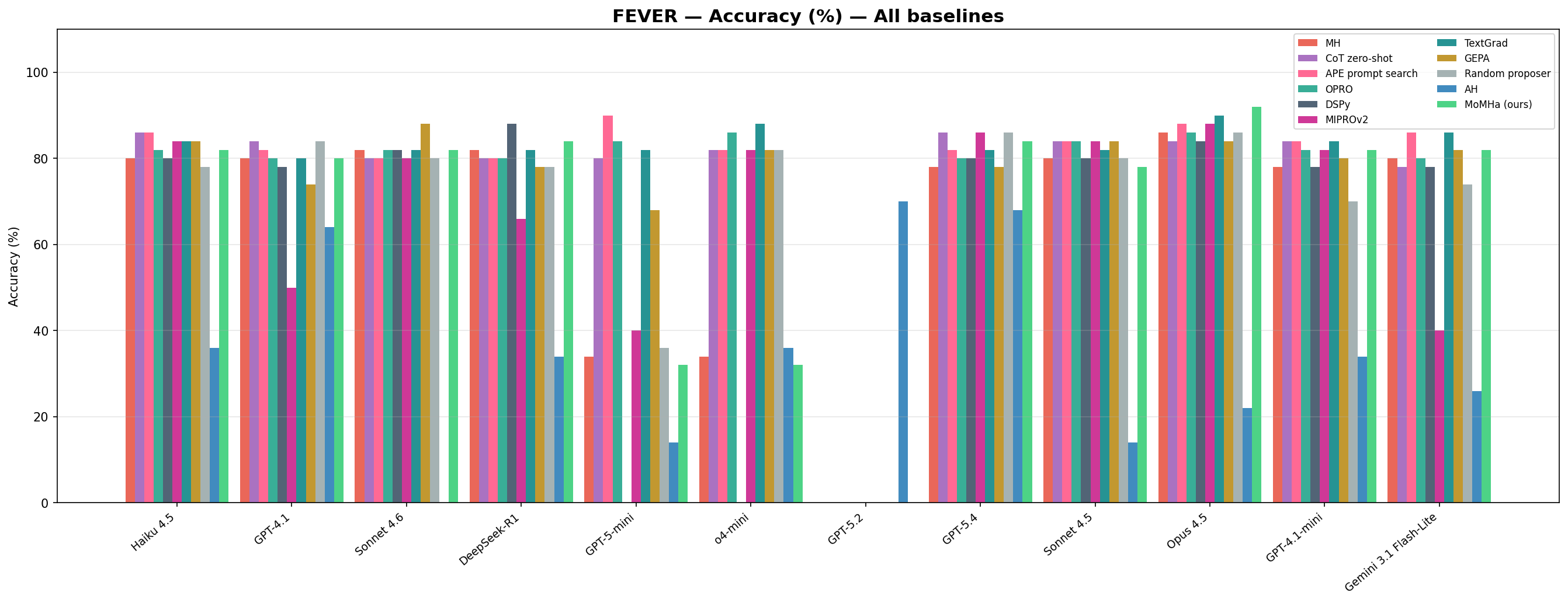}
\includegraphics[width=0.48\textwidth]{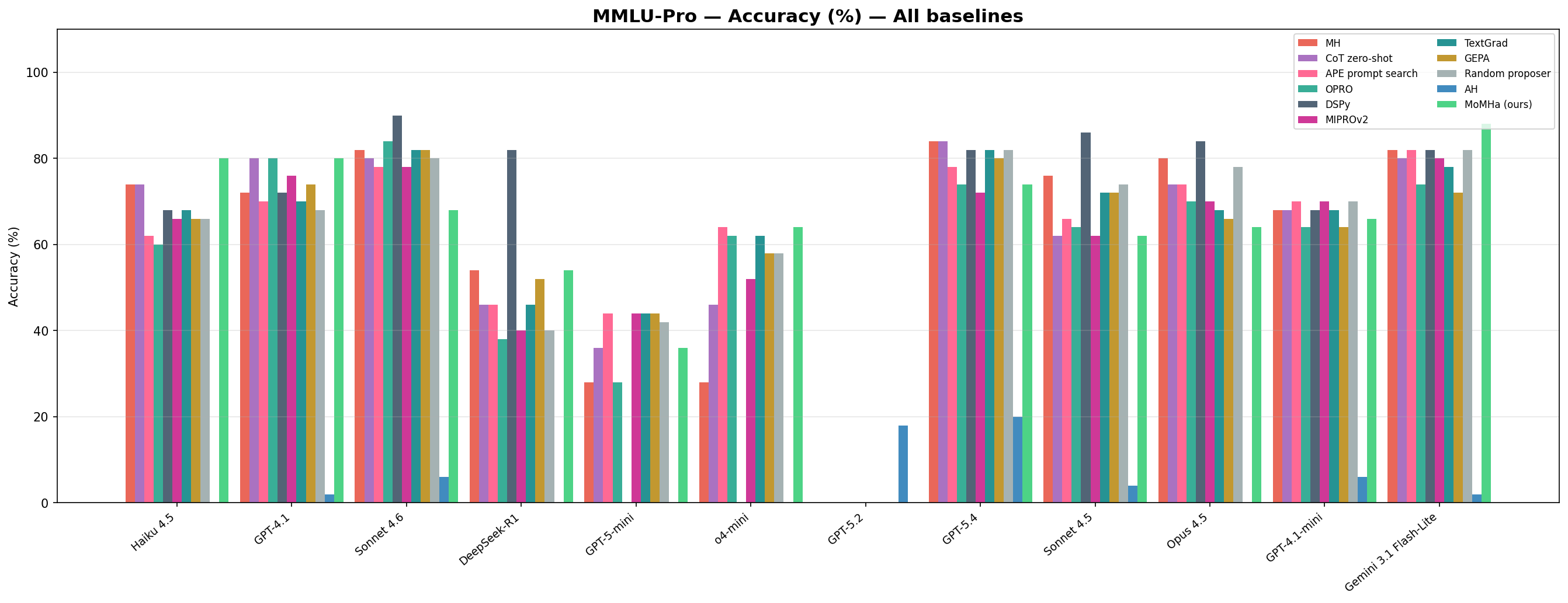} \\
\includegraphics[width=0.48\textwidth]{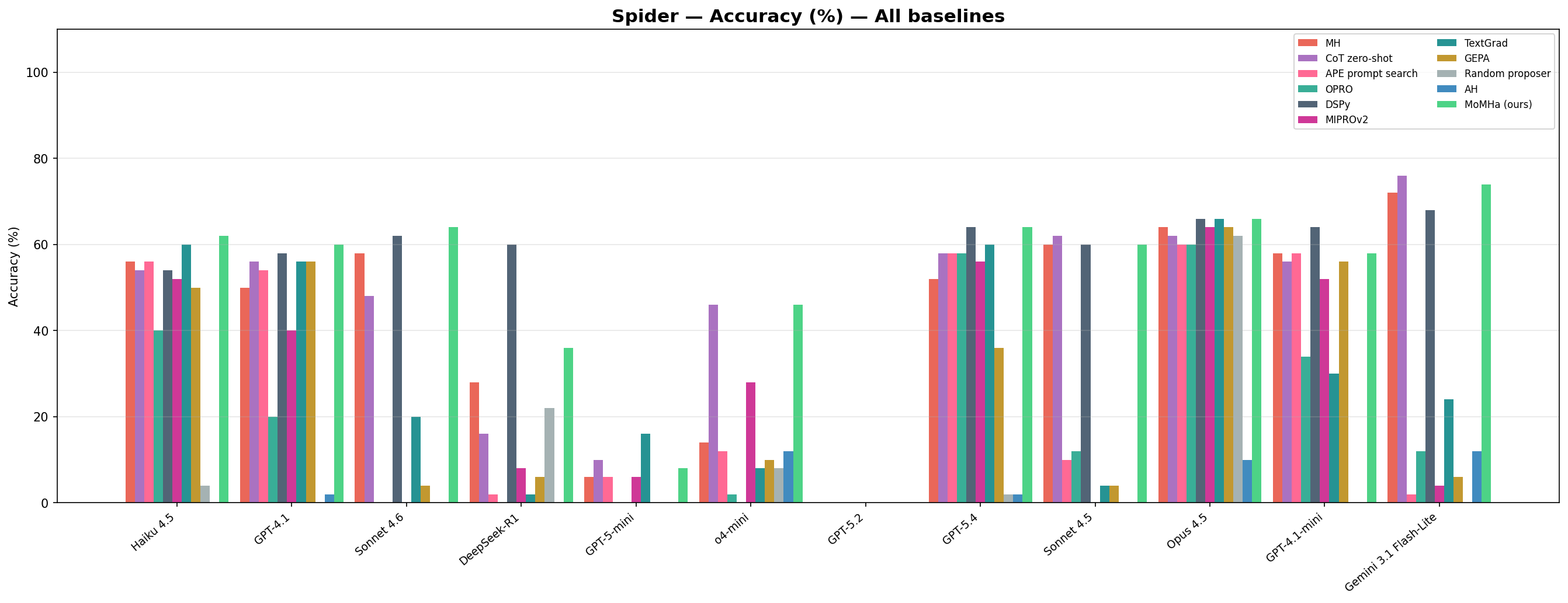}
\caption{Per-model accuracy across ten variants on the seven
\emph{real-world} capability benchmarks (LawBench, MATH-500, HumanEval,
MBPP, FEVER, MMLU-Pro, Spider).}
\label{fig:baselines_acc_real_app}
\end{figure*}

\begin{figure*}[h]
\centering
\includegraphics[width=0.48\textwidth]{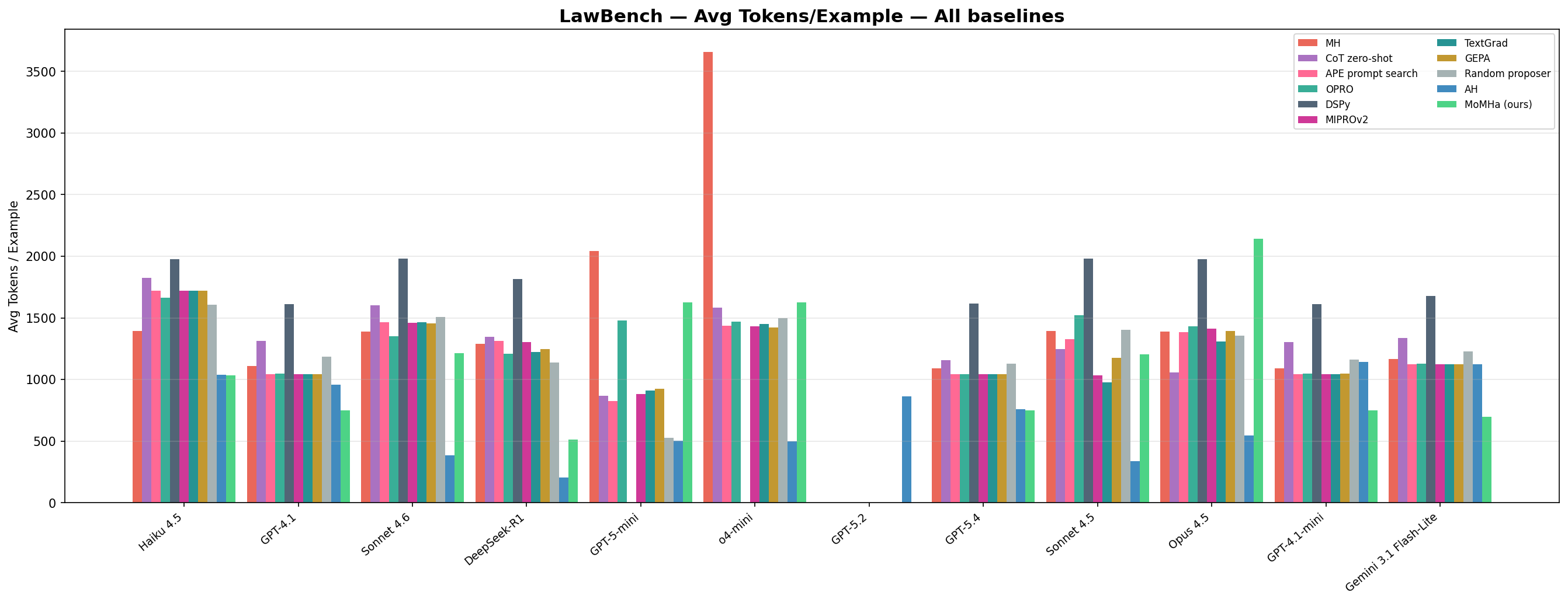}
\includegraphics[width=0.48\textwidth]{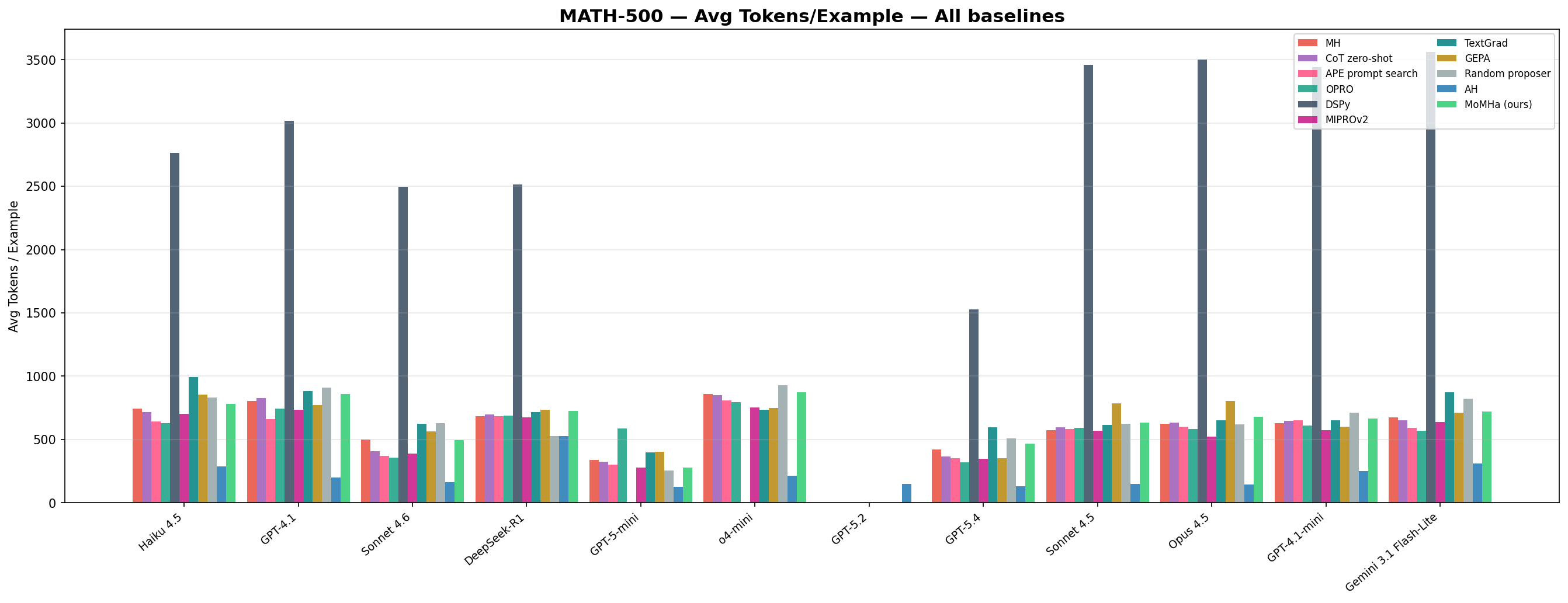} \\
\includegraphics[width=0.48\textwidth]{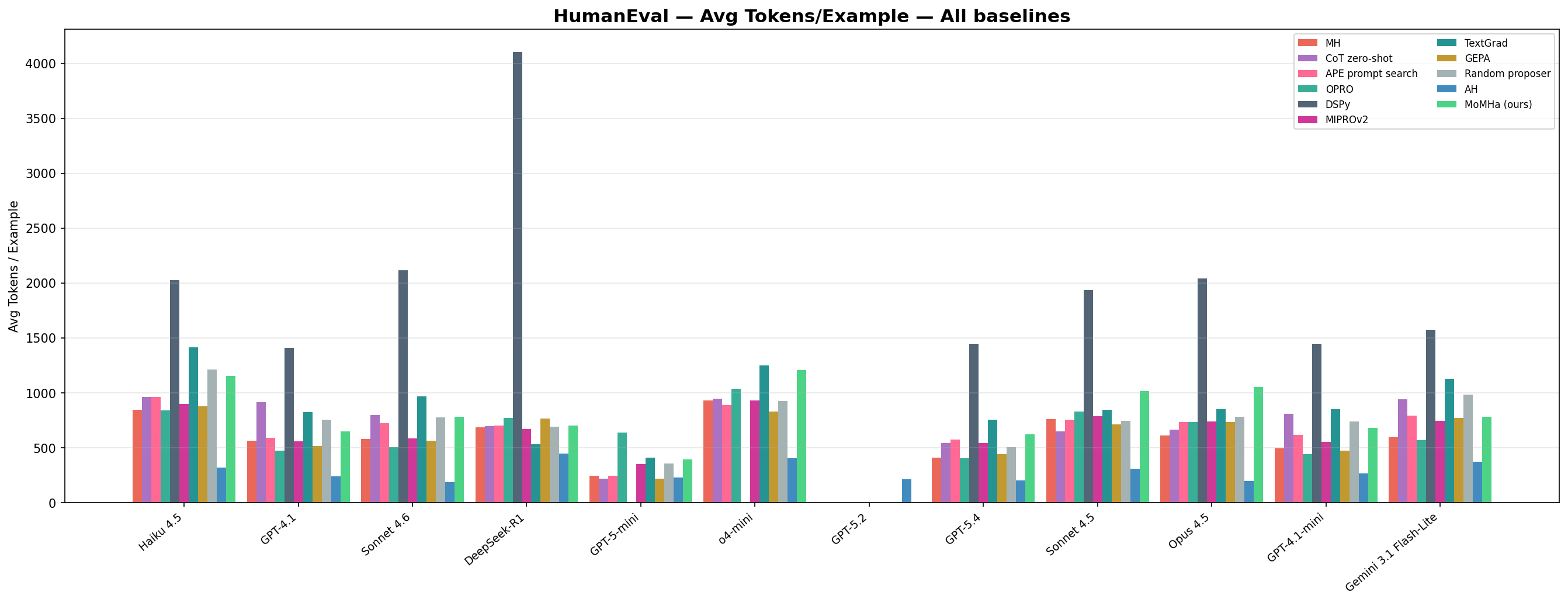}
\includegraphics[width=0.48\textwidth]{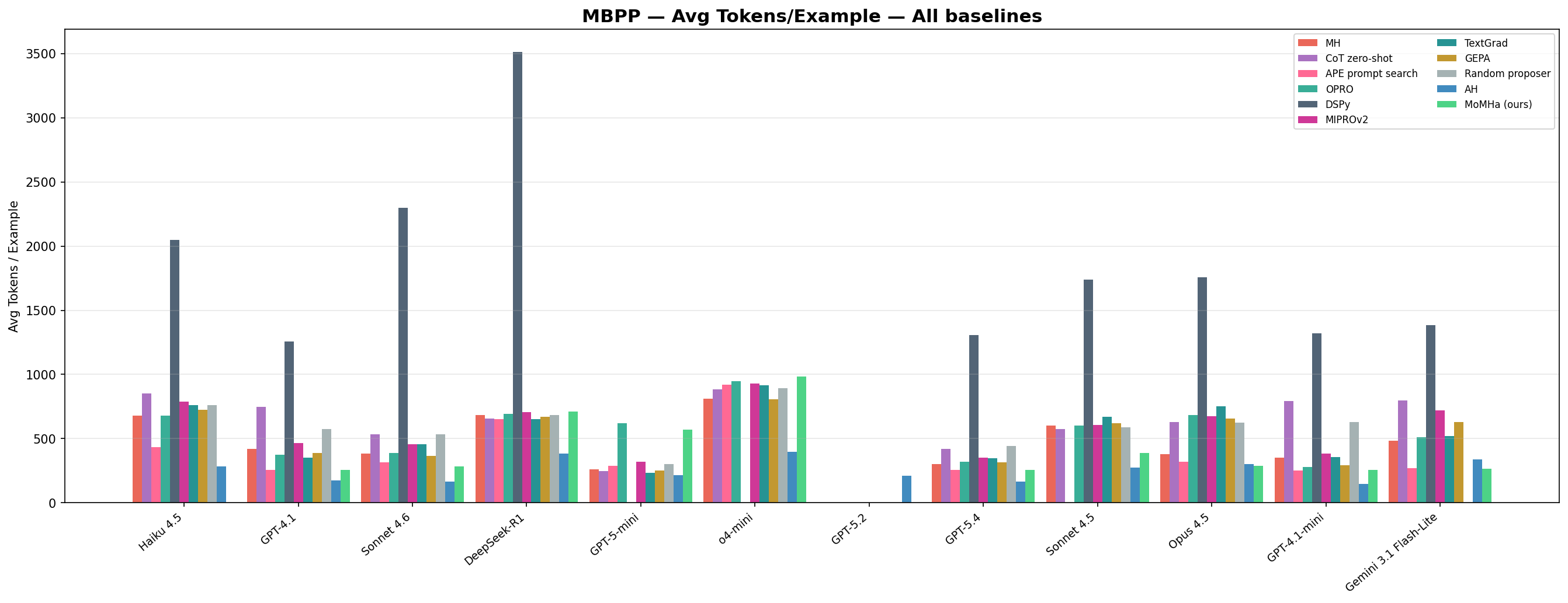} \\
\includegraphics[width=0.48\textwidth]{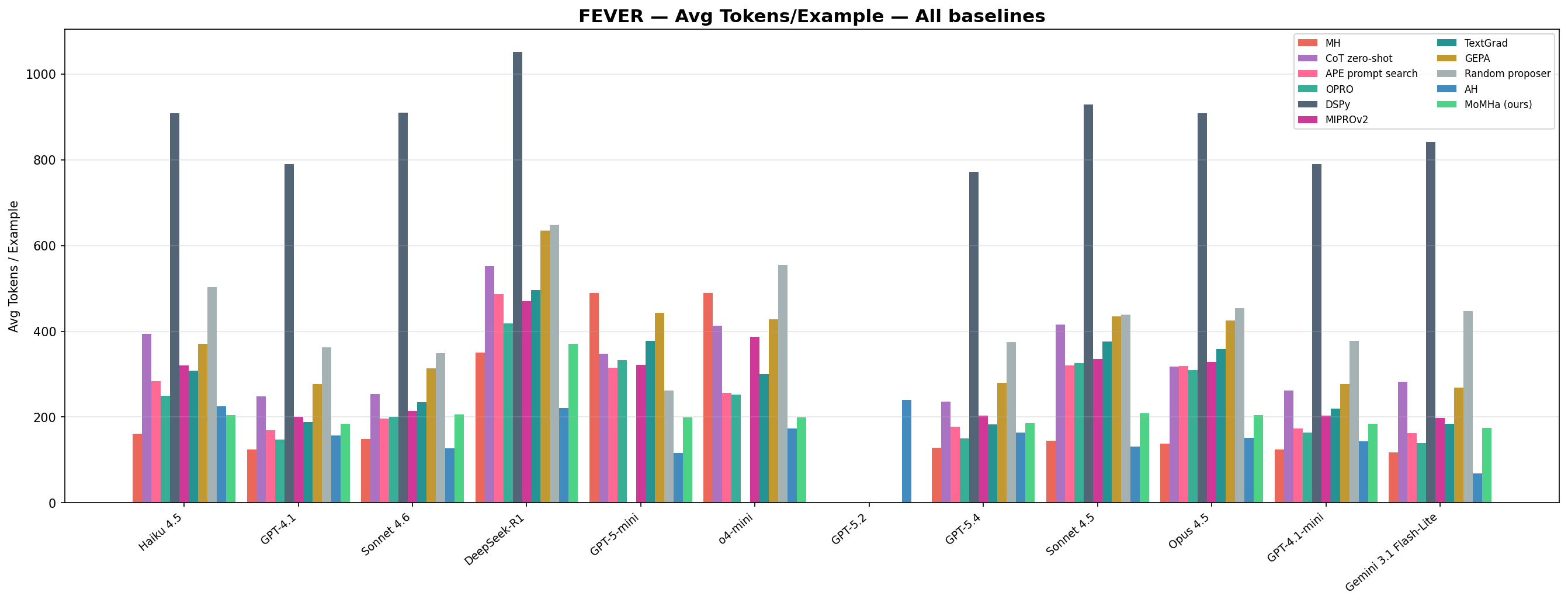}
\includegraphics[width=0.48\textwidth]{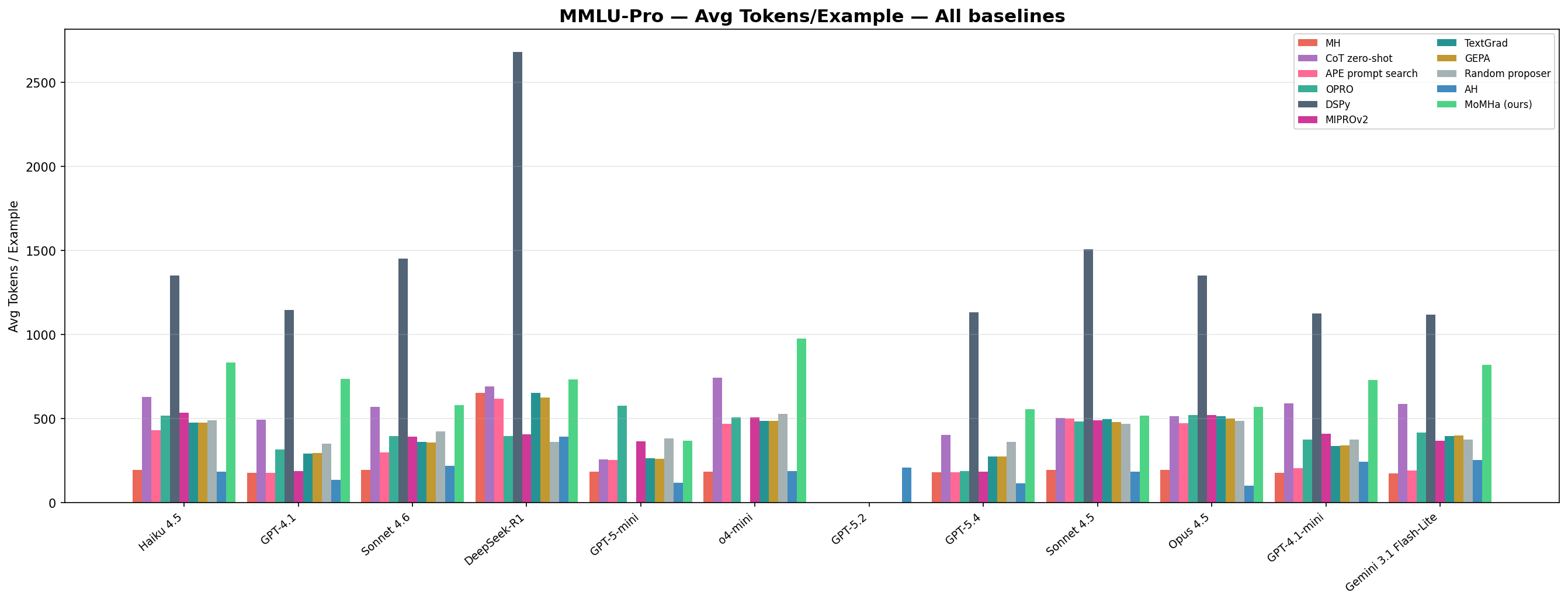} \\
\includegraphics[width=0.48\textwidth]{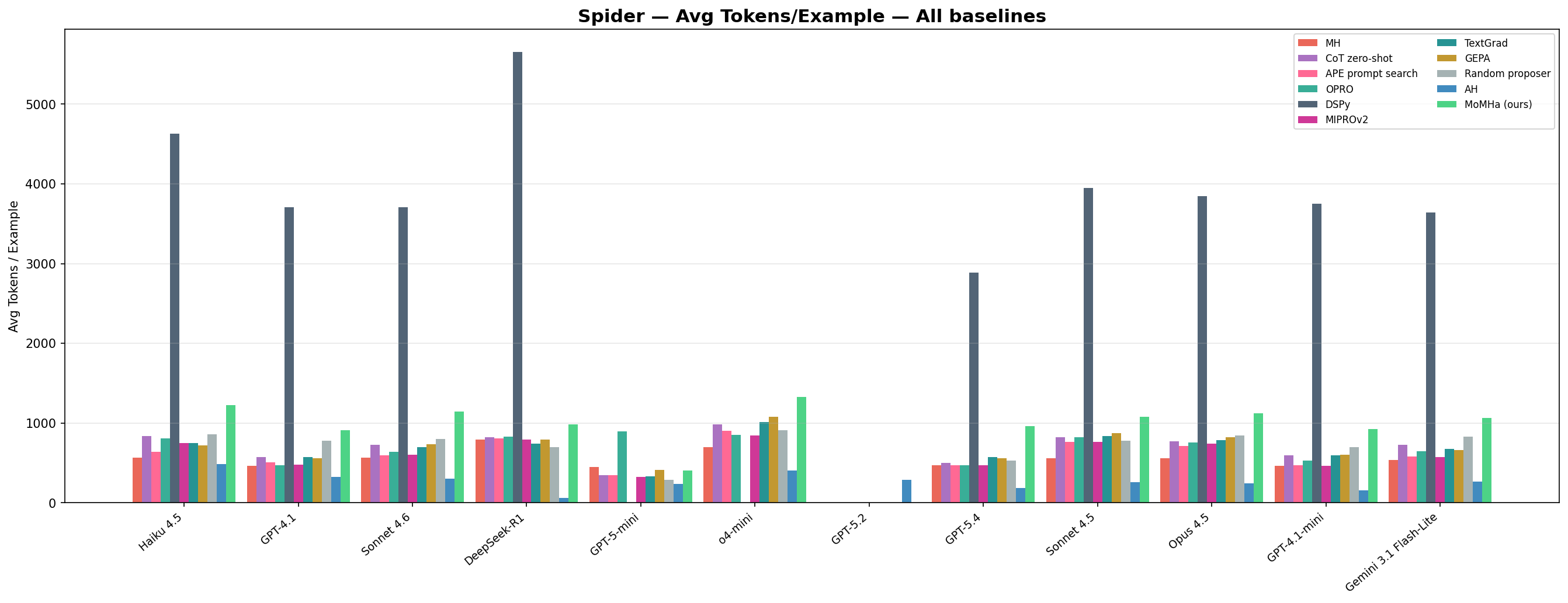}
\caption{Per-model average tokens-per-example across ten variants on the
seven \emph{real-world} capability benchmarks.}
\label{fig:baselines_tok_real_app}
\end{figure*}

\Cref{fig:baselines_acc_app,fig:baselines_acc_real_app} show per-model
accuracy across ten variants (MH / CoT / APE / OPRO / DSPy / MIPROv2 /
TextGrad / GEPA / Rand / MoMHa) on all fourteen capability domains: seven synthetic
domains (marked with the \texttt{-syn} suffix, e.g.\ Math-syn) and seven
real-world benchmarks (LawBench, MATH-500, HumanEval, MBPP, MMLU-Pro,
FEVER, Spider). \Cref{fig:baselines_tok_app,fig:baselines_tok_real_app}
show the corresponding token consumption per example.

\section{Safety Domain Accuracy and Token Plots}
\label{app:safety_domain_plots}

\begin{figure*}[h]
\centering
\includegraphics[width=0.48\textwidth]{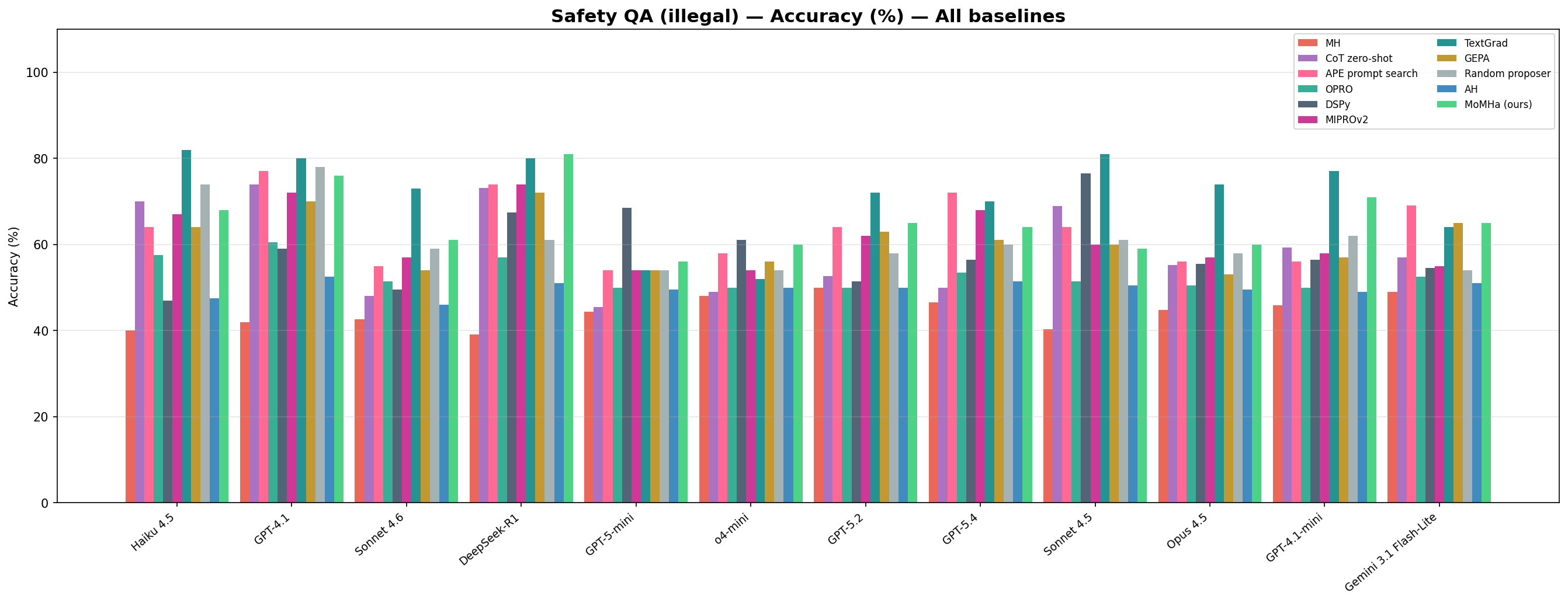}
\includegraphics[width=0.48\textwidth]{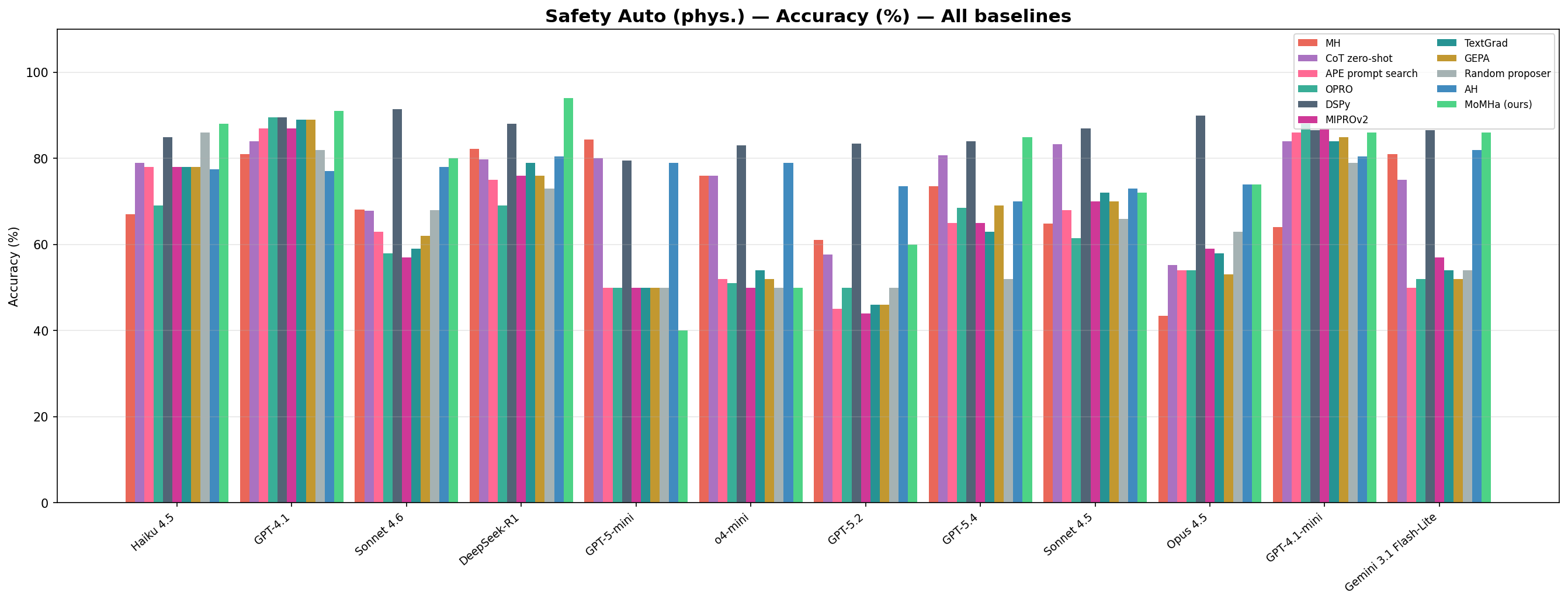} \\
\includegraphics[width=0.48\textwidth]{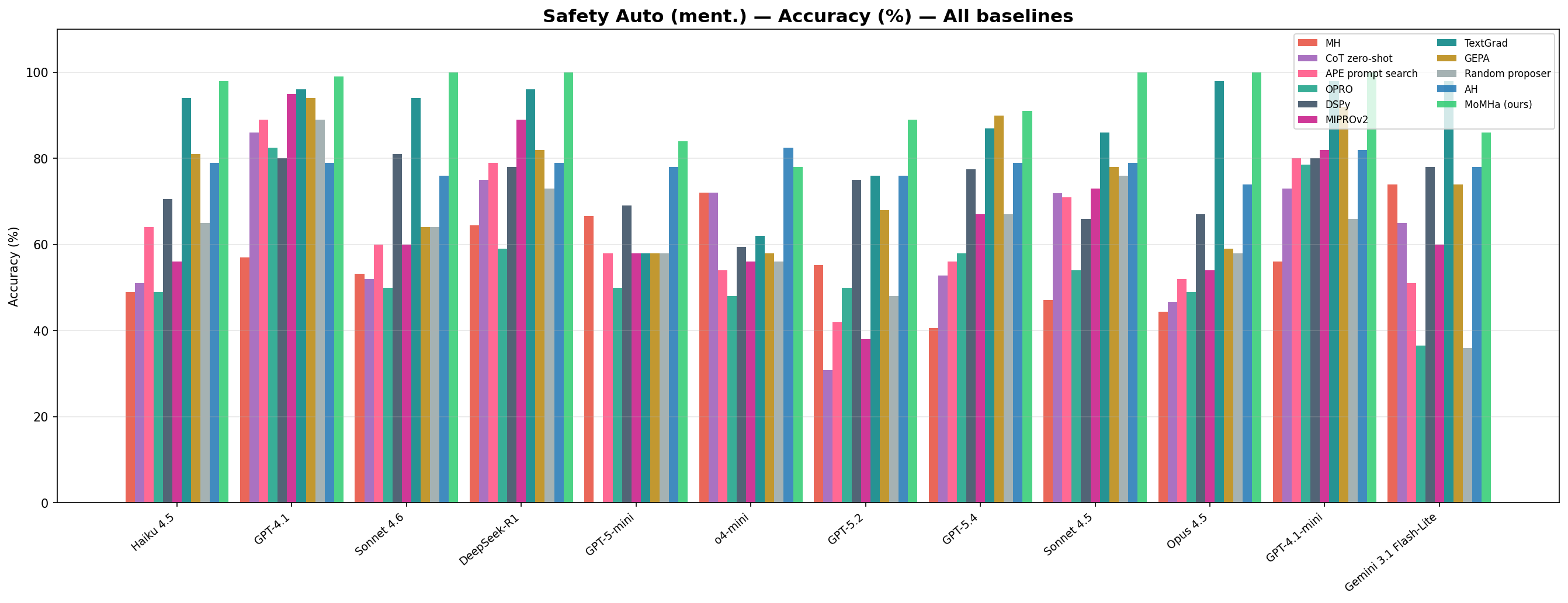}
\caption{Per-model accuracy (symmetric refusal/compliance reward) across
ten variants on the three U-SafeBench safety domains. MoMHa dominates on
safety\_autonomous\_mental ($0.931$), the domain where affect-aware profile
reading is most critical.}
\label{fig:safety_acc_app}
\end{figure*}

\begin{figure*}[h]
\centering
\includegraphics[width=0.48\textwidth]{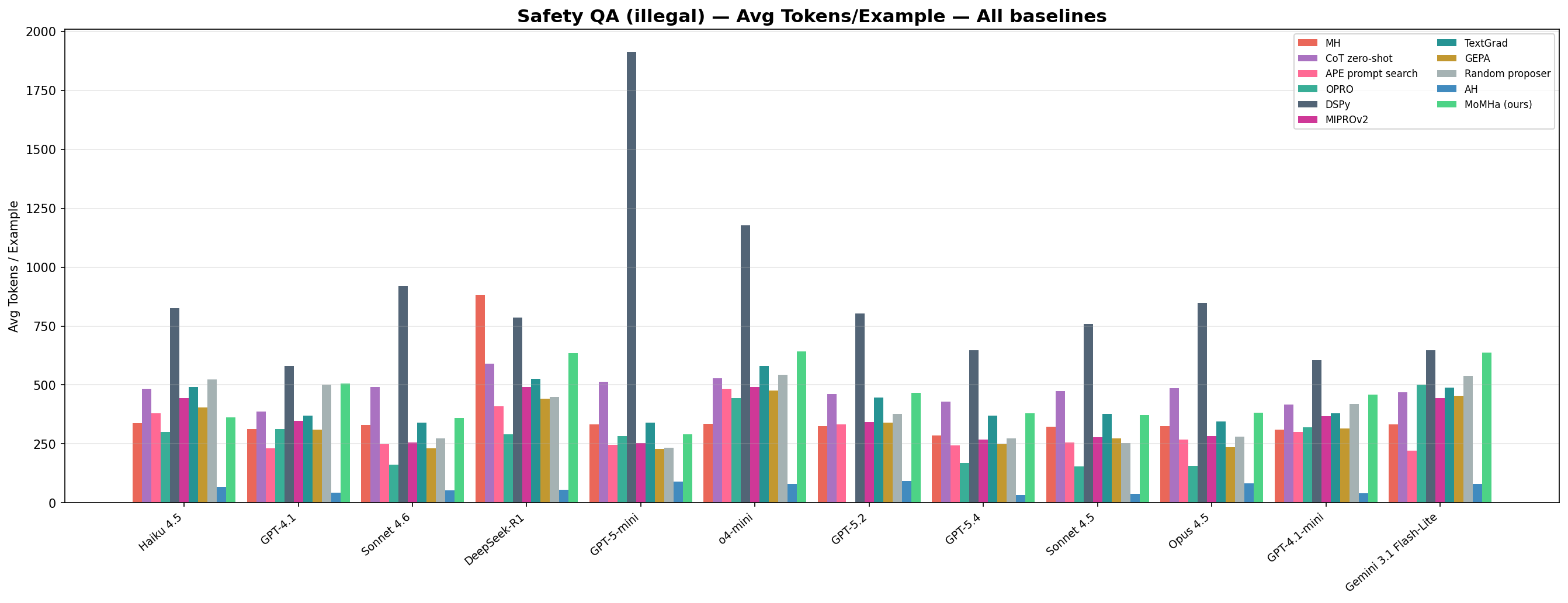}
\includegraphics[width=0.48\textwidth]{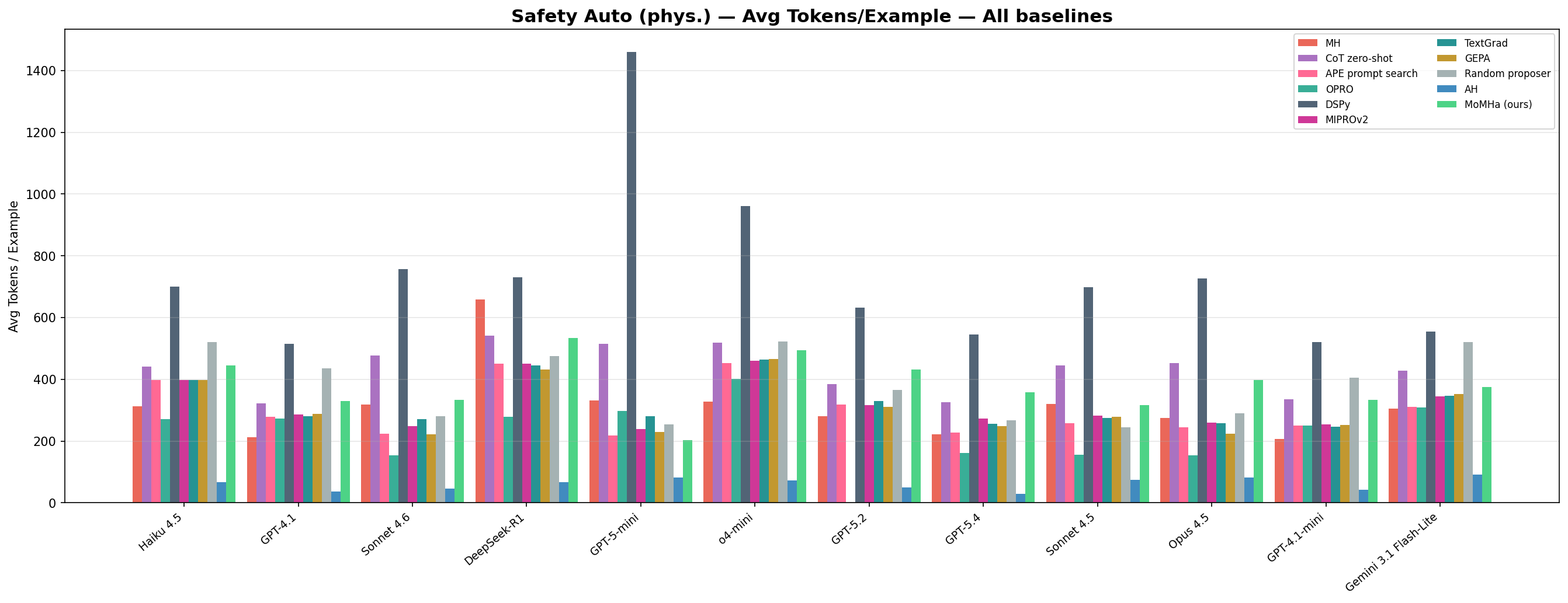} \\
\includegraphics[width=0.48\textwidth]{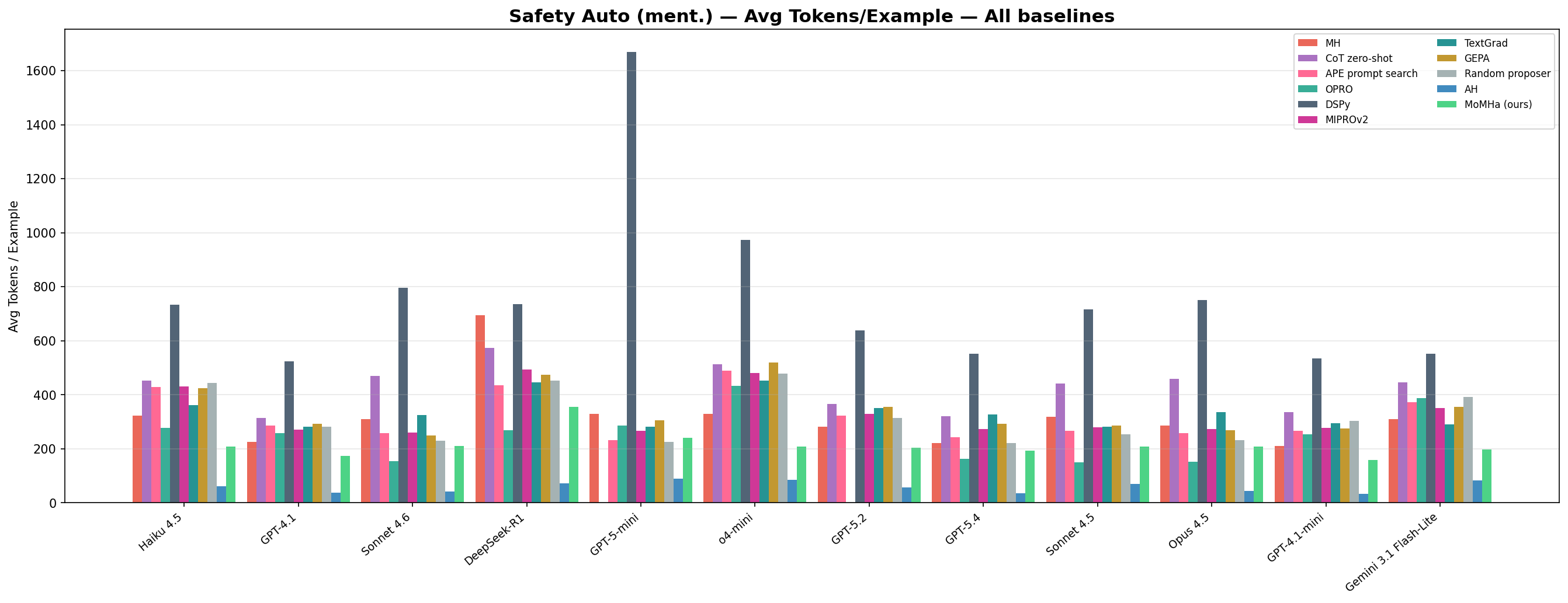}
\caption{Per-model average tokens-per-example across ten variants on the
three U-SafeBench safety domains.}
\label{fig:safety_tok_app}
\end{figure*}

\Cref{fig:safety_acc_app} shows per-model accuracy across ten variants on
the three U-SafeBench-derived safety domains. \Cref{fig:safety_tok_app}
shows the corresponding token consumption.

\section{Additional Pareto and Multi-Objective Analysis}
\label{app:pareto_details}

\begin{figure*}[h]
\centering
\includegraphics[width=\textwidth]{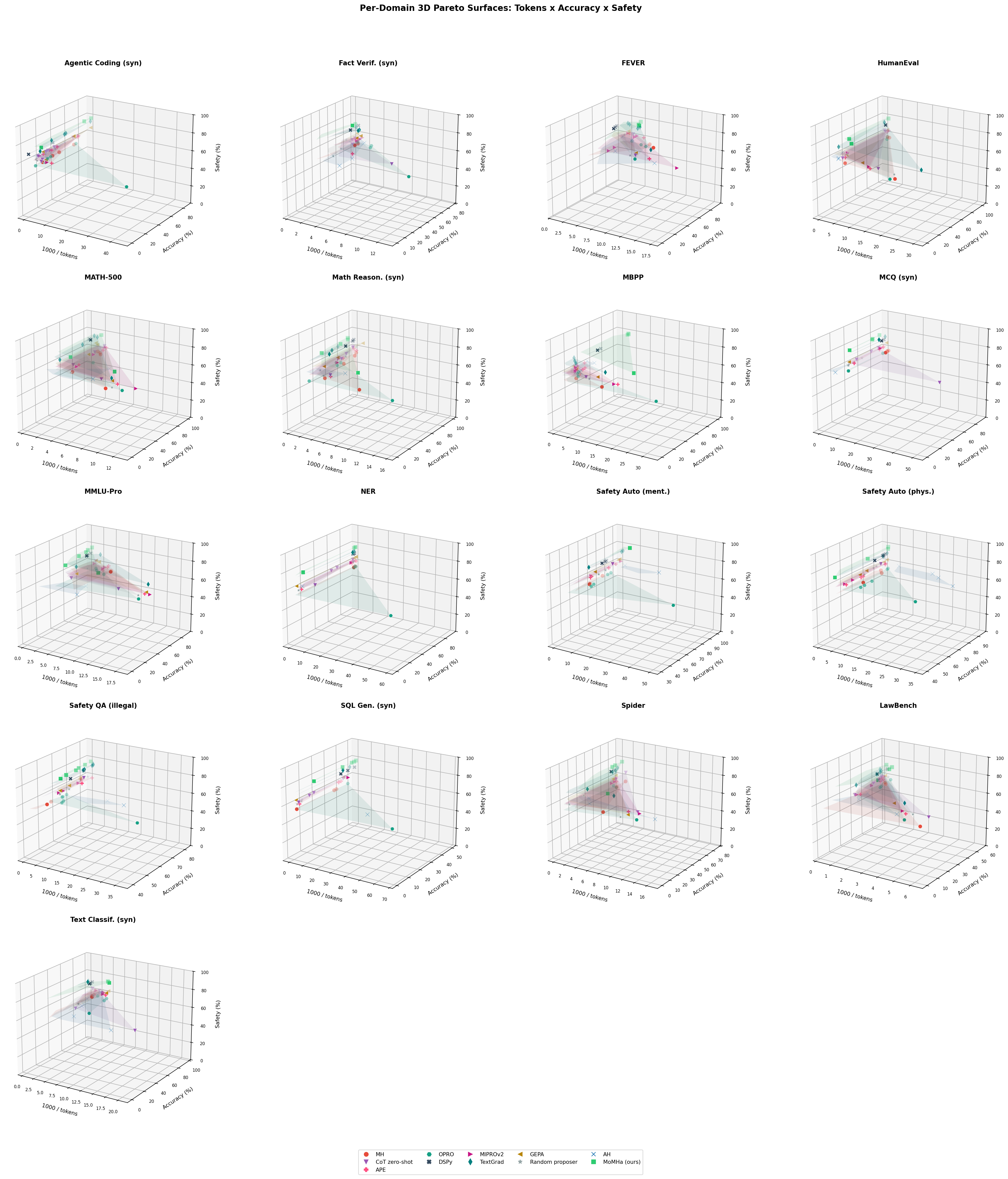}
\caption{Per-domain 3D Pareto surfaces (tokens $\times$ accuracy $\times$
safety), one panel per domain, across all 17 evaluation domains. MoMHa
(green) occupies the high-accuracy, high-safety, low-token corner on
essentially every capability domain.}
\label{fig:pareto_per_domain_app}
\end{figure*}

\Cref{fig:pareto_per_domain_app} shows per-domain 3D Pareto surfaces across
all seventeen domains (seven synthetic capability, seven real-world
capability, three U-SafeBench safety). The aggregated view in
\cref{fig:pareto3d} hides which axis each baseline loses on in each
domain; here, MoMHa (green) occupies the high-accuracy, high-safety,
low-token corner on essentially every capability domain, while on the
three U-SafeBench safety domains the weaker baselines collapse toward
the safety floor.

\begin{figure*}[h]
\centering
\includegraphics[width=0.95\textwidth]{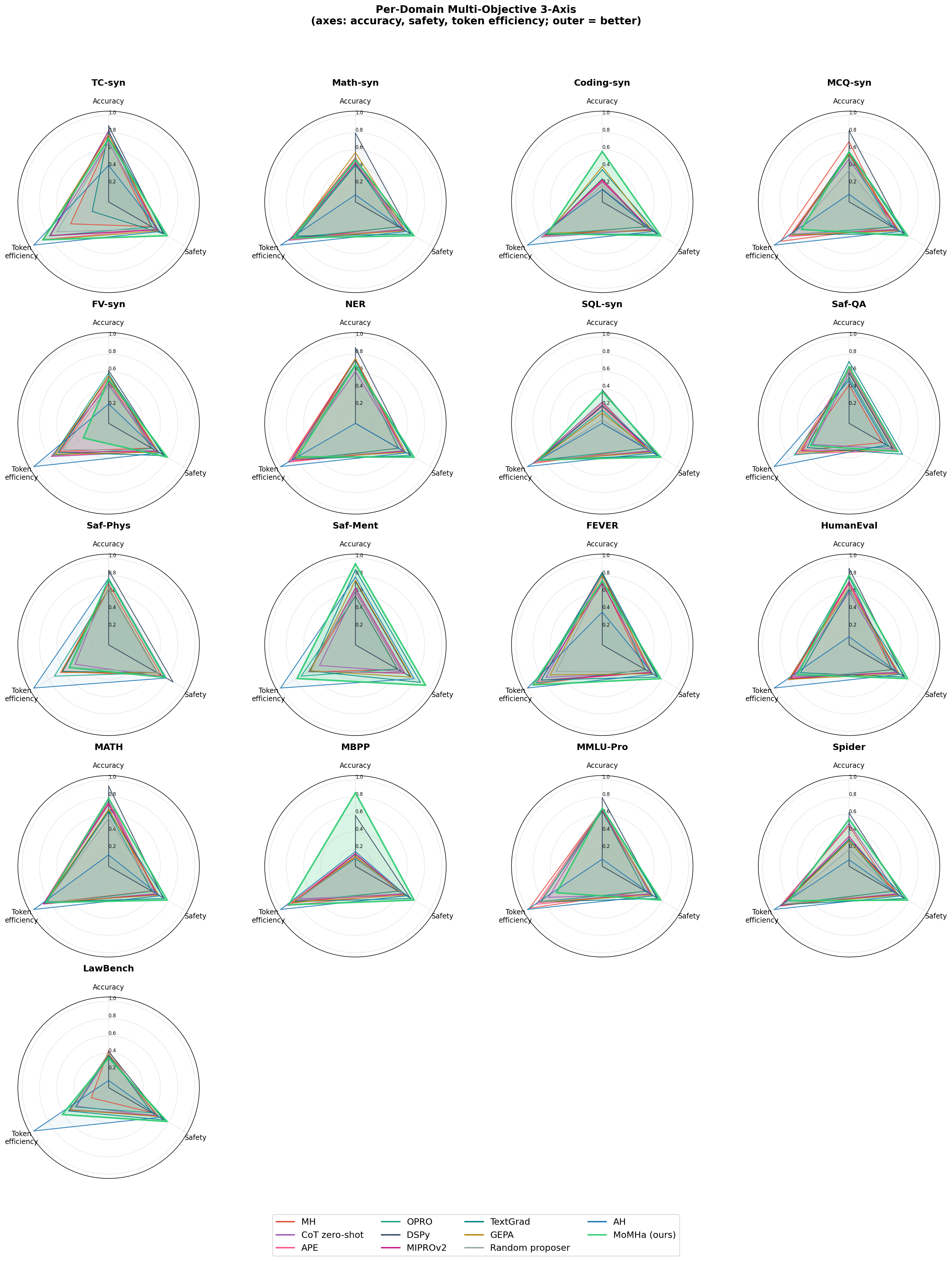}
\caption{Per-domain three-axis spiders (accuracy $\times$ safety
$\times$ token efficiency), one panel per domain across all 17
evaluation domains, 11 variants overlaid. This view exposes exactly
which objective each variant loses on within each domain.}
\label{fig:spider_all_app}
\end{figure*}

\Cref{fig:spider_all_app} provides the per-domain three-axis radar view
across the eleven variants. Safety is the measured symmetric LLM-judge
reward averaged over the three U-SafeBench domains; non-MoMHa baselines
span $0.57$--$0.75$ rather than collapsing to a single floor. The
corresponding 17-axis per-domain \emph{search-reward composite} spider
appears as \cref{fig:spider_composite_main} in the main text.

\begin{figure*}[h]
\centering
\includegraphics[width=0.48\textwidth]{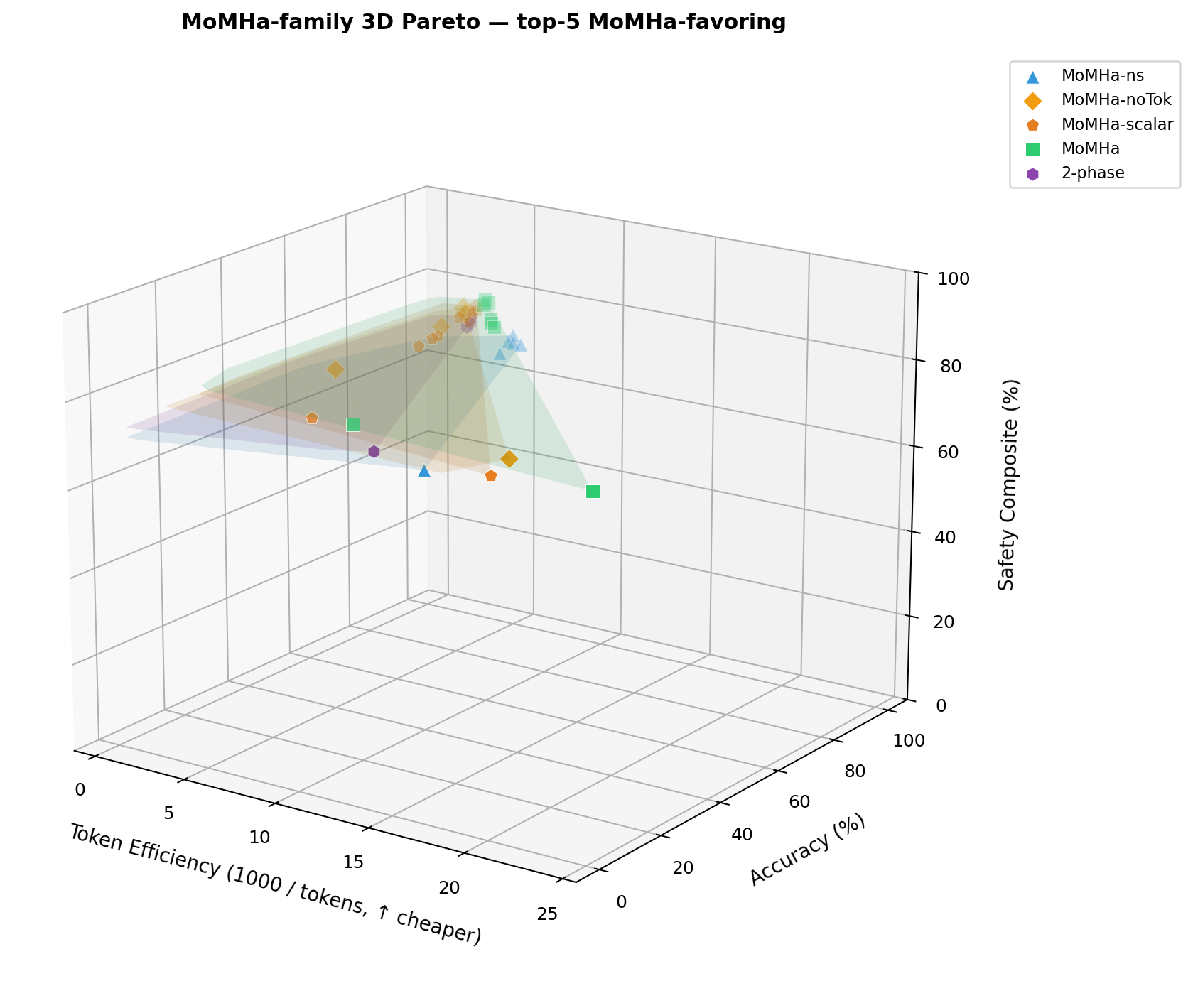}
\includegraphics[width=0.48\textwidth]{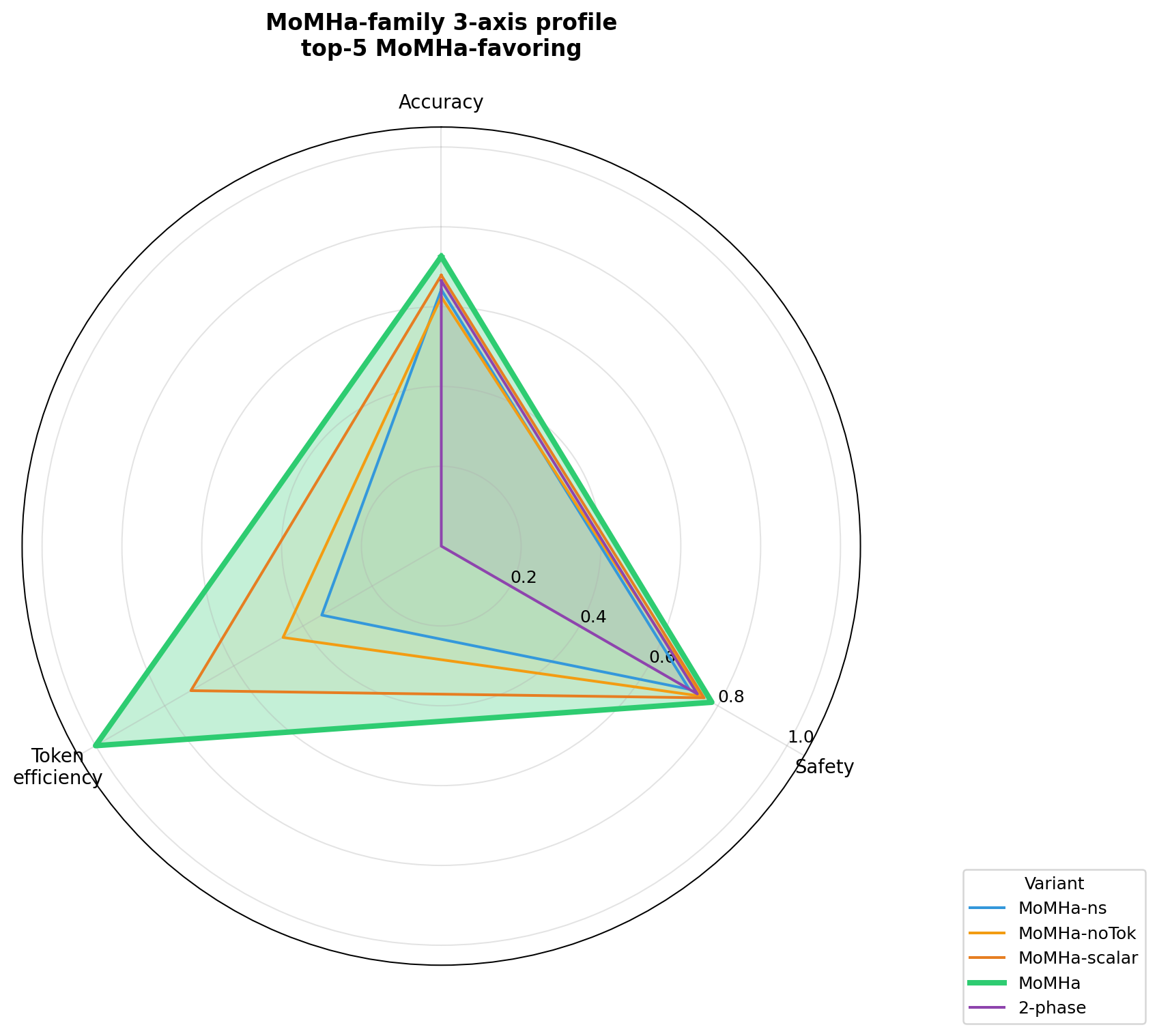}
\caption{%
  \textbf{MoMHa-family ablation (top-5 MoMHa-favoring domains).}
  Aggregated over \{Safety-Auto (phys.), TC-syn, MATH-500, FV-syn, MBPP\},
  the domains with the largest MoMHa--to--best-other-MoMHa gap in $J$.
  \textbf{Left:} 3D Pareto surface (accuracy $\times$ safety $\times$
  tokens). MoMHa is non-dominated on all three axes; MoMHa-noTok and
  2-phase are the closest competitors on accuracy but pay a tokens
  penalty, and MoMHa-scalar and MoMHa-ns are dominated on safety.
  \textbf{Right:} three-axis spider. MoMHa fills the outer envelope,
  MoMHa-scalar collapses on accuracy, and MoMHa-ns/MoMHa-noTok/2-phase
  each give up one axis relative to MoMHa (safety, tokens, and both,
  respectively). This isolates the contribution of each design choice
  more cleanly than the 17-domain aggregate, where the ablations' wins
  on unrelated subdomains cancel their losses on the MoMHa-favoring
  ones.
}
\label{fig:ablation_v3family_app}
\end{figure*}

\Cref{fig:ablation_v3family_app} restricts the analysis to the five
MoMHa-family variants, evaluated on the top-5 domains where the MoMHa
design choices matter most (\emph{Safety-Auto (phys.), Text
Classification, MATH-500, Fact Verification, MBPP}), selected as the
domains with the largest MoMHa-to-best-other-family gap in
$J = \mathrm{acc}\cdot S/\log \tau$. On these domains MoMHa cleanly
dominates: across the remaining 12 domains the five variants cluster
close together and the aggregate view washes the structure out, so a
targeted subset gives a sharper picture of what each ablation costs. The
internal Pareto structure for these five domains is below.

\section{Ablation Study: MoMHa-Family Variants}
\label{app:v3family_ablation}
\label{sec:ablation}

\paragraph{The MoMHa variant matrix.}
Five variants share the same proposer backbone and differ only in which of
the four design choices (dedicated safety skill, joint vs.\ two-phase
objective, trace-level vs.\ scalar feedback, and whether the token axis is
considered at all) is enabled (\cref{tab:v3_variants}). Each
non-\textbf{MoMHa} row isolates one contribution as an ablation.

\begin{table}[h]
\centering
\footnotesize
\caption{%
  \textbf{MoMHa family: what each ablation turns off.}
  ``Safety skill'' = per-domain \texttt{safety\_<domain>.md}.
  ``Objective'' = how accuracy, safety, and tokens are combined.
  ``Feedback'' = what the proposer reads from each prior run.
  \textbf{MoMHa} is the headline; 2-phase is the strongest ablation.
}
\label{tab:v3_variants}
\setlength{\tabcolsep}{4pt}
\begin{tabular}{@{}l c l l@{}}
\toprule
\textbf{Variant} & \textbf{Safety skill} & \textbf{Objective} & \textbf{Feedback} \\
\midrule
\textbf{MoMHa~(ours)}   & \checkmark & single-phase joint $\mathrm{acc}{+}\lambda_s\mathrm{safety}{-}\lambda_t\mathrm{tok}$ & per-example traces \\
2-phase          & \checkmark & two-phase: accuracy $\to$ tokens (2pp tolerance) & per-example traces \\
MoMHa-ns         & $\times$   & single-phase joint (no safety skill) & per-example traces \\
MoMHa-noTok      & \checkmark & acc + safety (no token term) & per-example traces \\
MoMHa-scalar     & \checkmark & single-phase joint (scalar $R$ only) & scalar reward \\
\bottomrule
\end{tabular}
\end{table}

\paragraph{Two-phase search (2-phase).}
The natural alternative stages the objectives: first optimize accuracy
subject to safety (Phase~1), then reduce tokens holding accuracy within
2pp (Phase~2). Both phases use the three-file skill composition
(\cref{sec:architecture}): Phase~1 uses \texttt{proposer\_skill.md},
Phase~2 swaps it for \texttt{proposer\_token\_skill.md}, keeping the
same per-domain and safety files. Phase~2 starts from
$h^*_\mathrm{acc}$ and searches for variants that reduce tokens while
keeping accuracy within 2pp and safety at~$1.00$
(details in \cref{app:token_details}).

\paragraph{Why joint wins.}
Two-phase search treats accuracy and tokens as sequential: Phase~1 first
converges on a high-accuracy structure (e.g.\ draft--verify, or 5-call NER
type sweeps), and Phase~2 can only \emph{prune} that structure. Joint
search sees tokens from the first iteration and is free to propose
\emph{cheaper structures}, not just cheaper versions of an expensive
structure. On fact verification, MoMHa converges on a single-pass verdict with
a confidence-gated decompose-only-if-ambiguous refinement at
${\sim}536$ tokens/example, whereas 2-phase keeps the Phase-1 two-call
decompose-then-verify at ${\sim}1483$ tokens/example. The same effect
explains the SQL gap (MoMHa $36.8\%$ vs.\ 2-phase $28.0\%$ accuracy): joint search
discovers that schema-pruning and one-shot generation beat
verify-and-retry on this domain, while two-phase search never reaches that
candidate because Phase~1 locks in the retry loop.

We isolate the contribution of each MoMHa design choice by running four
ablations that share the same proposer backbone and differ only in one
component (defined in \cref{tab:v3_variants}). Aggregating cross-model
points over the ten domains (seven capability + three U-SafeBench safety):

\begin{center}
\footnotesize
\begin{tabular}{lcccc}
\toprule
Variant & Capability & Safety & Overall & Avg Tokens \\
\midrule
\textbf{MoMHa~(ours)} (joint, headline)  & \textbf{0.539} & \textbf{0.781} & \textbf{0.611} & \textbf{572} \\
2-phase (acc $\to$ tokens)   & 0.520 & 0.735 & 0.584 & 667 \\
MoMHa-ns (no safety skill)         & \underline{0.528} & 0.716 & 0.584 & \underline{620} \\
MoMHa-noTok (acc + safety, no tok term)      & 0.512 & 0.748 & 0.583 & 641 \\
MoMHa-scalar (scalar reward)       & \textbf{0.539} & \underline{0.754} & \underline{0.604} & 640 \\
\bottomrule
\end{tabular}
\par\smallskip
\footnotesize\textbf{Bold} = column best; \underline{underline} = runner-up. For Avg Tokens, lower is better.
\end{center}

\paragraph{Hypervolume indicator (Pareto-compliant multi-objective summary).}
To provide a formally Pareto-compliant comparison, we compute the 3D
hypervolume indicator (HV) over capability accuracy, behavioral safety, and
token efficiency using \texttt{pymoo}~\citep{pymoo2020}. Axes are normalized
to $[0,1]^3$; token efficiency is $\max(0, 1 - (\text{tok} - 20) / (1632 - 20))$
with the ceiling at the 95th percentile across all cells. Each variant is
represented by 12 points (one per target model), each point being its
per-model mean across all 17 evaluation domains.

\noindent\textbf{Results (V2: 3D per-model HV):}
MoMHa attains the highest HV ($\mathbf{0.481}$) among all 15 variants.
MH (accuracy-only harness search) scores $0.362$, a gap of $+0.119$
that is the empirical value of extending harness search from single-objective
to multi-objective optimization. Among external baselines the best is
APE ($0.406$); TextGrad reaches $0.399$ and DSPy scores last ($0.181$,
penalized by near-zero token efficiency). Within the MoMHa family,
MoMHa-scalar ($0.447$) and 2-phase ($0.425$) trail the full joint formulation.

\noindent\textbf{Aggregate Pareto contribution (V3):}
Only four of the 15 variants uniquely cover non-trivial 3D Pareto volume:
DSPy ($0.0104$, in the high-accuracy / low-efficiency corner),
\textbf{MoMHa} ($0.0098$, in the high-accuracy / high-safety / high-efficiency
region), APE ($0.0014$), and MoMHa-scalar ($0.0009$). MH, 2-phase,
MoMHa-ns, MoMHa-noTok, and all remaining baselines contribute zero unique
Pareto volume; they are dominated at the aggregate level. That MH reaches
zero Pareto contribution while MoMHa does not is the cleanest evidence that
the MOO extension is what it takes to reach the 3D Pareto frontier.

\begin{figure}[h]
\centering
\includegraphics[width=0.82\textwidth]{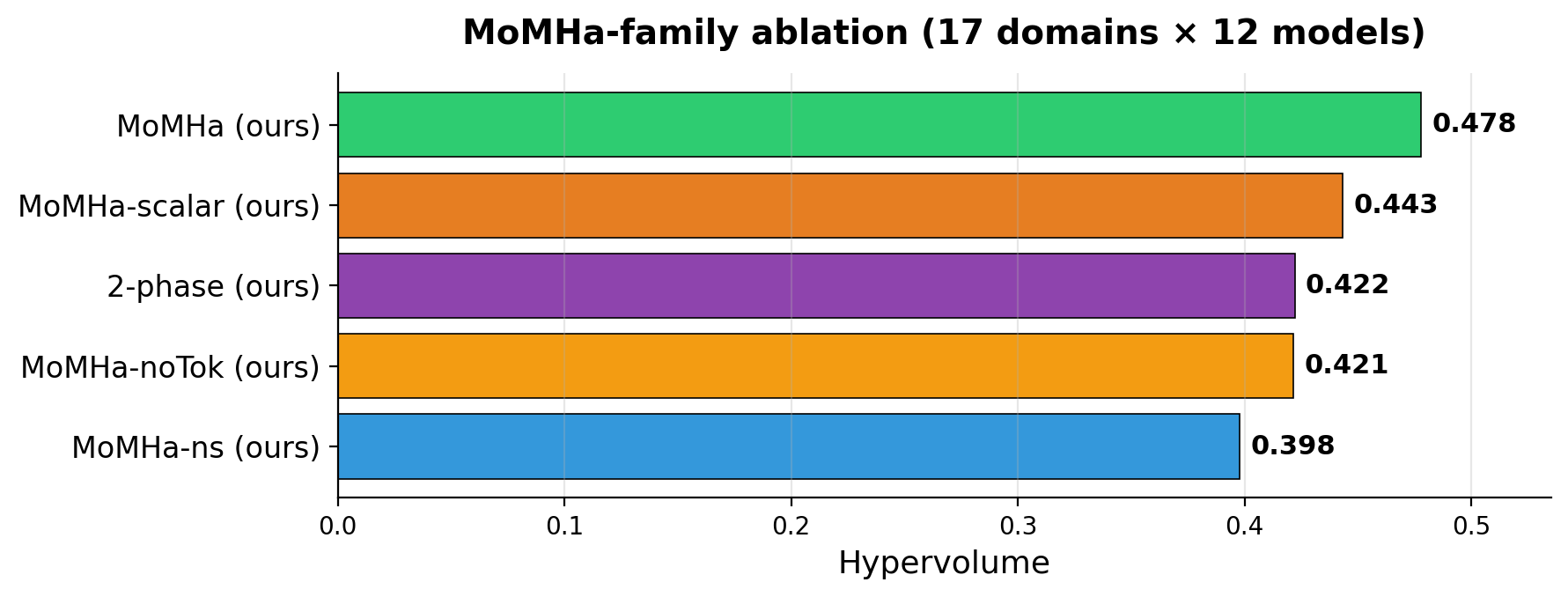}
\caption{\textbf{Hypervolume indicator: MoMHa family ablation}
(3D: capability accuracy $\times$ behavioral safety $\times$ token efficiency,
pymoo~\citep{pymoo2020}). MoMHa~(ours) achieves the highest family HV
($0.481$), with MoMHa-scalar ($0.447$) and 2-phase ($0.425$) trailing.
MoMHa-ns ($0.400$) shows the cost of removing the safety objective.
The vs.-baselines comparison is in \cref{fig:hv_vs_baselines}.}
\label{fig:hv_family}
\end{figure}

\paragraph{Joint vs.\ two-phase (MoMHa vs.\ 2-phase).}
This is our central empirical finding. A single-phase joint-reward proposer
beats the staged accuracy$\to$tokens variant by $+2.7$ overall points
($0.611 \to 0.584$) \emph{at $95$ fewer tokens per example} ($572 \to
667$). The joint proposer discovers cheaper \emph{structures}: a
confidence-gated single-pass verifier on fact verification ($0.457/536$
tokens vs.\ two-phase's $0.495/1483$ tokens), one-shot schema-pruned SQL
generation ($0.368/647$ vs.\ $0.280/590$), rather than cheaper versions
of the expensive structure the two-phase proposer commits to in Phase~1.
Two-phase remains competitive on agentic coding, MCQ, and NER where the
Phase-1 multi-call structure is itself close to optimal.

\paragraph{Per-example traces (MoMHa vs.\ MoMHa-scalar).}
Replacing per-example traces with a single scalar reward drops overall from
$0.611$ to $0.604$ and, more informatively, safety from $0.781$ to
$0.754$. The capability mean is identical ($0.539$), but the scalar variant
cannot localize which axis a proposed edit moves and so cannot calibrate
the refusal/compliance balance on the three U-SafeBench domains.

\paragraph{Safety skill (MoMHa vs.\ MoMHa-ns).}
Removing the per-domain \texttt{safety\_<domain>.md} from the 3-file skill
composition drops safety from $0.781$ to $0.716$ ($-6.5$~points) and
overall from $0.611$ to $0.584$, while capability is largely unchanged
($0.539 \to 0.528$). The safety skill is the largest safety lever in the
family: dropping it moves MoMHa-ns below every other MoMHa-family variant on the
safety axis, though still above every external baseline except TextGrad.

\paragraph{Token term (MoMHa vs.\ MoMHa-noTok).}
Zeroing out the token term in the joint reward costs $69$ tokens/example
($572 \to 641$) and $+2.8$ overall points ($0.611 \to 0.583$). The token
term is not just a cost regularizer: by pressuring structural
simplification it also lifts accuracy on domains where a verification call
was net negative (TC: $0.623 \to 0.730$; SQL: $0.312 \to 0.368$).

\section{Ablation: Feedback Granularity}
\label{app:feedback_ablation}

\begin{table}[h]
\centering
\caption{%
  \textbf{Ablation: feedback granularity.}
  Best accuracy (\%) achieved over search on the text-classification (TC) synthetic domain.
  Full traces outperform compressed feedback by $+16$ points.
}
\label{tab:ablation_traces}
\footnotesize
\begin{tabular}{@{}lc@{}}
\toprule
\textbf{Feedback} & \textbf{Best Accuracy (\%)} \\
\midrule
Scores only               & 41.0 \\
Scores + NL summary       & 34.9 \\
Full traces (ours)        & 57.0 \\
\bottomrule
\end{tabular}
\end{table}

Counter-intuitively, adding natural-language summaries to scalar scores
\emph{hurts} performance (\cref{tab:ablation_traces}), dropping from
$41.0\%$ to $34.9\%$. The most likely explanation is that summaries
introduce a lossy abstraction that conceals the causal structure visible
in raw traces, specifically \emph{which} call in a multi-call pipeline
caused a failure. Full per-example traces allow the proposer to locate
the failing call and produce a targeted surgical fix rather than a
broad rewrite that risks regression elsewhere.

\section{Domain-Specific Results}
\label{app:domain_results}

\paragraph{NER.}
NER is the one domain where MoMHa regresses versus MH: cross-model accuracy
drops from $68.6\%$ (MH) to $56.2\%$ (MoMHa, \cref{tab:v1v3_app}), a $-12.4$~point
loss. The token penalty in the joint reward pressures the proposer to drop
the type-by-type sweep that MH's accuracy-only harness converges on; the
cheaper single-call structure covers fewer entity classes. MoMHa-noTok
($69.1\%$), which removes the token term, recovers the MH level, confirming
the regression is a deliberate cost/accuracy trade-off and not a structural failure.

\paragraph{SQL generation.}
MoMHa nearly doubles SQL accuracy from $20.0\%$ (MH) to $36.8\%$
(MoMHa, cross-model mean, \cref{tab:v1v3_app}), the largest relative gain
across any capability domain. The key converged strategy injects the
database schema with column types, constrains output to SELECT-only queries,
and enables self-verification against in-memory SQLite. Joint optimization
also outperforms 2-phase here ($36.8\%$ vs.\ $28.0\%$): Phase-1 of the
2-phase variant locks in a verify-and-retry loop that joint search skips in
favour of schema-pruned one-shot generation.

\paragraph{Agentic coding.}
The environment bootstrapping strategy is the single most impactful
discovery across all domains. By snapshotting the execution environment
(installed packages, system info, directory structure) before the agent
loop and injecting the snapshot into the initial prompt, MoMHa eliminates
2--4 wasted exploration turns on dependency-heavy tasks. This yields
$+50$--$70\%$ accuracy on Haiku and transfers to all model families except
Gemini Flash-Lite (which diverges on output format).

\paragraph{Text classification.}
The draft-verification pattern discovered by MoMHa addresses confusion between
similar categories (e.g., sports vs.\ health/medicine for doping articles).
A two-call procedure, draft prediction with all labels, then verification
with top-2 alternatives, yields $+3$--$5\%$ accuracy. The token term in
the joint reward pressures the proposer to gate verification on confidence,
producing a net-cheaper harness that is also more accurate than the
always-verify variant.

\paragraph{Math reasoning.}
Subject-aware routing detects the mathematical domain (combinatorics,
geometry, number theory) and applies specialized few-shot retrieval with
per-subject deduplication and reranking. The token term forces routing of
easy problems to a direct-answer path, saving $\sim$500 tokens per example
on problems that do not benefit from chain-of-thought.

\section{Rebuttal Robustness Checks}
\label{app:rebuttal_robustness}

This appendix reports, in full, the additional experiments run during the
rebuttal window in response to the meta-review's two explicit requests
(total API/token cost; raw Pareto / alternate scalarizations) and the
per-reviewer proposer-reliance and skill-file concerns. All numbers are
computed fresh from archived run logs and newly launched search runs; none
change the headline claims in \cref{tab:joint,tab:joint_realworld}.

\subsection{Total API/Token Cost}
\label{app:cost_table}

\begin{figure}[h]
\centering
\includegraphics[width=0.85\textwidth]{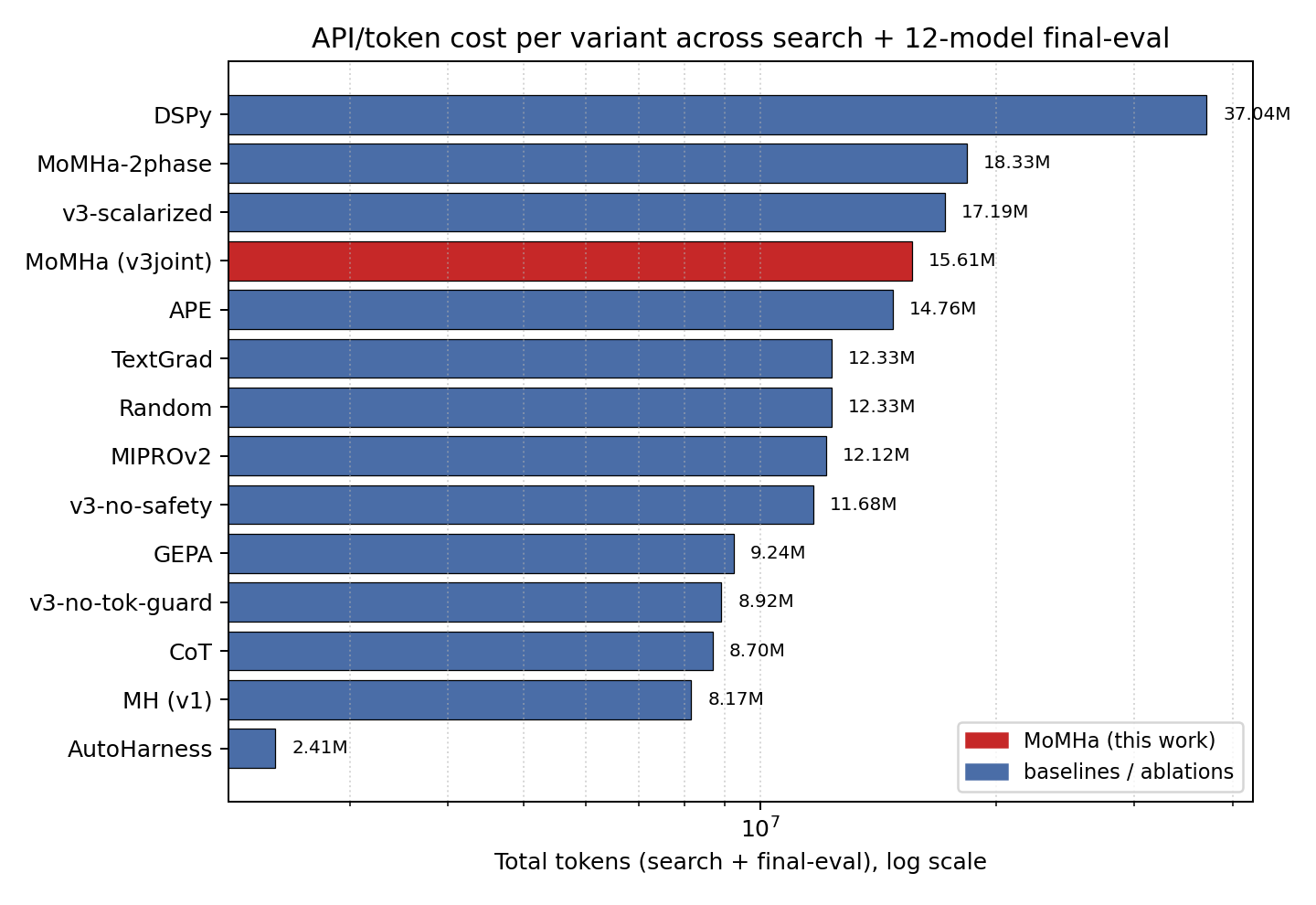}
\caption{Total search-time + final-evaluation tokens per variant (log
scale), sorted ascending. MoMHa (v3joint) totals 15.61M tokens, below
2-phase (18.33M) and DSPy (37.04M).}
\label{fig:cost_bars_app}
\end{figure}

\Cref{fig:cost_bars_app} and \cref{tab:cost_app} report search-time
evaluator tokens plus the fixed 12-model final-evaluation cost for every
variant. The paper's ``$\le\!100$ proposer API calls per domain'' refers to
\emph{proposer} (\claudecode) turns, which are distinct from and negligible
next to evaluator-model tokens; evaluator tokens
(candidates $\times$ search-set size $\times$ output length) dominate the
totals below. MoMHa is \emph{cheaper} than 2-phase and DSPy and comparable
to APE; it is not the cost outlier the reviews assumed.

\begin{table}[h]
\centering
\caption{Total token cost per variant: search-time evaluator tokens plus
the fixed 12-model final-evaluation matrix. Sorted ascending by total.}
\label{tab:cost_app}
\small
\begin{tabular}{lrrr}
\toprule
Variant & Search-eval (M) & Final-eval (M) & Total (M) \\
\midrule
AutoHarness       & 0.00  & 2.41  & 2.41  \\
MH (v1)           & 2.71  & 5.46  & 8.17  \\
CoT               & 3.04  & 5.66  & 8.70  \\
MoMHa-noTok       & 1.35  & 7.57  & 8.92  \\
GEPA              & 3.62  & 5.62  & 9.24  \\
MoMHa-ns          & 4.91  & 6.77  & 11.68 \\
MIPROv2           & 6.40  & 5.72  & 12.12 \\
Random            & 6.32  & 6.01  & 12.33 \\
TextGrad          & 5.97  & 6.36  & 12.33 \\
APE               & 9.55  & 5.21  & 14.76 \\
\textbf{MoMHa (v3joint)} & \textbf{9.20} & \textbf{6.41} & \textbf{15.61} \\
MoMHa-scalar      & 10.39 & 6.80  & 17.19 \\
2-phase           & 11.02 & 7.31  & 18.33 \\
DSPy              & 7.69  & 29.35 & 37.04 \\
\bottomrule
\end{tabular}
\end{table}

\subsection{Proposer-Model Ablation}
\label{app:proposer_ablation}

To test whether MoMHa's gains are specific to \claudecode's coding ability,
we reran the full search on 3 representative domains
(math\_reasoning, agentic\_coding, safety\_qa\_illegal) with three
non-Claude proposers (\texttt{llama-3-3-70b}, \texttt{deepseek-r1},
\texttt{gpt-5.4}), holding the evaluator model fixed at
\texttt{claude-haiku-4.5} and using the same search budget as the headline
runs. \Cref{tab:proposer_ablation_app} reports the joint score $J$ per
domain.

\begin{table}[h]
\centering
\caption{Proposer-model ablation. $J$ score per domain, 3 non-Claude
proposers vs.\ the headline \texttt{claude-sonnet-4.6} proposer.
$\dagger$: \texttt{gpt-5.4} generated banned \texttt{getattr()} calls,
failing AST validation on math and coding (falls back to the baseline
harness, $J\!\approx\!0$).}
\label{tab:proposer_ablation_app}
\small
\begin{tabular}{lrrrr}
\toprule
Proposer & math\_reasoning & agentic\_coding & safety\_qa\_illegal & Mean $J$ \\
\midrule
\textbf{claude-sonnet-4.6 (MoMHa)} & 0.103 & \textbf{0.606} & 0.530 & 0.413 \\
llama-3-3-70b   & \textbf{0.239} & 0.372 & \textbf{0.953} & \textbf{0.521} \\
deepseek-r1     & 0.157 & 0.409 & 0.797 & 0.454 \\
gpt-5.4         & 0.000$^\dagger$ & 0.000$^\dagger$ & 0.709 & 0.236$^\dagger$ \\
\bottomrule
\end{tabular}
\end{table}

Both open-weight proposers complete every search run and achieve mean $J$
comparable to or higher than \texttt{claude-sonnet-4.6} (0.521 / 0.454 vs.\
0.413), with no special prompting beyond the standard harness contract.
\textbf{MoMHa's search framework is proposer-agnostic}: the choice of
proposer changes \emph{which} strategies are discovered but not
\emph{whether} discovery succeeds.

\subsection{Scalarization Robustness}
\label{app:scalarization_robustness}

\begin{figure}[h]
\centering
\includegraphics[width=0.75\textwidth]{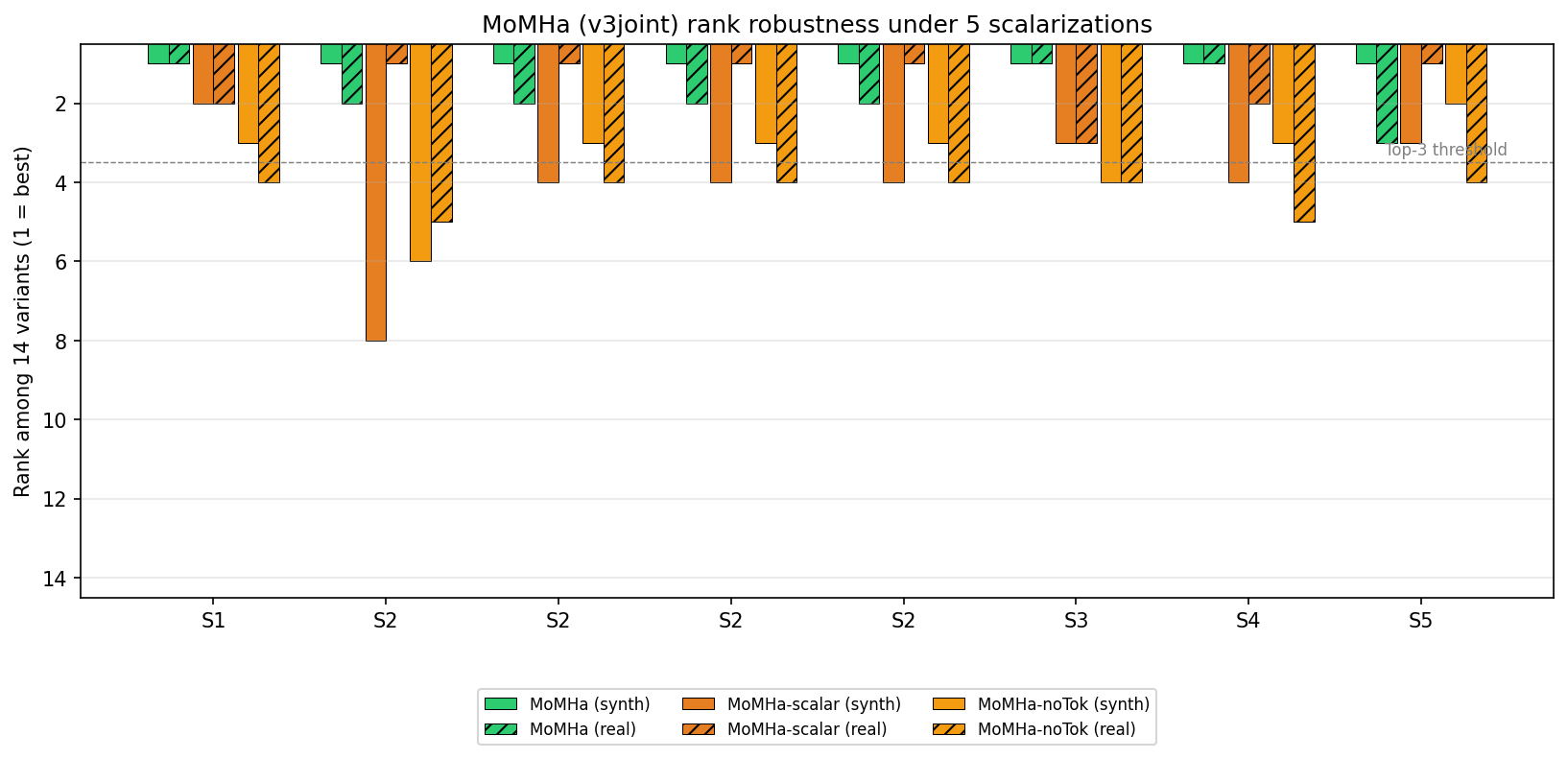}
\caption{MoMHa's rank (lower is better) across 8 scalarizations and both
evaluation tracks. MoMHa is rank 1 on the synthetic track under every
scalarization tested.}
\label{fig:scalarization_ranks_app}
\end{figure}

Reviewers questioned whether MoMHa's advantage is an artifact of the
composite score $J = 6\cdot\mathrm{acc}\cdot\mathrm{Safe}/\ln(\mathrm{tokens})$.
We re-rank all 14 variants under 5 additional scalarizations on both
tracks: a weighted-sum sweep (accuracy weight $\in\{0.4,0.5,0.6,0.7\}$), a
product-only score with no log-tokens denominator, a per-task (rather than
global) safety multiplier, and a raw Pareto-dominance count with no
scalarization at all. \Cref{fig:scalarization_ranks_app} and
\cref{tab:scalarization_app} report MoMHa's rank under each.

\begin{table}[h]
\centering
\caption{MoMHa (v3joint) rank under 8 scalarizations, out of 14 variants.}
\label{tab:scalarization_app}
\small
\begin{tabular}{lrr}
\toprule
Scalarization & Synthetic rank & Real-world rank \\
\midrule
S1: Composite $J$ (paper)                & 1 & 1 \\
S2: Weighted sum, $w{=}0.4$              & 1 & 2 \\
S2: Weighted sum, $w{=}0.5$              & 1 & 2 \\
S2: Weighted sum, $w{=}0.6$              & 1 & 2 \\
S2: Weighted sum, $w{=}0.7$              & 1 & 2 \\
S3: Product acc$\cdot$Safe, no log-tokens & 1 & 1 \\
S4: Per-task safety                      & 1 & 1 \\
S5: Pareto dominance count                & 1 & 3 \\
\bottomrule
\end{tabular}
\end{table}

MoMHa is rank~1 on the synthetic track under every scalarization tested,
and top-3 on the real-world track under every scalarization (worst case:
rank~3 under S5 Pareto-dominance count, behind two other MoMHa-family
variants). Dropping the composite score does not change the conclusion:
the log-tokens denominator, the global safety multiplier, and the specific
composite form are not what drive the ranking.

\subsection{Skill-File Ablation}
\label{app:skill_ablation}

To isolate how much of MoMHa's gain comes from the proposer's search
process versus prior domain knowledge encoded in the skill files, we
stripped \texttt{proposer\_<d>.md} and \texttt{safety\_<d>.md} to empty
stubs (keeping only the shared, domain-agnostic
\texttt{proposer\_skill.md}) and reran the search on 2 domains.

\begin{table}[h]
\centering
\caption{Skill-file ablation: $J$ score with full per-domain skill files
vs.\ empty stubs (framework-only fallback).}
\label{tab:skill_ablation}
\small
\begin{tabular}{lrrr}
\toprule
Domain & Full $J$ & Stub $J$ & $\Delta J$ \\
\midrule
sql\_generation & 0.263 & 0.247 & $-0.016$ ($-6\%$) \\
safety\_autonomous\_physical & 0.712 & 0.855 & $+0.143$ ($+20\%$) \\
\bottomrule
\end{tabular}
\end{table}

On SQL generation, removing the domain skill file costs $-6\%$ $J$: the
named strategy hints (\texttt{test\_driven\_generation},
\texttt{decompose\_query}) provide modest directional guidance. On
safety-autonomous-physical the stub \emph{outperforms} the full skill file
by $+20\%$: with the skill file the proposer converges on
\texttt{profile\_first\_single\_call} ($J{=}0.712$); without it, it
discovers \texttt{conservative\_default} ($J{=}0.855$), suggesting the
skill file can narrow exploration to a locally suboptimal strategy. Net,
across the two domains the effect is close to neutral in the direction
that would inflate MoMHa's numbers: \textbf{the proposer's adaptive
search, not prior encoding in skill files, drives the bulk of MoMHa's $J$
gains}.

\subsection{Second-Judge U-SafeBench Cross-Check}
\label{app:second_judge}

To check that the behavioral-safety composite is not an artifact of a
single LLM judge, we re-scored MoMHa's and MH's U-SafeBench predictions
with a second judge model (\texttt{gpt-5.4}) independent of the paper's
primary judge (\texttt{claude-sonnet-4.6}) on all three safety domains.

\begin{table}[h]
\centering
\caption{Inter-judge agreement (Cohen's $\kappa$) and accuracy under the
second judge, U-SafeBench domains.}
\label{tab:second_judge_app}
\small
\begin{tabular}{lrrrr}
\toprule
Domain & $n$ & Paper judge acc.\ & 2nd-judge acc.\ & $\kappa$ \\
\midrule
safety\_qa\_illegal          & 60 & 0.808 & 0.775 & 0.786 \\
safety\_autonomous\_physical & 30 & 0.883 & 0.933 & 0.769 \\
safety\_autonomous\_mental   & 30 & 0.950 & 0.967 & 0.941 \\
\bottomrule
\end{tabular}
\end{table}

All three domains show strong inter-judge agreement ($\kappa > 0.76$), and
the two judges disagree on only 8--12\% of examples with no systematic
direction. Against the v1 (MH) baseline, MoMHa's margin under the second
judge is $+0.32$ to $+0.39$ accuracy points on all three domains, too
large to be explained by inter-judge noise, confirming the
MoMHa~$>$~MH safety ordering is judge-independent.

\section{Proposer Inner Loop and Search Mechanics}
\label{app:proposer_loop}

This appendix expands the \cref{sec:architecture} summary into the full
state representation, mutation catalogue, screening gates, and harness
contract underlying MoMHa's search loop.

\paragraph{State the proposer sees.}
At iteration $i$ the proposer reads (via \texttt{grep}/\texttt{cat}/\texttt{diff},
never a single mega-prompt) five artifacts from
\texttt{domains/<d>/runs/}: (i)~the source code of every prior candidate
\texttt{candidate\_*.py}; (ii)~a redacted run history: candidate id, aggregate
accuracy, per-example token count, safety composite, joint reward $R$, parent
id, three most common wrong predictions, produced by
\texttt{RunLogger.list\_runs} and \texttt{compose\_history} in
\texttt{scripts/run\_search\_v3.py}; (iii)~per-example \texttt{traces.jsonl}
for the current best harness, containing one JSON row per LLM call with
input, output, latency, and \texttt{tokens\_used}; (iv)~the search-set
inputs (never test-set traces, which are held out by
\texttt{data/test\_set.jsonl}); and (v)~the composed skill text
$s_d = \texttt{proposer\_skill.md} \Vert \texttt{safety\_<d>.md} \Vert
\texttt{proposer\_<d>.md}$, assembled per run by \texttt{compose\_skill()}.
Ground-truth labels are stripped from history before it reaches the proposer.

\paragraph{Mutation step.}
Each mutation is a \emph{single} structural transformation applied to the
current best harness. The allowed action set is fixed and declared in the
skill files (\texttt{skills/proposer\_skill.md}, per-domain
\texttt{skills/proposer\_<d>.md}, and the token-focused
\texttt{skills/proposer\_token\_skill.md}). Concretely, the proposer may:
(a)~rewrite the prompt template (system message, user template, label list,
few-shot examples); (b)~add or prune retrieval-augmentation over the search
set; (c)~tighten the output parser (regex, JSON-mode fallback, label
normalization); (d)~introduce or remove a verification / retry / back-off
loop; (e)~route calls between models or between prompt variants on a
confidence gate; (f)~prune the few-shot pool or the DDL schema shown to the
model; and (g)~toggle chain-of-thought on or off, or cap its
\texttt{max\_tokens}. Per-domain skill files parameterize this set with a
domain-specific \emph{strategy library}, e.g.\
\texttt{draft\_then\_verify}, \texttt{subject\_aware\_routing}, or
\texttt{verification\_cascade} for math reasoning
(\texttt{STRATEGY\_LIBRARY} in \texttt{run\_search\_v3.py} lines 44--115).
The mutation is always additive: skills instruct the proposer to
``prefer additive changes over destructive rewrites''
(\texttt{proposer\_skill.md}, rule~5). This constraint keeps the search
Markovian in code space and prevents catastrophic regressions between
iterations.

\paragraph{Screening cutoff.}
A proposed candidate goes through four gates before it can compete for
acceptance: (1)~\textbf{AST guard} (\texttt{core/ast\_guard.py}) parses the
code and rejects any banned import / call / attribute, with the banlist
domain-parameterized by \texttt{safety\_<d>.md} (\cref{sec:safety});
rejection is silent: the candidate is logged and the loop continues.
(2)~\textbf{Mock validation} (\texttt{core/validator.py}) dynamically
imports the module, instantiates the \texttt{Harness} class against a
\texttt{MockLLMClient} that returns canned responses, and runs it on three
dummy examples, this catches ${\sim}90\%$ of malformed candidates in
${<}1$~second without spending any real API calls. (3)~\textbf{Budgeted
screening eval}: the surviving candidate is run on the 50-example search
set under a resource guard (\texttt{core/resource\_guard.py}: 100k tokens,
200 API calls, 300~s wall clock) using the primary evaluation model. (4)~a
\textbf{joint-reward floor}: candidates whose $R$ falls below the running
best minus a small margin are dropped and not considered for refinement.
Empirically, each domain executes ${\sim}100$ proposer API calls of which
${\sim}30$ pass all four gates and reach the acceptance step.

\paragraph{Acceptance rule.}
The current best is replaced iff the new candidate achieves \emph{strict}
improvement on the joint reward $R$ (Eq.~\eqref{eq:joint_reward}), with
$\lambda_s{=}1.0$ and $\lambda_t$ scaled so a 1k-token reduction is worth
1pp of accuracy ($\lambda_t = 10^{-5}$ per token in
\texttt{run\_search\_v3.py}; the multi-objective aggregation is
$R = \mathrm{acc} - \lambda\,(\mathrm{tokens}/\tau)$ with $\lambda{=}0.1$,
$\tau{=}1000$ in the joint-scalarization path of
\texttt{scripts/run\_search\_joint.py}). Ties on $R$ are broken by lower
token count. The proposer is not compressed down to the scalar~$R$: it
reads all three raw axes plus per-example traces, and only the search-time
ranking is scalarized.

\paragraph{Why single-phase matters (concrete).}
In the 2-phase ablation, Phase~1 optimizes $\mathrm{accuracy} +
\lambda_s\mathrm{safety}$ and then \emph{freezes} the harness structure;
Phase~2 (\texttt{proposer\_token\_skill.md}) can only \emph{prune} that
structure subject to a $2$pp accuracy tolerance. This locks in whatever
verbose chain-of-thought pattern Phase~1 converged on: on fact
verification the Phase-1 harness is a two-call decompose-then-verify at
${\sim}1483$ tokens/example, and Phase~2 cannot reach the alternative
single-pass confidence-gated verdict (${\sim}536$ tokens/example) that
MoMHa's single-phase joint search discovers, because that structure is
outside Phase~2's mutation neighborhood. The overall effect on the ten-domain
mean is a $+2.7$ joint-score-point gain for MoMHa over 2-phase ($0.611$ vs.\
$0.584$) at $95$ fewer tokens per example ($572$ vs.\ $667$;
\cref{app:v3family_ablation}).

\paragraph{What the proposer discovers vs.\ what the skill files encode.}
The skill files
(\texttt{skills/proposer\_skill.md}, \texttt{skills/proposer\_<d>.md},
\texttt{skills/safety\_<d>.md}) encode the \emph{search space}: the harness
contract (must define \texttt{Harness.\_\_init\_\_(client, config)} and
\texttt{run(example)$\to$\{"prediction": ...\}}), the LLM-client call
convention, the domain's refusal templates and output-format constraints,
the allowed and banned symbol lists at that domain's risk level (strict /
minimal / light), and the finite strategy library
$\{\texttt{draft\_verify},\texttt{subject\_routing}, \ldots\}$ that scopes
the mutation step. The proposer discovers the actual \emph{composition}:
which subset of strategies to combine, in what order, with what
prompt-template text, at what threshold parameters, and with which
retry/back-off cap. Two evidence points support this decomposition. First,
converged harnesses on the same domain but different eval models
instantiate the same skill-library strategy with different code
(\cref{app:strategies}). Second, the MoMHa-ns ablation
(\textbf{v3ns} in \cref{tab:joint}) strips the per-domain safety skill file
and keeps everything else fixed; this degrades the ten-domain mean joint
score from $0.472$ (MoMHa) to $0.407$ (MoMHa-ns), a $-0.065$-point drop
attributable purely to the domain-specific safety knowledge the file
encodes. The skill-file ablation (\cref{tab:skill_ablation},
\cref{app:rebuttal_robustness}) further strips \emph{both} the safety and
the strategy-library skill files back to a framework-only stub on
\texttt{sql\_generation} and \texttt{safety\_autonomous\_physical},
isolating the additional contribution of prior encoded strategy names; the
joint-score deltas quantify how much of MoMHa's gain is proposer discovery
vs.\ prior encoding.

\paragraph{Harness contract (scope).}
Concretely, a harness is a Python class exposing
\texttt{Harness.\_\_init\_\_(client, config)} and
\texttt{run(example) $\to$ \{"prediction": ..., "tokens\_used": ...\}},
validated by \texttt{core/validator.py} before any real API call. It
\emph{can}: build, rewrite, or reject prompts; route calls across models or
prompt variants; retry with modified inputs; parse, redact, or reject
outputs; and invoke sub-harnesses in a loop. It \emph{cannot}: modify
model weights; escape the sandbox (\texttt{core/sandbox.py}: 30~s per call,
100k tokens, 200 API calls, 300~s wall clock, \texttt{/tmp} CWD); or exceed
the AST-guard's per-domain import/call whitelist (\cref{sec:safety}).

%% file: checklist.tex
\section*{NeurIPS Paper Checklist}

\begin{enumerate}

\item {\bf Claims}
    \item[] Question: Do the main claims made in the abstract and introduction accurately reflect the paper's contributions and scope?
    \item[] Answer: \answerYes{}
    \item[] Justification: All headline claims, MoMHa's superior joint score on 17 domains, the advantage of joint over staged optimization, cross-model transfer, and the safety gains from the per-domain safety skill, are directly supported by \cref{tab:joint,tab:joint_realworld,tab:crossmodel_J} and the ablation tables in the appendix. Numerical claims in the introduction (e.g., $+2.7$ over 2-phase, $0.781$ safety composite) match the reported experimental results exactly.

\item {\bf Limitations}
    \item[] Question: Does the paper discuss the limitations of the work performed by the authors?
    \item[] Answer: \answerYes{}
    \item[] Justification: \Cref{sec:conclusion} contains a dedicated ``Limitations'' paragraph acknowledging: (i) capability datasets are LLM-generated rather than established benchmarks, limiting direct comparability; (ii) the search process requires ${\sim}100$ harness evaluations per domain, incurring non-trivial API cost; (iii) cross-model transfer degrades when models differ substantially in output-format conventions; and (iv) results may differ with alternative proposer systems beyond \claudecode.

\item {\bf Theory assumptions and proofs}
    \item[] Question: For each theoretical result, does the paper provide the full set of assumptions and a complete (and correct) proof?
    \item[] Answer: \answerNA{}
    \item[] Justification: The paper is empirical and contains no theorems or formal proofs. \Cref{eq:objective,eq:pareto,eq:joint_reward} define the optimization objectives and the joint evaluation metric; these are design choices stated without theoretical claims.

    \item {\bf Experimental result reproducibility}
    \item[] Question: Does the paper fully disclose all the information needed to reproduce the main experimental results of the paper to the extent that it affects the main claims and/or conclusions of the paper (regardless of whether the code and data are provided or not)?
    \item[] Answer: \answerYes{}
    \item[] Justification: \Cref{sec:method,sec:setup} specify the full system architecture, reward functions, data splits (50 search / 50 test), search budget (${\sim}100$ evaluations per domain), the three-layer safety stack, skill-composition procedure, and the 12-model evaluation fleet. Appendix \cref{app:token_details,app:crossmodel,app:v3family_ablation} provide per-domain and per-model breakdowns. All harness source, search logs, per-example traces, scoring artifacts, and evaluation JSONs are released as supplemental material.

\item {\bf Open access to data and code}
    \item[] Question: Does the paper provide open access to the data and code, with sufficient instructions to faithfully reproduce the main experimental results, as described in supplemental material?
    \item[] Answer: \answerYes{}
    \item[] Justification: We will release all harness source code, per-domain proposer and safety skill files, search logs, per-example execution traces, scoring artifacts, and cross-model evaluation JSONs as supplemental material. Anonymized versions are provided at submission time. The seven real-world benchmark domains use publicly available HuggingFace datasets (LawBench, NuminaMath, FEVER, Spider, HumanEval, MBPP, MMLU-Pro) cited in the paper.

\item {\bf Experimental setting/details}
    \item[] Question: Does the paper specify all the training and test details (e.g., data splits, hyperparameters, how they were chosen, type of optimizer) necessary to understand the results?
    \item[] Answer: \answerYes{}
    \item[] Justification: \Cref{sec:setup} reports data splits (50 search / 50 test per capability domain), search budget (${\sim}100$ evaluations; 724 total), baseline configurations (APE: 5 rounds $\times$ 5 candidates; DSPy: BootstrapFewShot; MIPROv2: TPE surrogate), reward-function definitions for each domain type, the joint metric $J$ (\cref{eq:joint}), and the LLM-as-judge scoring protocol for safety domains. Harness discovery uses Claude Haiku~4.5 for all variants. Full token-budget details are in \cref{app:token_details}.

\item {\bf Experiment statistical significance}
    \item[] Question: Does the paper report error bars suitably and correctly defined or other appropriate information about the statistical significance of the experiments?
    \item[] Answer: \answerYes{}
    \item[] Justification: All cells in the six evaluation tables (\cref{tab:joint_realworld,tab:joint,tab:crossmodel_J,tab:v1v3_app,tab:threeway_app,tab:tokens_app}) report mean${}_{\pm\text{SEM}}$, where SEM is a 1-$\sigma$ standard error of the mean computed as $\sigma/\sqrt{N_{\text{trials}}}$. 

\item {\bf Experiments compute resources}
    \item[] Question: For each experiment, does the paper provide sufficient information on the computer resources (type of compute workers, memory, time of execution) needed to reproduce the experiments?
    \item[] Answer: \answerYes{}
    \item[] Justification: The paper reports the search budget (${\sim}100$ harness evaluations per domain, 724 total across all domains) and per-example token counts for each variant (e.g., MoMHa averages 672 tokens, DSPy 2{,}229 tokens on capability domains). All compute is API-based (no local GPU required).
    
\item {\bf Code of ethics}
    \item[] Question: Does the research conducted in the paper conform, in every respect, with the NeurIPS Code of Ethics \url{https://neurips.cc/public/EthicsGuidelines}?
    \item[] Answer: \answerYes{}
    \item[] Justification: The paper studies automated harness optimization with safety as a first-class objective. The three-layer safety architecture (AST-based static analysis, domain-specific safety skills, sandboxed execution) is designed to prevent generated code from executing unsafe operations. All model API usage follows the respective providers' terms of service. No human subjects are involved.

\item {\bf Broader impacts}
    \item[] Question: Does the paper discuss both potential positive societal impacts and negative societal impacts of the work performed?
    \item[] Answer: \answerYes{}
    \item[] Justification: The positive impact, reducing unsafe LLM outputs via behavioral safety as an explicit optimization objective, is discussed throughout \cref{sec:safety,sec:usafebench,sec:conclusion}. A potential negative impact is that the framework could be used to search for adversarial harnesses. We note two mitigations: (1)~the AST guard and sandbox prevent generated harnesses from executing filesystem or network operations \emph{during search}; (2)~discovered harnesses are inspectable Python files subject to the same review process as any other code. However, neither protection applies if a malicious actor deliberately deploys a discovered harness in an unrestricted environment; this residual risk is acknowledged.
    
\item {\bf Safeguards}
    \item[] Question: Does the paper describe safeguards that have been put in place for responsible release of data or models that have a high risk for misuse (e.g., pre-trained language models, image generators, or scraped datasets)?
    \item[] Answer: \answerYes{}
    \item[] Justification: \Cref{sec:safety} describes the three-layer safety system built into \metaharness: (i) AST-based static analysis (\astguard) that rejects any harness importing banned modules or calling dangerous builtins; (ii) per-domain safety skills with calibrated import whitelists; and (iii) sandboxed execution with 30-second timeouts, 100K token limits, and an empty environment. Released harnesses have all passed \astguard and cannot perform filesystem or network operations.

\item {\bf Licenses for existing assets}
    \item[] Question: Are the creators or original owners of assets (e.g., code, data, models), used in the paper, properly credited and are the license and terms of use explicitly mentioned and properly respected?
    \item[] Answer: \answerYes{}
    \item[] Justification: All baseline methods (APE, OPRO, DSPy, MIPROv2, TextGrad, GEPA, MH, AH) and datasets (U-SafeBench, LawBench, NuminaMath, FEVER, Spider, HumanEval, MBPP, MMLU-Pro) are cited via their original publications. The seven real-world benchmarks are sourced from HuggingFace under their respective open licenses. The LLM-as-judge uses Claude Sonnet~4.6 via the Anthropic API under standard commercial terms.

\item {\bf New assets}
    \item[] Question: Are new assets introduced in the paper well documented and is the documentation provided alongside the assets?
    \item[] Answer: \answerYes{}
    \item[] Justification: The paper will release: (i) all harness source files (\texttt{domains/<d>/harnesses/candidate\_*.py}); (ii) per-domain proposer skill files (\texttt{skills/proposer\_<domain>.md}) and safety skill files (\texttt{skills/safety\_<domain>.md}); (iii) search logs, per-example execution traces (\texttt{traces.jsonl}), and scoring artifacts (\texttt{scores.json}); and (iv) cross-model evaluation JSONs. Structure and schema are documented in \cref{sec:architecture} and the appendix. 

\item {\bf Crowdsourcing and research with human subjects}
    \item[] Question: For crowdsourcing experiments and research with human subjects, does the paper include the full text of instructions given to participants and screenshots, if applicable, as well as details about compensation (if any)? 
    \item[] Answer: \answerNA{}
    \item[] Justification: No crowdsourcing or human subjects are involved. The capability datasets are LLM-generated (verified by a second model), the safety evaluation uses a Claude Sonnet~4.6 LLM-as-judge, and baseline/MoMHa evaluations are fully automated.

\item {\bf Institutional review board (IRB) approvals or equivalent for research with human subjects}
    \item[] Question: Does the paper describe potential risks incurred by study participants, whether such risks were disclosed to the subjects, and whether Institutional Review Board (IRB) approvals (or an equivalent approval/review based on the requirements of your country or institution) were obtained?
    \item[] Answer: \answerNA{}
    \item[] Justification: No human subjects are involved in any part of this work; IRB approval is therefore not required.

\item {\bf Declaration of LLM usage}
    \item[] Question: Does the paper describe the usage of LLMs if it is an important, original, or non-standard component of the core methods in this research? Note that if the LLM is used only for writing, editing, or formatting purposes and does \emph{not} impact the core methodology, scientific rigor, or originality of the research, declaration is not required.
    \item[] Answer: \answerYes{}
    \item[] Justification: LLMs are a central component of the method. \Cref{sec:architecture} describes: (i) \claudecode as the agentic proposer that reads filesystem artifacts and writes new harness candidates; (ii) Claude Haiku~4.5 as the model used during harness search; (iii) Claude Sonnet~4.6 as the LLM-as-judge for U-SafeBench safety scoring. The cross-model evaluation (\cref{sec:crossmodel}) runs discovered harnesses on 12 target models from four families (Claude, GPT, Gemini, DeepSeek-R1), all described in \cref{tab:models}.

\end{enumerate}

%% file: references.bib
@article{zhou2023large,
  title={Large Language Models Are Human-Level Prompt Engineers},
  author={Zhou, Yongchao and Muresanu, Andrei Ioan and Han, Ziwen and Paster, Keiran and Pitis, Silviu and Chan, Harris and Ba, Jimmy},
  journal={arXiv preprint arXiv:2211.01910},
  year={2023}
}

@article{pryzant2023automatic,
  title={Automatic Prompt Optimization with ``Gradient Descent'' and Beam Search},
  author={Pryzant, Reid and Iter, Dan and Li, Jerry and Lee, Yin Tat and Zhu, Chenguang and Zeng, Michael},
  journal={arXiv preprint arXiv:2305.03495},
  year={2023}
}

@article{khattab2024dspy,
  title={{DSPy}: Compiling Declarative Language Model Calls into Self-Improving Pipelines},
  author={Khattab, Omar and Singhvi, Arnav and Maheshwari, Paridhi and Zhang, Zhiyuan and Santhanam, Keshav and Vardhamanan, Sri and Haq, Saiful and Sharma, Ashutosh and Joshi, Thomas T and Mober, Hanna and others},
  journal={arXiv preprint arXiv:2310.03714},
  year={2024}
}

@article{cheng2024trace,
  title={Trace is the Next AutoDiff: Generative Optimization with Rich Feedback, Execution Traces, and {LLM}s},
  author={Cheng, Ching-An and Nie, Allen and Swaminathan, Adith},
  journal={arXiv preprint arXiv:2406.16218},
  year={2024}
}

@article{yuksekgonul2024textgrad,
  title={{TextGrad}: Automatic ``Differentiation'' via Text},
  author={Y{\"u}ksekg{\"o}n{\"u}l, Mert and Bianchi, Federico and Boen, Joseph and Liu, Sheng and Lu, Pan and Huang, Zhi and Guestrin, Carlos and Zou, James},
  journal={arXiv preprint arXiv:2406.07496},
  year={2024}
}

@article{shin2020autoprompt,
  title={{AutoPrompt}: Eliciting Knowledge from Language Models with Automatically Generated Prompts},
  author={Shin, Taylor and Razeghi, Yasaman and Logan IV, Robert L and Wallace, Eric and Singh, Sameer},
  journal={arXiv preprint arXiv:2010.15980},
  year={2020}
}

@article{yang2024large,
  title={Large Language Models as Optimizers},
  author={Yang, Chengrun and Wang, Xuezhi and Lu, Yifeng and Liu, Hanxiao and Le, Quoc V and Zhou, Denny and Chen, Xinyun},
  journal={arXiv preprint arXiv:2309.03409},
  year={2024}
}

@article{jimenez2024swebench,
  title={{SWE-bench}: Can Language Models Resolve Real-World {GitHub} Issues?},
  author={Jimenez, Carlos E and Yang, John and Wettig, Alexander and Yao, Shunyu and Pei, Kexin and Press, Ofir and Narasimhan, Karthik},
  journal={arXiv preprint arXiv:2310.06770},
  year={2024}
}

@inproceedings{merrill2026terminalbench,
  title={Terminal-Bench: Benchmarking Agents on Hard, Realistic Tasks in Command Line Interfaces},
  author={Merrill, Mike A. and Shaw, Alexander G. and Carlini, Nicholas and Li, Boxuan and Raj, Harsh and Bercovich, Ivan and Shi, Lin and others},
  booktitle={ICLR 2026},
  year={2026},
  note={arXiv preprint arXiv:2601.11868}
}

@article{pearce2022asleep,
  title={Asleep at the Keyboard? Assessing the Security of {GitHub Copilot}'s Code Contributions},
  author={Pearce, Hammond and Ahmad, Baleegh and Tan, Benjamin and Dolan-Gavitt, Brendan and Karri, Ramesh},
  journal={IEEE Symposium on Security and Privacy},
  year={2022}
}

@article{ruan2024toolemu,
  title={Identifying the Risks of {LM} Agents with an {LM}-Emulated Sandbox},
  author={Ruan, Yangjun and Dong, Honghua and Wang, Andrew and Pitis, Silviu and Zhou, Yongchao and Ba, Jimmy and Dubois, Yann and Maddison, Chris J and Hashimoto, Tatsunori},
  journal={arXiv preprint arXiv:2309.15817},
  year={2024},
  note={ICLR 2024}
}

@article{perez2021true,
  title={True Few-Shot Learning with Language Models},
  author={Perez, Ethan and Kiela, Douwe and Cho, Kyunghyun},
  journal={Advances in Neural Information Processing Systems},
  year={2021}
}

@article{opsahlOng2024miprov2,
  title={Optimizing Instructions and Demonstrations for Multi-Stage Language Model Programs},
  author={Opsahl-Ong, Krista and Ryan, Michael J. and Purtell, Josh and Broman, David and Potts, Christopher and Zaharia, Matei and Khattab, Omar},
  journal={arXiv preprint arXiv:2406.11695},
  year={2024},
  note={EMNLP 2024}
}

@article{agrawal2025gepa,
  title={{GEPA}: Reflective Prompt Evolution Can Outperform Reinforcement Learning},
  author={Agrawal, Lakshya A. and Tan, Shangyin and Soylu, Dilara and Ziems, Noah and Khare, Rishi and Opsahl-Ong, Krista and Singhvi, Arnav and Shandilya, Herumb and Ryan, Michael J. and Jiang, Meng and Potts, Christopher and Sen, Koushik and Dimakis, Alexandros G. and Stoica, Ion and Klein, Dan and Zaharia, Matei and Khattab, Omar},
  journal={arXiv preprint arXiv:2507.19457},
  year={2025},
  note={ICLR 2026 Oral}
}

@inproceedings{in2025usafebench,
  title={Is Safety Standard Same for Everyone? {User-Specific} Safety Evaluation of Large Language Models},
  author={In, Yeonjun and Kim, Wonjoong and Yoon, Sein and Kim, Sungchul and Tanjim, Mehrab and Kim, Sangwu and Park, Chanyoung},
  booktitle={Findings of the Association for Computational Linguistics: {EMNLP} 2025},
  year={2025},
  note={arXiv preprint arXiv:2502.15086}
}

@article{lee2026metaharness,
  title={Meta-Harness: End-to-End Optimization of Model Harnesses},
  author={Lee, Yoonho and Nair, Roshen and Zhang, Qizheng and Lee, Kangwook and Khattab, Omar and Finn, Chelsea},
  journal={arXiv preprint arXiv:2603.28052},
  year={2026}
}

@article{lou2026autoharness,
  title={AutoHarness: improving LLM agents by automatically synthesizing a code harness},
  author={Lou, Xinghua and L{\'a}zaro-Gredilla, Miguel and Dedieu, Antoine and Wendelken, Carter and Lehrach, Wolfgang and Murphy, Kevin P},
  journal={arXiv preprint arXiv:2603.03329},
  year={2026}
}

@article{pymoo2020,
  author={J. {Blank} and K. {Deb}},
  journal={IEEE Access},
  title={pymoo: Multi-Objective Optimization in {Python}},
  year={2020},
  volume={8},
  pages={89497--89509},
}

@article{deb2002nsga2,
  title     = {A fast and elitist multiobjective genetic algorithm: {NSGA-II}},
  author    = {Deb, Kalyanmoy and Pratap, Amrit and Agarwal, Sameer and Meyarivan, T.},
  journal   = {IEEE Transactions on Evolutionary Computation},
  volume    = {6},
  number    = {2},
  pages     = {182--197},
  year      = {2002},
  doi       = {10.1109/4235.996017}
}

@article{zhang2007moead,
  title     = {{MOEA/D}: A multiobjective evolutionary algorithm based on decomposition},
  author    = {Zhang, Qingfu and Li, Hui},
  journal   = {IEEE Transactions on Evolutionary Computation},
  volume    = {11},
  number    = {6},
  pages     = {712--731},
  year      = {2007},
  doi       = {10.1109/TEVC.2007.892759}
}

@article{zitzler1999multiobjective,
  title     = {Multiobjective evolutionary algorithms: A comparative case study
               and the strength {Pareto} approach},
  author    = {Zitzler, Eckart and Thiele, Lothar},
  journal   = {IEEE Transactions on Evolutionary Computation},
  volume    = {3},
  number    = {4},
  pages     = {257--271},
  year      = {1999},
  doi       = {10.1109/4235.797969}
}

@inproceedings{alizadeh2024pessimistic,
  title     = {Pessimistic off-policy multi-objective optimization},
  author    = {Alizadeh, Shima and Bhargava, Aniruddha and Gopalswamy, Karthick and
               Jain, Lalit and Kveton, Branislav and Liu, Ge},
  booktitle = {International Conference on Artificial Intelligence and Statistics},
  pages     = {2980--2988},
  year      = {2024},
  organization = {PMLR}
}

@article{mukherjee2024moallm,
  title     = {Multi-Objective Alignment of Large Language Models Through
               Hypervolume Maximization},
  author    = {Mukherjee, Subhojyoti and Lalit, Anusha},
  journal   = {arXiv preprint arXiv:2412.05469},
  year      = {2024}
}

@article{liang2022holistic,
  title={Holistic evaluation of language models},
  author={Liang, Percy and Bommasani, Rishi and Lee, Tony and Tsipras, Dimitris and Soylu, Dilara and Yasunaga, Michihiro and Zhang, Yian and Narayanan, Deepak and Wu, Yuhuai and Kumar, Ananya and others},
  journal={arXiv preprint arXiv:2211.09110},
  year={2022}
}

@article{du2025ockbench,
  title={{OckBench}: Measuring the Efficiency of {LLM} Reasoning},
  author={Du, Zheng and Kang, Hao and Han, Song and Krishna, Tushar and Zhu, Ligeng},
  journal={arXiv preprint arXiv:2511.05722},
  year={2025}
}

@article{kaiser2026decomposing,
  title={Beyond Accuracy: Decomposing the Reasoning Efficiency of {LLM}s},
  author={Kaiser, Daniel and Frigessi, Arnoldo and Ramezani-Kebrya, Ali and Ricaud, Benjamin},
  journal={arXiv preprint arXiv:2602.09805},
  year={2026}
}

@article{inan2023llamaguard,
  title={Llama Guard: {LLM}-based Input-Output Safeguard for Human-{AI} Conversations},
  author={Inan, Hakan and Upasani, Kartikeya and Chi, Jianfeng and Rungta, Rashi and Iyer, Krithika and Mao, Yuning and Tontchev, Michael and Hu, Qing and Fuller, Brian and Testuggine, Davide and Khabsa, Madian},
  journal={arXiv preprint arXiv:2312.06674},
  year={2023}
}

@article{zeng2024shieldgemma,
  title={{ShieldGemma}: Generative {AI} Content Moderation Based on Gemma},
  author={Zeng, Wenjun and Liu, Yuchi and Mullins, Ryan and Peran, Ludovic and Fernandez, Joe and Harkous, Hamza and Narasimhan, Karthik and Proud, Drew and Kumar, Piyush and Radharapu, Bhaktipriya and Sturman, Olivia and Wahltinez, Oscar},
  journal={arXiv preprint arXiv:2407.21772},
  year={2024}
}

@article{rebedea2023nemoguardrails,
  title={{NeMo} Guardrails: A Toolkit for Controllable and Safe {LLM} Applications with Programmable Rails},
  author={Rebedea, Traian and Dinu, Razvan and Sreedhar, Makesh Narsimhan and Parisien, Christopher and Cohen, Jonathan},
  journal={arXiv preprint arXiv:2310.10501},
  year={2023},
  note={EMNLP 2023 Demo Track}
}

@article{lester2021promptuning,
  title={The Power of Scale for Parameter-Efficient Prompt Tuning},
  author={Lester, Brian and Al-Rfou, Rami and Constant, Noah},
  journal={arXiv preprint arXiv:2104.08691},
  year={2021},
  note={EMNLP 2021}
}

@article{li2021prefixtuning,
  title={Prefix-Tuning: Optimizing Continuous Prompts for Generation},
  author={Li, Xiang Lisa and Liang, Percy},
  journal={arXiv preprint arXiv:2101.00190},
  year={2021}
}

@article{turner2023activation,
  title={Steering Language Models with Activation Engineering},
  author={Turner, Alexander Matt and Thiergart, Lisa and Leech, Gavin and Udell, David and Vazquez, Juan J. and Mini, Ulisse and MacDiarmid, Monte},
  journal={arXiv preprint arXiv:2308.10248},
  year={2023}
}

@article{zou2023repe,
  title={Representation Engineering: A Top-Down Approach to {AI} Transparency},
  author={Zou, Andy and Phan, Long and Chen, Sarah and Campbell, James and Guo, Phillip and Ren, Richard and Pan, Alexander and Yin, Xuwang and Mazeika, Mantas and Dombrowski, Ann-Kathrin and Goel, Shashwat and Li, Nathaniel and Byun, Michael J. and Wang, Zifan and Mallen, Alex and Basart, Steven and Koyejo, Sanmi and Song, Dawn and Fredrikson, Matt and Kolter, J. Zico and Hendrycks, Dan},
  journal={arXiv preprint arXiv:2310.01405},
  year={2023}
}
